%% file: article.tex
\documentclass{svjour3}                     
\smartqed  
\usepackage{graphicx}
\usepackage[usenames,dvipsnames]{xcolor}
\usepackage{amsmath}
\usepackage{amssymb}
\usepackage{hyperref}
\usepackage{natbib}
\usepackage[section]{placeins}
\usepackage{lscape} 
\usepackage{etoolbox}
\usepackage{lineno}


\newtoggle{arxiv}
\toggletrue{arxiv}

\begin{document}

\iftoggle{arxiv}{}{

\linenumbers}

\title{The advent and fall of a vocabulary learning bias from communicative efficiency 
}


\author{David Carrera-Casado \and Ramon Ferrer-i-Cancho
}


\institute{David Carrera-Casado \& Ramon Ferrer-i-Cancho \at
           Complexity and Quantitative Linguistics Lab \\
           LARCA Research Group \\
           Departament de Ci\`encies de la Computaci\'o \\
           Universitat Polit\`ecnica de Catalunya \\
           Campus Nord, Edifici Omega\\
           Jordi Girona Salgado 1-3 \\
           08034 Barcelona, Catalonia, Spain \\
           \email{david.carrera@estudiantat.upc.edu,rferrericancho@cs.upc.edu}           
}

\date{Received: date / Accepted: date}

\maketitle

\begin{abstract}
Biosemiosis is a process of choice-making between simultaneously alternative options. It is well-known that, when sufficiently young children encounter a new word, they tend to interpret it as pointing to a meaning that does not have a word yet in their lexicon rather than to a meaning that already has a word attached. In previous research, the strategy was shown to be optimal from an information theoretic standpoint. In that framework, interpretation is hypothesized to be driven by the minimization of a cost function: the option of least communication cost is chosen. However, the information theoretic model employed in that research neither explains the weakening of that vocabulary learning bias in older children or polylinguals nor reproduces Zipf's meaning-frequency law, namely the non-linear relationship between the number of meanings of a word and its frequency. Here we consider a generalization of the model that is channeled to reproduce that law. The analysis of the new model reveals regions of the phase space where the bias disappears consistently with the weakening or loss of the bias in older children or polylinguals. The model is abstract enough to support future research on other levels of life that are relevant to biosemiotics.
In the deep learning era, the model is a transparent low-dimensional tool for future experimental research and illustrates the predictive power of a theoretical framework  originally designed to shed light on the origins of Zipf's rank-frequency law. 
\keywords{biosemiosis \and vocabulary learning \and mutual exclusivity \and Zipfian laws \and information theory \and quantitative linguistics}
\end{abstract}

\iftoggle{arxiv}{\tableofcontents}{}

\section{Introduction}

\label{sec:introduction}

\input{introduction}



\section{The mathematical model}

\label{sec:model}




\input{model}

\section{Results}

\label{sec:results}


\input{results}

\section{Discussion}

\label{sec:discussion}

\input{discussion}

\begin{acknowledgements}
We are grateful to two anonymous reviewers for their valuable feeback and recommendations to improve the article.  We are also grateful to A. Hern\'andez-Fern\'andez and G. Boleda for their revision of the article and many recommendations to improve it. The article has benefited from discussions with T. Brochhagen, S. Semple and M. Gustison. Finally, we thank C. Hobaiter for her advice and inspiration for future research.
DCC and RFC are supported by the grant TIN2017-89244-R from MINECO (Ministerio de Econom\'ia, Industria y Competitividad). RFC is also supported by the recognition 2017SGR-856 (MACDA) from AGAUR (Generalitat de Catalunya). 
\end{acknowledgements}

%
%

\bibliographystyle{spbasic}      
\bibliography{biblio}   

%
%

\appendix

\section{The mathematical model in detail}

\label{app:model}

\input{model_appendix}

\section{Form degrees and number of links}

\label{app:form_degrees_appendix}

\input{form_degrees_appendix}

\section{Complementary heatmaps for other values of $\phi$}

\label{app:other_values_phi}

\input{other_values_phi_appendix}

\section{Complementary figures with discrete degrees}

\label{app::discrete_degrees}

\input{discrete_degrees_appendix}

\end{document}

%% file: introduction.tex
Biosemiotics can be defined as a science of signs in living systems \citep[p. 386]{Kull1999a}. Here we join the effort of developing such a science. Focusing on the problem of ``learning'' new signs, we hope to contribute (i) to place choice at the core of semiotic theory of learning \citep{Kull2018a} and (ii) to make biosemiotics compatible with the information theoretic perspective that is regarded as currently dominant in physics, chemistry, and molecular biology \citep{Deacon2015a}.

\begin{figure}
\includegraphics[width=\textwidth]{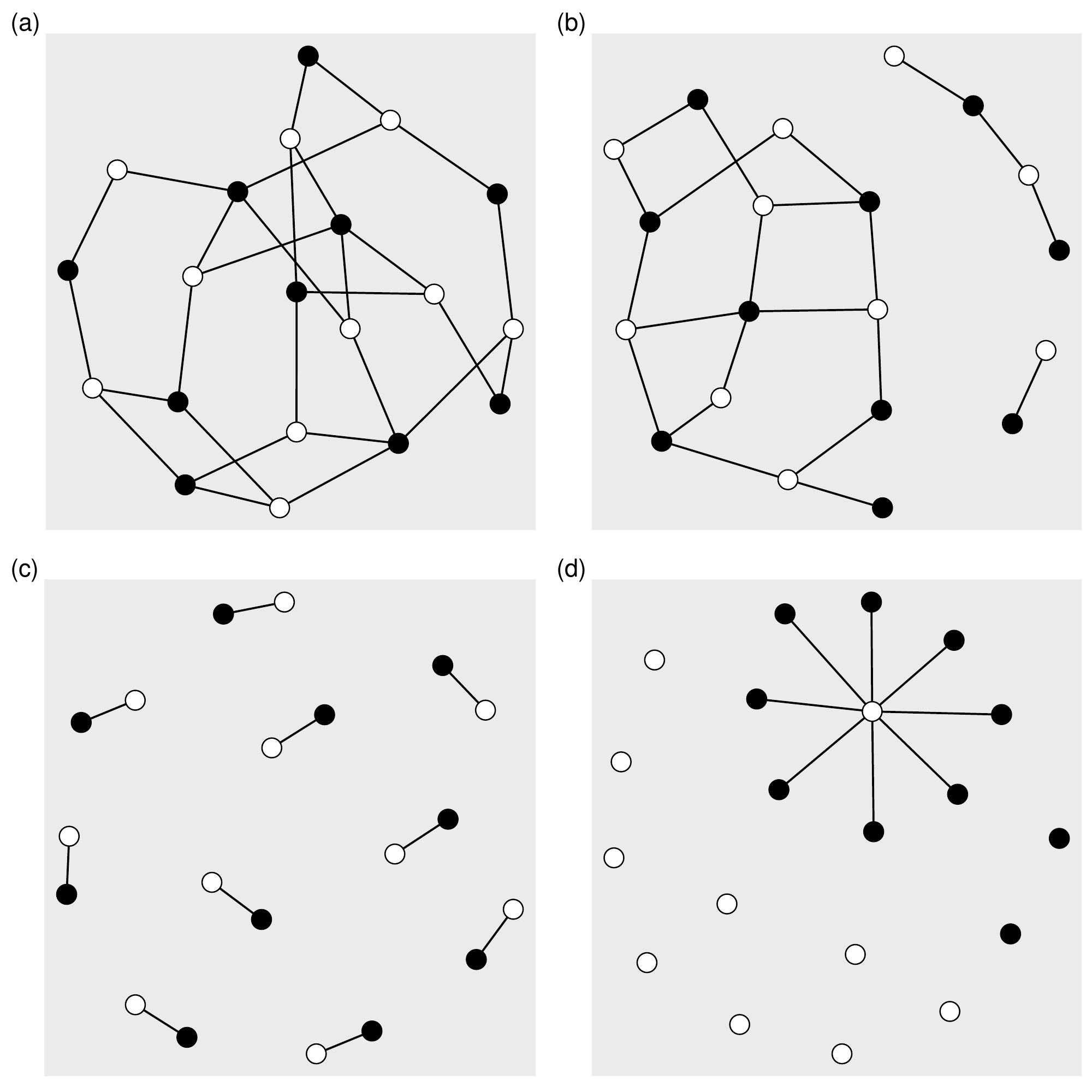}
\caption{\label{bipartite_graph_figure} A bipartite graph linking forms (white circles) with their counterparts (black circles). (a) a connected graph (b) an almost connected graph (c) a one-to-one mapping between forms and counterparts (d) a mapping where only one form is linked with counterparts. }
\end{figure}

\begin{figure}
\includegraphics[width=\textwidth]{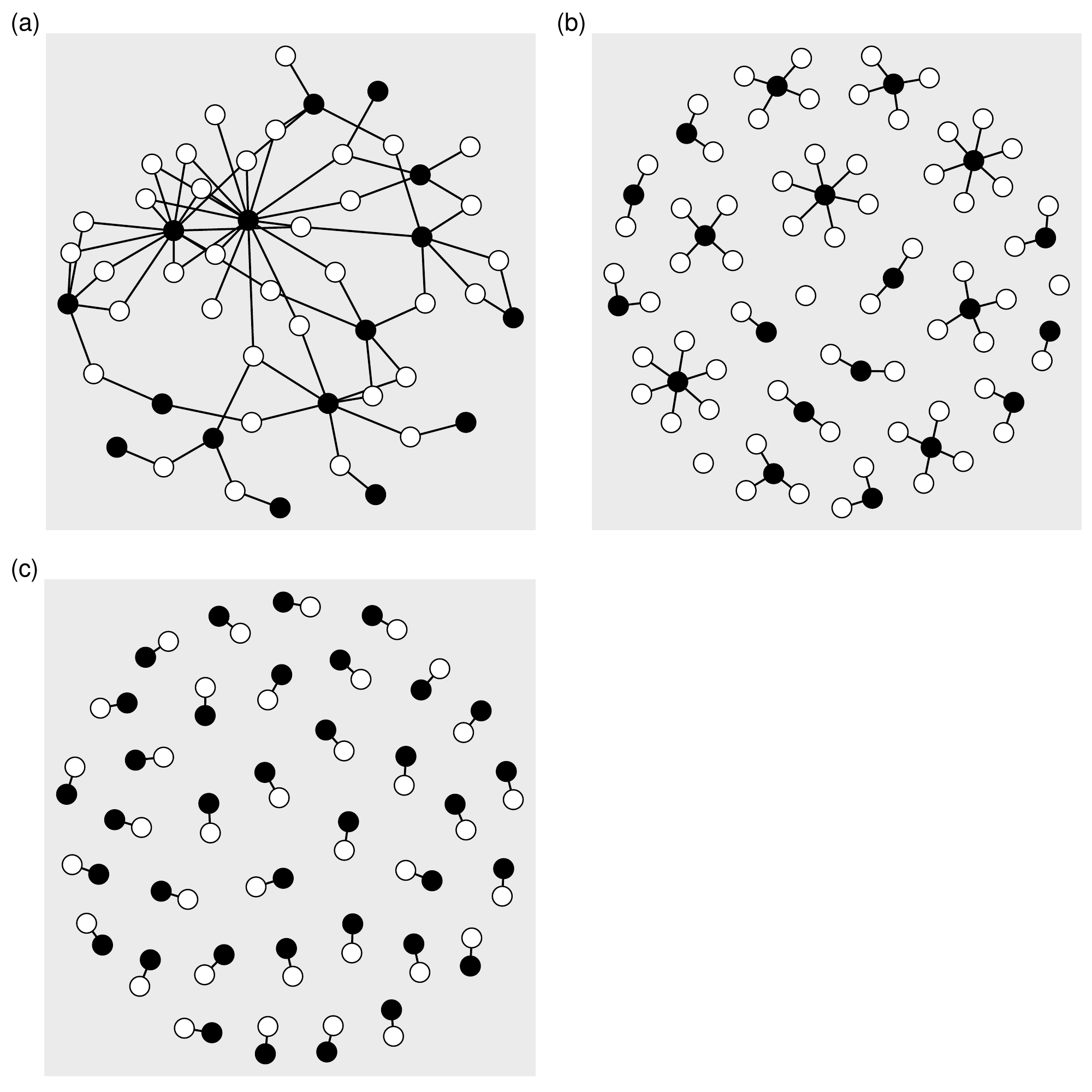}
\caption{\label{real_bipartite_graph_figure} Real bipartite graphs linking forms (white circles) with their counterparts (black circles). (a) Chimpanzee gestures and their meaning \citep[Table S3]{Hobaiter2014a}. This table was chosen for its broad coverage of gesture types (see other tables satisfying other constraints, e.g. only gesture-meaning associations employed by a sufficiently large number of individuals). (b) Codon translation into amino acids, where forms are 64 codons and counterparts are 20 amino acids 
(c) The international Morse code, where forms are strings of dots and dashed and the counterparts are letters of the English alphabet ($A,B,...,Z$) and digits ($0,1,...,9$). }
\end{figure}

Languages use words to convey information. From a semantic perspective, words stand for meanings \citep{Fromkin2014a}. Correlates of word meaning have been investigated in other species \cite[e.g.][]{Hobaiter2014a,Genty2014a,Moore2014a}.
From a neurobiological perspective, words can be seen as the counterparts of cell assemblies with distinct cortical topographies \citep{Pulvermuller2001,Pulvermuller2013a}.  
From a formal standpoint, the essence of that research is some binding between a sign or a form, e.g., a word or an ape gesture, and a counterpart, e.g. a 'meaning' or an assembly of cortical cells. Mathematically, that binding can be formalized as a bipartite graph where vertices are forms and their counterparts (Fig. \ref{bipartite_graph_figure}).
Such abstract setting allows for a powerful exploration of natural systems across levels of life, from the mapping of animal vocal or gestural behaviors (Fig. \ref{real_bipartite_graph_figure} (a)) into their ``meanings'' down to the mapping from codons into amino acids (Figure \ref{real_bipartite_graph_figure} (b)) while allowing for a comparison against ``artificial'' coding systems such as the Morse code (Fig. \ref{real_bipartite_graph_figure} (c)) or those emerging in artificial naming games \citep{Hurford1989a,Steels1996a}.
In that setting, almost connectedness has been hypothesized to be the mathematical condition required for the emergence of a rudimentary form of syntax and symbolic reference \citep{Ferrer2004f,Ferrer2005e}. By symbolic reference, we mean here Deacon's revision of Pierce's view \citep{Deacon1997a}. The almost connectedness condition is met when it is possible to reach practically any other vertex of the network by starting a walk from any possible vertex (as in Fig. \ref{bipartite_graph_figure} (a)-(b) but not in Figs. \ref{bipartite_graph_figure} (c)-(d)).

Since the pioneering research of G. K. \citet{Zipf1949a}, statistical laws of language have been interpreted as manifestations of the minimization of cognitive costs \citep{Zipf1949a,Ellis1986,Ferrer2007a,Gustison2016a,Ferrer2019c}. Zipf argued that the law of abbreviation, the tendency of more frequent words to be shorter, resulted from a minimization of a cost function involving, for every word, its frequency, its ``mass'' and its ``distance'', which in turn implies the minimization of the size of words \citep[p.59]{Zipf1949a}. Recently, it as been shown mathematically that the minimization of the average of the length of words (the mean code length in the language of information theory) predicts a correlation between frequency and duration that cannot be positive, extending and generalizing previous results from information theory \citep{Ferrer2019c}. The framework addresses the general problem of assigning codes as short as possible to counterparts represented by distinct numbers while warranting certain constraints, e.g., that every number will receive a distinct code (e.g. non-singular coding in the language of information theory). If the counterparts are word types from a vocabulary, it predicts the law of abbreviation as it occurs in the vast majority of languages \citep{Bentz2016a}. If these counterparts are meanings, it predicts that more frequent meanings should tend to be assigned smaller codes (e.g., shorter words) as found in real experiments \citep{Kanwal2017a,Brochhagen2021a}. Table \ref{tab:prediction} summarizes these and other predictions of compression.     

\begin{landscape}
\begin{table}
\begin{tabular}{lll}
{\em linguistic laws}      & $\longrightarrow$ {\em principles}                                                                                & $\longrightarrow$ {\em predictions} \\
                           &                                                                                                                   & \citep{Koehler1987,Altmann1993} \\ 
\\
Zipf's law of abbreviation & $\longrightarrow$ compression                                                                                     & $\longrightarrow$ Menzerath's law \\                          
                           &                                                                                                                   & \citep{Gustison2016a,Ferrer2019c} \\
\\                          
                           &                                                                                                                   & $\longrightarrow$ Zipf's rank-frequency law \\                          
                           &                                                                                                                   & \citep{Ferrer2016b} \\
                           &                                                                                                                   & \\
                           &                                                                                                                   & $\longrightarrow$ ``shorter words'' for more frequent ``meanings'' \\                          
                           &                                                                                                                   & \citep{Ferrer2019c,Kanwal2017a,Brochhagen2021a} \\
                           &                                                                                                                   & \\
Zipf's rank-frequency law  & $\longrightarrow$ \begin{tabular}{cc}mutual information maximization \\ + \\ surprisal minimization \end{tabular} & $\longrightarrow$ {\bf a vocabulary learning bias} \\
                           &                                                                                                                   & \citep{Ferrer2013g} \\
\\
                           &                                                                                                                   & $\longrightarrow$ the principle of contrast \\                          &                                                                                                                   & \citep{Ferrer2013g} \\
\\
                           &                                                                                                                   & $\longrightarrow$ range or variation of $\alpha$ \\                          &                                                                                                                   & \citep{Ferrer2004a,Ferrer2005e} \\
\\                          

\end{tabular}
\caption{\label{tab:prediction} The application of the scientific method in quantitative linguistics (italics) with various concrete examples (roman). 
$\alpha$ is the exponent of Zipf's rank-frequency law \citep{Zipf1949a}. The prediction that is the target of the current article is shown in boldface. }
\end{table}
\end{landscape}

\subsection{A family of probabilistic models}

The bipartite graph of form-counterpart associations is the {\em skeleton}  (Figs. \ref{bipartite_graph_figure} and \ref{real_bipartite_graph_figure}) on which a family of models of communication has been built \citep{Ferrer2007a,Ferrer2017b}. The target of the first of these models \citep{Ferrer2002a} was Zipf's rank-frequency law, that defines the relationship between the frequency of a word $f$ and its rank $i$, approximately as 
\begin{equation*}
f \approx i^{-\alpha}.
\end{equation*}
These early models were aimed at shedding light on mainly three questions: 
\begin{enumerate}
\item
The origins of this law \citep{Ferrer2002a,Ferrer2004e}. 
\item
The range of variation of $\alpha$ in human language \citep{Ferrer2004a,Ferrer2005e}.
\item
The relationship between $\alpha$ and the syntactic and referential complexity of a communication system \citep{Ferrer2004f,Ferrer2005e}.  
\end{enumerate}
The main assumption of these models is that word frequency is an epiphenomenon of the structure of the skeleton or the probability of the meanings. Following the metaphor of the skeleton, the models are {\em bodies} whose {\em flesh} are probabilities that are calculated from the skeleton.
The first models defined $p(s_i| r_j)$, the probability that a speaker produces $s_i$ given a counterpart $r_j$, as the same for all words connected to $r_j$. In the language of mathematics, 
\begin{equation}
p(s_i | r_j) = \frac{a_{ij}}{\omega_j},
\label{form_conditional_probability_equation}
\end{equation}
where $a_{ij}$ is a boolean (0 or 1) that indicates if $s_i$ and $r_j$ are connected and $\omega_j$ is the degree of $r_j$, namely the number of connections of $r_j$ with forms, i.e. 
\begin{equation*}
\omega_j = \sum_{i} a_{ij}.
\end{equation*} 
These models are often portrayed as {\em models of the assignment of meanings to forms} \citep{Futrell2020_twitter,Piantadosi2014a} but 
this description falls short because:
\begin{itemize}
\item
They are indeed models of production as they define the probability of producing a form given some counterparts (as in Eq. \ref{form_conditional_probability_equation}) or simply the marginal probability of a form. The claim that {\em theories of language production or discourse do not explain the law} \citep{Piantadosi2014a} has no basis and raises the questions of which theories of language production are deemed acceptable. 
\item
They are also models of understanding, as they define symmetric conditional probabilities such as $p(r_j | s_i)$, the probability that a listener interprets $r_j$ when receiving $s_i$.
\item
The models are flexible. In addition to ``meaning'', other counterparts were deemed possible from their birth. See for instance the use of the term ``stimuli'' \citep[e.g.][]{Ferrer2007a}, as a replacement for meaning that was borrowed from neurolinguistics \citep{Pulvermuller2001}.  
\item
The models fit in the distributional semantics framework \citep{Lund1996a} for two reasons: their flexibility, as counterparts can be dimensions in some hidden space, and also because of representing a form as a vector of their joint or conditional probabilities with ``counterparts'' that is inferred from the network structure, as we have already explained \citep{Ferrer2017b}.  
\end{itemize}
Contrary to the conclusions of \citep{Piantadosi2014a}, there are derivations of Zipf’s law that do account for psychological processes of word production, especially the intentionality of choosing words in order to convey a desired meaning. 

The family of models assume that the skeleton that determines all the probabilities, the bipartite graph, is shaped by a combination of minimization of the entropy (or surprisal) of words ($H$) and the maximization of the mutual information between words and meanings ($I$), two principles that are cognitively motivated and that capture speaker and listener's requirements \citep{Ferrer2015b}. When only the entropy of words is minimized, configurations where only one form is linked as in Fig. \ref{bipartite_graph_figure} (d) are predicted. When only the mutual information between forms and counterparts is maximized, one-to-one mappings between forms and counterparts are predicted (when the number of forms and counterparts is the same) as in Figure \ref{bipartite_graph_figure} (c) or Fig. \ref{real_bipartite_graph_figure} (d). Real language is argued to be in-between these two extreme configurations \citep{Ferrer2007a}. 
Such a trade-off between simplicity (Zipf's unification) and effective communication (Zipf's diversification) is also found in information theoretic models of communication based on the information bottleneck approach (see \cite{Zaslavsky2021a} and references there in).

In quantitative linguistics, scientific theory is not possible without taking into consideration language laws \citep{Koehler1987,Debowski2020a}. Laws are seen as manifestations of principles (also referred as {\em ``requirements''} by \citet{Koehler1987}), which are key components of explanations of linguistic phenomena. As part of the scientific method cycle, novel predictions are key aim \citep{Altmann1993} and key to validation and refinement of theory \citep{Bunge2001a}. Table \ref{tab:prediction} synthesizes this general view as chains of the form: {\em laws}, {\em principles} that are inferred from them, and {\em predictions} that are made from those principles, giving concrete examples from previous research.

Although one of the initial goals of the family of models was to shed light on the origins of Zipf's law for word frequencies, a member of the family of models turned out to generate a novel prediction on vocabulary learning in children and the tendency of words to contrast in meaning \citep{Ferrer2013g}: when encountering a new word, children tend to infer that it refers to a concept that does not have a word attached to it \citep{Markman1988a,Merriman1989a,Clark1993}. The finding is cross-linguistically robust: it has been found in children speaking English \citep{Markman1988a}, Canadian French \citep{Nicoladis2020a}, Japanese \citep{Haryu1991a}, Mandarin Chinese \citep{Byers-Heinlein2013a, Hung2015a}, Korean \citep{EunNam2017a}. These languages correspond to four distinct linguistic families (Indo-European, Japonic, Sino-Tibetan, Koreanic). Furthermore, the finding has also been replicated in adults \citep{Hendrickson2019a, Yurovsky2008a} and other species \cite{Kaminski2004a}. This phenomenon is a example of biosemiosis, namely a process of choice-making between simultaneously alternative options \citep[p. 454]{Kull2018a}.




\begin{figure}
\includegraphics[width=\textwidth]{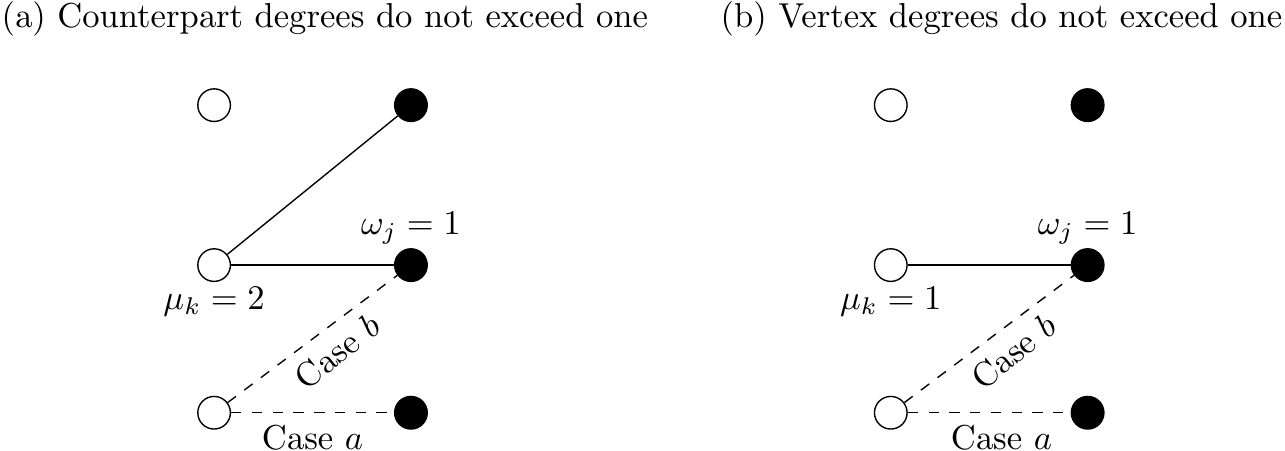}
\caption{\label{vocabulary_learning_figure} Strategies for linking a new word to a meaning. Strategy $a$ consists of linking a word to a free meaning, namely an unlinked meaning. Strategy $b$ consists of linking a word to a meaning that is already linked. We assume that the meaning that is already linked is connected to a single word of degree $\mu_k$. Two simplifying assumptions are considered. (a) Counterpart degrees do not exceed one, implying $\mu_k \geq 1$. (b) Vertex degrees do not exceed one, implying $\mu_k = 1$.}
\end{figure}

As an explanation for vocabulary learning, the information theoretic model suffers from some limitations that motivate the present article. The first one is that the vocabulary learning bias weakens in older children \citep{Kalashnikova2016a, Yildiz2020a} or in polylinguals \citep{Houston-Price2010a, Kalashnikova2015a}, while the current version of the model predicts the vocabulary learning bias only provided that mutual information maximization is not neglected \citep{Ferrer2013g}. 

The second limitation is inherited from the family of models, where the definition of the probabilities over the bipartite graph skeleton leads to a linear relationship between the frequency of a form and its number of counterparts \citep{Ferrer2017b}. However, this is inconsistent with Zipf's prediction, namely that the number of meanings $\mu$ a word of frequency $f$ should follow \citep{Zipf1945a}
\begin{equation}
\mu \approx f^{\delta},
\label{meaning-frequency_law_equation}
\end{equation} 
with $\delta = 0.5$. Eq. \ref{meaning-frequency_law_equation} is known as Zipf's meaning-frequency law \citep{Zipf1949a}.
To overcome such a limitation, \citet{Ferrer2017b} proposed different ways of modifying the definition of the probabilities from the skeleton. Here we borrow a proposal of defining the joint probability of a form and its counterpart as
\begin{equation}
p(s_i, r_j) \propto a_{ij}(\mu_i\omega_j)^{\phi},
\label{joint_probability_equation}
\end{equation} 
where $\phi$ is a parameter of the model and $\mu_i$ and $\omega_j$ are, respectively, the degree (number of connections) of the form $s_i$ and the counterpart $r_j$.  
Previous research on vocabulary learning in children with these models \citep{Ferrer2013g} assumed $\phi = 0$, which leads to $\delta = 1$ \citep{Ferrer2014d}. When $\phi = 1$, the system is channeled to reproduce Zipf's meaning-frequency law, i.e. Eq. \ref{meaning-frequency_law_equation} with $\delta=0.5$ \citep{Ferrer2017b}. 

\subsection{Overview of the present article}

It has been argued that there cannot be meaning without interpretation \citep{Eco1986a}. 
As \citet{Kull2020a} puts it, ``{\em Interpretation (which is the same as primitive decision-making) assumes that there exists a choice between two or more
options. The options can be described as different codes applicable simultaneously in the same situation.}''
The main aim to of this article is to shed light on the choice between strategy $a$, i.e. attaching the new form to a counterpart that is unlinked, and strategy $b$, i.e. attaching the new form to a counterpart that is already linked (Fig. \ref{vocabulary_learning_figure}).

The remainder of the article is organized as follows. Section \ref{sec:model} considers a model of a communication system that has three components: 
\begin{enumerate}
\item
A {\em skeleton} that is defined by a binary matrix $A$ that indicates the form-counterpart connections. 
\item
A {\em flesh} that is defined over the skeleton with Eq. \ref{joint_probability_equation},
\item 
A {\em cost function}, that defines the cost of communication as 
\begin{equation}
\Omega = -\lambda I + (1-\lambda)H,
\label{cost_equation}  
\end{equation}
where $\lambda$ is a parameter that regulates the weight of mutual information ($I$) maximization and word entropy ($H$) minimization such that $ 0 \leq \lambda \leq 1$. $I$ and $H$ are inferred from matrix $A$ and Eq. \ref{joint_probability_equation} (further details are given in Section \ref{sec:model}). 
\end{enumerate} 
This section introduces $\Delta$, i.e. the difference in the cost of communication between strategy $a$ and strategy $b$ according to $\Omega$ (Fig. \ref{vocabulary_learning_figure}). $\Delta < 0$ indicates that the cost of communication of strategy $a$ is lower than that of $b$. Our main hypothesis is that interpretation is driven by the $\Omega$ cost function and that a receiver will choose the option that minimizes the resulting $\Omega$. By doing this, we are challenging the longstanding and limiting belief that information theory is dissociated from semiotics and not concerned about meaning \citep[e.g.][]{Deacon2015a}. This article is a just one counterexample (see also \citet{Zaslavsky2018a}). Information theory, as any abstract powerful mathematical tool, can serve applications that do not assume meaning (or meaning-making processes) as in the original setting of telecommunication where it was developed by Shannon, as well as others that do, although they were not his primary concern for historical and sociological reasons.

In general, the formula of $\Delta$ is complex and the analysis of the conditions where $a$ is advantageous (namely $\Delta < 0$) requires making some simplifying assumptions. 
If $\phi = 0$, then one obtains that \cite{Ferrer2013g}
\begin{equation}
  \Delta  = -\lambda \frac{(\omega_j + 1) \log (\omega_j + 1) - \omega_j \log(\omega_j)}{M+1},
  \label{eq:delta_lambda_phi_0}
\end{equation}
where $M$ is the number of edges in the skeleton and $\omega_j$ is the degree of the already linked counterpart that is selected in strategy $b$ (Fig. \ref{vocabulary_learning_figure}). Eq. \ref{eq:delta_lambda_phi_0} indicates that strategy $a$ will be advantageous provided that mutual information maximization matters (i.e. $\lambda >0$) and its advantage will increase as mutual information maximization becomes more important (i.e. for larger $\lambda$), the linked counterpart has more connections (i.e. larger $\omega_j$) or when the skeleton has less connections (i.e. smaller $M$).
To be able to analyze the case $\phi > 0$, we will examine two classes of skeleta that are presented next.

\paragraph{Counterpart degrees do not exceed one.} In this class, the degrees of counterparts are restricted to not exceed one, namely a counterpart can only be disconnected or connected to just one form.
If meanings are taken as counterparts, this class matches the view that {\em ``no two words ever have {\em exactly} the same meaning''} \citep[p. 256]{Fromkin2014a}, based on the notion of absolute synonymy \citep{Dangli2009a}. 
This class also mirrors the linguistic principle that any two words should contrast in meaning \citep{Clark1987a}. 
Alternatively, if synonyms are deemed real to some extent, this class may capture early stages of language development in children or early stages in the evolution of languages where synonyms have not been learned or developed. 
From a theoretical standpoint, this class is required by the maximization of the mutual information between forms and counterparts when the number of forms does not exceed that of counterparts \citep{Ferrer2017b}.

We use $\mu_k$ to refer to degree of the word that will be connected to meaning selected in strategy $b$ (Fig. \ref{vocabulary_learning_figure}). We will show that, in this class, $\Delta$ is determined by $\lambda$, $\phi$, $\mu_k$ and the degree distribution of forms, namely the vector of form degrees $\vec{\mu}=(\mu_1,...,\mu_i,...\mu_n)$.

\paragraph{Vertex degrees do not exceed one.} In this class, the degrees of any vertex are restricted to not exceed one, namely a form (or a meaning) can only be disconnected or connected to just one counterpart (just one form).
This class is narrower than the previous one because it imposes that degrees do not exceed one both for forms and counterparts. Words lack homonymy (or polysemy). We believe that this class would correspond to even earlier stages of language development in children (where children have learned at most one meaning of a word) or earlier stages in the evolution of languages (where the communication system has not developed any homonymy). From a theoretical stand point, that class is a requirement of maximizing mutual information between forms and counterparts when $n = m$ \citep{Ferrer2017b}. We will show that $\Delta$ is determined just by $\lambda$, $\phi$ and $M$, the number of links of the bipartite skeleton.

Notice that meanings with synonyms have been found in chimpanzee gestures \citep{Hobaiter2014a}, which suggests that the two classes above do not capture the current state of the development of form-counterpart mappings in adults of other species.
Section \ref{sec:model} presents the formulae of $\Delta$ for each classes. 
Section \ref{sec:results} uses this formulae to explore the conditions that determine when strategy $a$ is more advantageous, namely $\Delta < 0$, for each of the two classes of skeleta above, that correspond to different stages of the development of language in children. While the condition $\phi = 0$ implies that strategy $a$ is always advantageous when $\lambda >0$, we find regions of the space of parameters where this is not the case when $\phi >0$ and $\lambda>0$.
In the more restrictive class, where vertex degrees do not exceed one, we find a region where $a$ is not advantageous when $\lambda$ is sufficiently small and $M$ is sufficiently large.
The size of that region increases as $\phi$ increases. From a complementary perspective, we find a region where $a$ is not advantageous ($\Delta \geq 0$) when $\lambda$ is sufficiency small and $\phi$ is sufficiently large; the size of the region increases as $M$ increases.  As $M$ is expected to be larger in older children or in polylinguals (if the forms of each language are mixed in the same skeleton), the model predicts the weakening of the bias in older children and polylinguals \citep{Liittschwager1994a,Kalashnikova2016a,Yildiz2020a,Houston-Price2010a,Kalashnikova2015a,Kalashnikova2019a}.
To ease the exploration of the phase space for the class where the degrees of counterparts do not exceed one, we will assume that word frequencies follow Zipf's rank-frequency law. Again, regions where $a$ is not advantageous ($\Delta \geq 0$) also appear but the conditions for the emergence of this regions are more complex. Our preliminary analyses suggest that the bias should weaken in older children even for this class. Section \ref{sec:discussion} discusses the findings, suggests future research directions and reviews the research program in light of the scientific method.

%% file: model.tex
Below we give more details about the model that we use to investigate the learning of new words and outlines the arguments that take from Eq.  \ref{joint_probability_equation} to concrete formulae of $\Delta$. Section
\ref{subsection_delta_formulae}
just presents the concrete formulae $\Delta$ for each of the two classes of skeleta.
Full details are given in Appendix \ref{app:model}. The model has four components that we review next. 

\paragraph{Skeleton ($A=a_{ij}$).} A bipartite graph that defines the associations between $n$ forms and $m$ counterparts that are defined by an adjacency matrix $A = \{a_{ij}\}$.

\paragraph{Flesh ($p(s_i,r_j)$).} The flesh consist of a definition of $p(s_i, r_j)$, the joint probability of a form (or word) and a counterpart (or meaning) and a series of probability definitions stemming from it. Probabilities depart from previous work \citep{Ferrer2002a,Ferrer2004e} by the addition of the parameter $\phi$. Eq. \ref{joint_probability_equation} defines 
$p(s_i, r_j)$ as proportional to the product of the degrees of the form and the counterpart to the power of $\phi$, which is a parameter of the model. By normalization, namely
\begin{equation*}
  \sum_{i=1}^n\sum_{j=1}^m p(s_i, r_j) = 1,
\end{equation*}
Eq. \ref{joint_probability_equation} leads to 
\begin{equation}
  p(s_i, r_j) = \frac{1}{M_\phi} a_{ij} (\mu_i \omega_j)^\phi,
  \label{normalized_joint_probability_equation}
\end{equation}
where
\begin{equation}
  M_\phi = \sum_{i=1}^n \sum_{j=1}^m a_{ij} (\mu_i \omega_j)^\phi.
  \label{normalization_factor_equation}
\end{equation}
From these expressions, the marginal probabilities of a form $p(s_i)$ and a counterpart $p(r_j)$ are obtained easily thanks to 
\begin{eqnarray*}
p(s_i) = \sum_{j=1}^m p(s_i, r_j)\\
p(r_j) = \sum_{i=1}^n p(s_i, r_j).
\end{eqnarray*}

\paragraph{The cost of communication ($\Omega$).}
The cost function is initially defined in Eq. \ref{cost_equation} as in previous research \cite[e.g.][]{Ferrer2007a}. In more detail, 
\begin{equation}
  \Omega = -\lambda I(S,R) + (1 - \lambda) H(S),
  \label{cost_equation_SR}
\end{equation}
where $I(S,R)$ is the mutual information between forms from a repertoire $S$ and counterparts from a repertoire $R$, and $H(S)$ is the entropy (or surprisal) of forms from a repertoire $S$.
Knowing that $I(S,R) = H(S) + H(R) - H(S,R)$ \cite{Cover2006a}, the final expression for the cost function in this article is 
\begin{equation}
  \Omega(\lambda) = (1-2\lambda) H(S) - \lambda H(R) + \lambda H(S,R).
  \label{cost_function_from_entropies}
\end{equation}
The entropies $H(S)$, $H(R)$ and $H(S,R)$ are easy to calculate applying the definitions of $p(s_i)$, $p(r_j)$ and $p(s_i,r_j)$, respectively. 

\paragraph{The difference in the cost of learning a new word ($\Delta$). }  
There are two possible strategies to determine the counterpart with which a new form (a previously unlinked form) should connect (Fig. \ref{vocabulary_learning_figure}):
\begin{enumerate}
\item[$a$.] Connect the new form to a counterpart that is not already connected to any other forms.
\item[$b$.] Connect the new form to a counterpart that is connected to at least one other form.
\end{enumerate}
The question we intend to answer is ``when does strategy $a$ result in a smaller cost than strategy $b$?'' Or, in the terminology of child language research, ``for which strategy is the assumption of mutual exclusivity more advantageous?'' To answer these questions, we define $\Delta$, as a the difference between the cost of each strategy. More precisely,   
\begin{equation}
  \Delta(\lambda) = \Omega'_a(\lambda) - \Omega'_b(\lambda),
  \label{eq:delta_from_omega_a_b}
\end{equation}
where $\Omega'_a(\lambda)$ and $\Omega'_b(\lambda)$ are the new value of $\Omega$ when a new link is created using strategy $a$ or $b$ respectively.
Then, our research question becomes ``When is $\Delta < 0$?''.

Formulae for $\Omega'_a(\lambda)$ and $\Omega'_b(\lambda)$ are derived in two steps. First, analyzing a general problem, i.e. $\Omega'$, the new value of $\Omega$ after producing a single mutation in $A$ (Appendix \ref{appendix_dynamic}). Second, deriving expressions for the case where that mutation results from linking a new form (an unlinked form) to a counterpart, that can be linked or unlinked
(Appendix \ref{appendix_vocabulary_learning_basic}).

\subsection{$\Delta$ in two classes of skeleta}
\label{subsection_delta_formulae}

In previous work, the value of $\Delta$ was already calculated for $\phi=0$, obtaining expressions equivalent to Eq. \ref{eq:delta_lambda_phi_0} (see Appendix \ref{vocabulary_learning_phi_0} for a derivation). The next sections just summarize the more complex formulae that are obtained for each class of skeleta for $\phi \geq 0$ (see Appendix \ref{app:model} for details on the derivation).

\subsubsection{Vertex degrees do not exceed one}

Here forms and counterparts both either have a single connection or are disconnected. Mathematically, this can be expressed as
\begin{align*}
  \mu_i \in \{0,1\} \mbox{~for each $i$ such that~} 1 \leq i \leq n \\
  \omega_j \in \{0,1\} \mbox{~for each $j$ such that~} 1 \leq j \leq m.
\end{align*}
Fig. \ref{vocabulary_learning_figure} (b) offers a visual representation of a bipartite graph of this class. In case $b$, the counterpart we connect the new form to is connected to only one form ($\omega_j=1$) and that form is connected to only one counterpart ($\mu_k=1$).
Under this class, $\Delta$ becomes 
\begin{equation}
  \Delta(\lambda) = (1 - 2\lambda) \left[ \log \left( 1 + \frac{2(2^\phi - 1)}{M + 1} \right) + \frac{2^{\phi+1} \log(2) \phi}{M + 2^{\phi+1} - 1} \right] - \lambda \frac{2^{\phi+1} \log(2)}{M + 2^{\phi+1} - 1},
  \label{eq:delta_w_mu_in_0_1}
\end{equation}
which can be rewritten as linear function of $\lambda$, i.e. 
\begin{equation*}
    \Delta(\lambda) = a \lambda + b,
\end{equation*}
with
\begin{align*}
  a &= 2 \log \left( 1 + \frac{2(2^\phi - 1)}{M + 1} \right) - (2 \phi + 1) \frac{2^{\phi+1} \log(2)}{M+2^{\phi+1}-1} \\
  b &= -\log \left( 1 + \frac{2(2^\phi - 1)}{M + 1} \right) + \phi \frac{2^{\phi+1} \log(2)}{M+2^{\phi+1}-1}.
\end{align*}
Importantly, notice that this expression of $\Delta$ is determined only by $\lambda$, $\phi$ and $M$ (the total number of links in the model).
See Appendix \ref{vocabulary_learning_wj_mui_in_01} for thorough derivations.

\subsubsection{Counterpart degrees do not exceed one}
\label{counterpart_degrees_do_not_exceed_one_section}

This class of skeleta is a relaxation of the previous class. Counterparts are either connected to a single form or disconnected. Mathematically,
\begin{equation*}
  \omega_j \in \{0,1\} \mbox{~for each $j$ such that~} 1 \leq j \leq m.
\end{equation*}
Fig. \ref{vocabulary_learning_figure} (a) offers a visual representation of a bipartite graph of this class. The number of forms the counterpart in case $b$ is connected to is still 1 ($\omega_j=1$) but this form may be connected to any number of counterparts; $\mu_k$ has to satisfy $1 \leq \mu_k \leq m$.

Under this class, $\Delta$ becomes 
\begin{equation}
  \begin{split}
    \Delta(\lambda) &= (1-2\lambda) \Bigg\{ \log \Bigg( \frac{M_\phi+1}{M_\phi+\left(2^\phi-1\right)\mu_k^\phi+2^\phi} \Bigg) \\
    &\qquad + \frac{1}{M_\phi+\left(2^\phi-1\right)\mu_k^\phi+2^\phi} \Bigg[ (\phi + 1) \frac{X(S,R) (2^\phi - 1) (\mu_k^\phi + 1)}{M_\phi + 1} \\
    &\qquad - \phi 2^\phi \log(2) + \mu_k^\phi \Big[ \log(\mu_k) (\mu_k + \phi) \\
    &\qquad - (\mu_k - 1 + 2^\phi) \log(\mu_k - 1 + 2^\phi) \Big] \Bigg] \Bigg\} \\
    &\qquad - \frac{1}{M_\phi+\left(2^\phi-1\right)\mu_k^\phi+2^\phi} \Bigg[ \lambda \big(\mu_k^\phi + 1\big) 2^\phi \log\big(\mu_k^\phi + 1\big) \\
    &\qquad - (1 - \lambda) \phi 2^\phi \mu_k^\phi \log(\mu_k) \Bigg],
  \end{split} \label{eq:delta_w_in_0_1}
\end{equation}
where
\begin{eqnarray}
X(S,R) & = & \sum_{i=1}^n \mu_i^{\phi+1} \log \mu_i \label{X_S_R_equation} \\
M_\phi & = & \sum_{i=1}^n \mu_i^{\phi+1}. \label{M_phi_equation}
\end{eqnarray}
Eq. \ref{eq:delta_w_in_0_1} can also be expressed as a linear function of $\lambda$ as
\begin{equation*}
  \Delta(\lambda) = a \lambda + b,
\end{equation*}
with 
\begin{eqnarray*}
    a &=& 2 \log\left(\frac{M_\phi + (2^\phi - 1) \mu_k^\phi + 2^\phi}{M_\phi + 1}\right) \\
      & &\qquad - \frac{1}{M_\phi + (2^\phi - 1) \mu_k^\phi + 2^\phi} \Bigg\{ 2^\phi \left[ (\mu_k^\phi + 1) \log(\mu_k^\phi + 1) + \phi \mu_k^\phi \log(\mu_k) \right] \\
      & &\qquad + 2 \Big[- (\phi + 1) \frac{X(S,R) (2^\phi - 1) \mu_k^\phi + 1}{M_\phi + 1} \\
      & &\qquad + \phi 2^\phi \log(2) - \mu_k^\phi \big[ \log(\mu_k) (\mu_k + \phi) - (\mu_k - 1 + 2^\phi) \log(\mu_k - 1 + 2^\phi) \big] \Big] \Bigg\} \\
    b &=& - \log\left(\frac{M_\phi + (2^\phi - 1) \mu_k^\phi + 2^\phi}{M_\phi + 1}\right) \\
      & &\qquad + \frac{1}{M_\phi + (2^\phi - 1) \mu_k^\phi + 2^\phi} \Bigg\{\phi 2^\phi \mu_k^\phi \log(\mu_k) - (\phi + 1) \frac{X(S,R) (2^\phi - 1) \mu_k^\phi + 1}{M_\phi + 1} \\
      & &\qquad + \phi 2^\phi \log(2) - \mu_k^\phi \left[ \log(\mu_k) (\mu_k + \phi) - (\mu_k - 1 + 2^\phi) \log(\mu_k - 1 + 2^\phi) \right] \Bigg\}.
\end{eqnarray*}
Being a relaxation of the previous class, the resulting expressions of $\Delta$ are more complex than those of the previous class, which are an in turn more complex than those of the case $\phi=0$ (Eq. \ref{eq:delta_lambda_phi_0}). See Appendix \ref{vocabulary_learning_wj_in_01} for further details on the derivation of $\Delta$. 

Notice that $X(S,R)$ (Eq. \ref{X_S_R_equation}) and $M_\phi$ (Eq. \ref{M_phi_equation}) are determined by the degrees of the forms ($\mu_i$'s). To explore the phase space with a realistic distribution of $\mu_i$'s, we assume, without any loss of generality, that the $\mu_i$'s are sorted decreasingly, i.e. $\mu_1 \geq \mu_2 \geq ... \mu_i \geq \mu_{i+1} \geq ... \mu_n$.
In addition, we assume 
\begin{enumerate}
\item
$\mu_{n} = 0$, because we are investigating the problem of linking and unlinked form with counterparts.
\item 
$\mu_{n-1} = 1$.
\item
Form degrees are continuous.
\item
The relationship between $\mu_i$ and its frequency rank is a right-truncated power-law, i.e.
\begin{equation}
  \mu_i = c i^{-\tau} 
  \label{degree_versus_rank_equation}
\end{equation}
for $1 \leq i \leq n - 1$. 
\end{enumerate} 
Appendix \ref{app:form_degrees_appendix} shows that forms then follow Zipf's rank-frequency law, i.e. 
\begin{equation*}
p(s_i) = c'i^{-\alpha}
\end{equation*}
with
\begin{eqnarray*}
\alpha & = & \tau(\phi+1) \\
c' & = & \frac{(n - 1)^\alpha}{M_\phi}.
\end{eqnarray*}

The value of $\Delta$ is determined by $\lambda$, $\phi$, $\mu_k$ and the sequence of degrees of the forms, which we have parameterized with $\alpha$ and $n$.
When $\tau = \frac{\alpha}{\phi + 1} = 0$, namely when $\alpha=0$ or when $\phi \rightarrow \infty$, we recover the class where vertex degrees do not exceed one but with just one form that is unlinked.

\begin{figure}
\centering
\includegraphics[width = \textwidth]{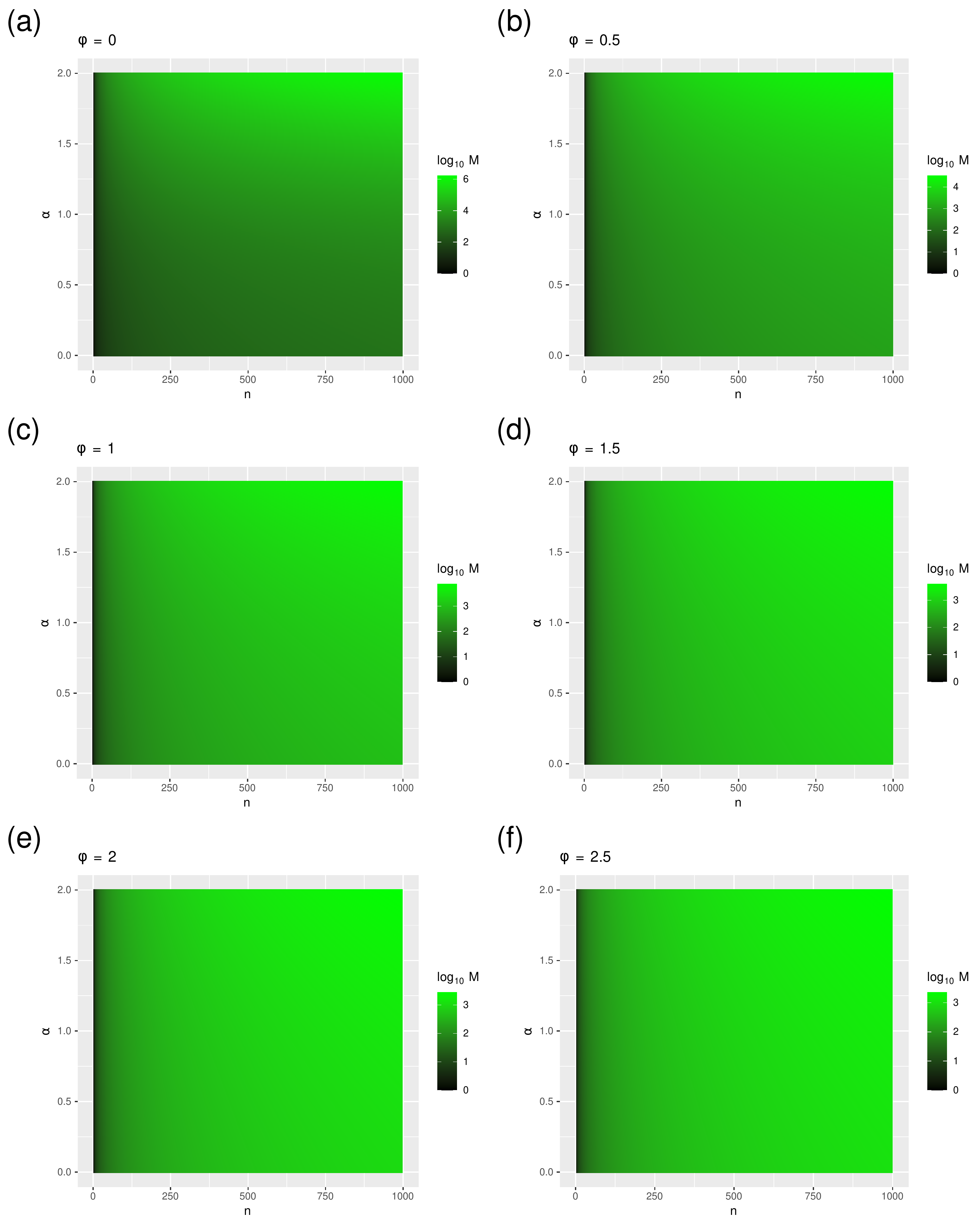}
\caption{\label{number_of_links_versus_alpha_and_n_figure} $\log_{10} M$, the logarithm of the number of links $M$, as a function of $n$ ($x$-axis) and $\alpha$ ($y$-axis) according to Eq. \ref{eq:number_of_links_alpha_and_n}. $\log_{10} M$ is used instead of $M$ to capture changes in order of magnitude of $M$. (a) $\phi = 0$, (b) $\phi = 0.5$, (c) $\phi = 1$, (d) $\phi = 1.5$, (e) $\phi = 2$ and (f) $\phi = 2.5$.
}
\end{figure}

A continuous approximation to the number of edges gives (Appendix \ref{app:form_degrees_appendix})
\begin{equation}
M = (n-1)^{\frac{\alpha}{\phi+1}} \sum_{i=1}^{n-1} i^{-\frac{\alpha}{\phi+1}}.
\label{eq:number_of_links_alpha_and_n}
\end{equation}
We aim to shed some light on the possible trajectory that children will describe on Fig. \ref{number_of_links_versus_alpha_and_n_figure} as they become older. One expects that $M$ tends to increase as children become older, due to word learning.
It is easy to see that Eq. \ref{eq:number_of_links_alpha_and_n} predicts that, if $\phi$ and $\alpha$ remain constant, $M$ is expected to increase as $n$ increases (Fig. \ref{number_of_links_versus_alpha_and_n_figure}). Besides, when $n$ remains constant, a reduction of $\alpha$ implies a reduction of $M$ when $\phi=0$ but that effect vanishes for $\phi>0$ (Fig. \ref{number_of_links_versus_alpha_and_n_figure}). Obviously, $n$ tends to increase as a child becomes older \citep{Saxton2010a_Chapter6} and thus children's trajectory will be from left to right in Fig. \ref{number_of_links_versus_alpha_and_n_figure}. As for the temporal evolution of $\alpha$, there are two possibilities. Zipf's pioneering investigations suggest that $\alpha$ remains close to 1 over time in English children \cite[Chapter IV]{Zipf1949a}. In contrast, a wider study reported a tendency of $\alpha$ to decrease over time in sufficiently old children of different languages \citep{Baixeries2012c} but the study did not determine the actual number of children where that trend was statistically significant after controlling for multiple comparisons.
Then children, as they become older, are likely to move either from left to right, keeping $\alpha$ constant, or from the left-upper corner (high $\alpha$, low $n$) to the bottom-right corner (low $\alpha$, high $n$) within each panel of Fig. \ref{number_of_links_versus_alpha_and_n_figure}. When $\phi$ is sufficiently large, the actual evolution of some children (decrease of $\alpha$ jointly with an increase of $n$) is dominated by the increase of $M$ that the growth of $n$ implies in the long run (Fig. \ref{number_of_links_versus_alpha_and_n_figure}).

When exploring the space of parameters, we must warrant that $\mu_k$ does not exceed the maximum degree that $n$, $\phi$ and $\alpha$ yield, namely $\mu_k \leq \mu_1$, where $\mu_1$ is defined according to Eq. \ref{degree_versus_rank_equation} with $i = 1$, i.e. 
\begin{eqnarray}
\mu_k & \leq & \mu_1 \nonumber \\
      & =    & c \nonumber \\
      & =    & (n-1)^\tau \nonumber \\
      & =    & (n-1)^\frac{\alpha}{\phi + 1}. \label{maximum_degree_equation}
\end{eqnarray}

%% file: results.tex

Here we will analyze $\Delta$, that takes a negative value when strategy $a$ (linking a new form to a new counterpart) is more advantageous than strategy $b$ linking a new form to an already connected counterpart), and a positive value otherwise. $|\Delta|$ indicates the strength of the bias towards strategy $a$ if $\Delta<0$; towards strategy $b$ if $\Delta > 0$. Therefore, when $\Delta <0$, the smaller the value of $\Delta$, the higher the bias for strategy $a$ whereas when $\Delta>0$, the greater the value of $\Delta$, the higher the bias for strategy $b$.
Each class of skeleta is analyzed separately, beginning by the most restrictive class. 

\subsection{Vertex degrees do not exceed one}

In this class of skeleta, corresponding to younger children, $\Delta$ depends only on $\phi$, $M$ and $\lambda$. We will explore the phase space with the help of two-dimensional heatmaps of $\Delta$ where the $x$-axis is always $\lambda$ and the $y$-axis is $M$ or $\phi$.
   
\begin{figure}
\centering
\includegraphics[width = \textwidth]{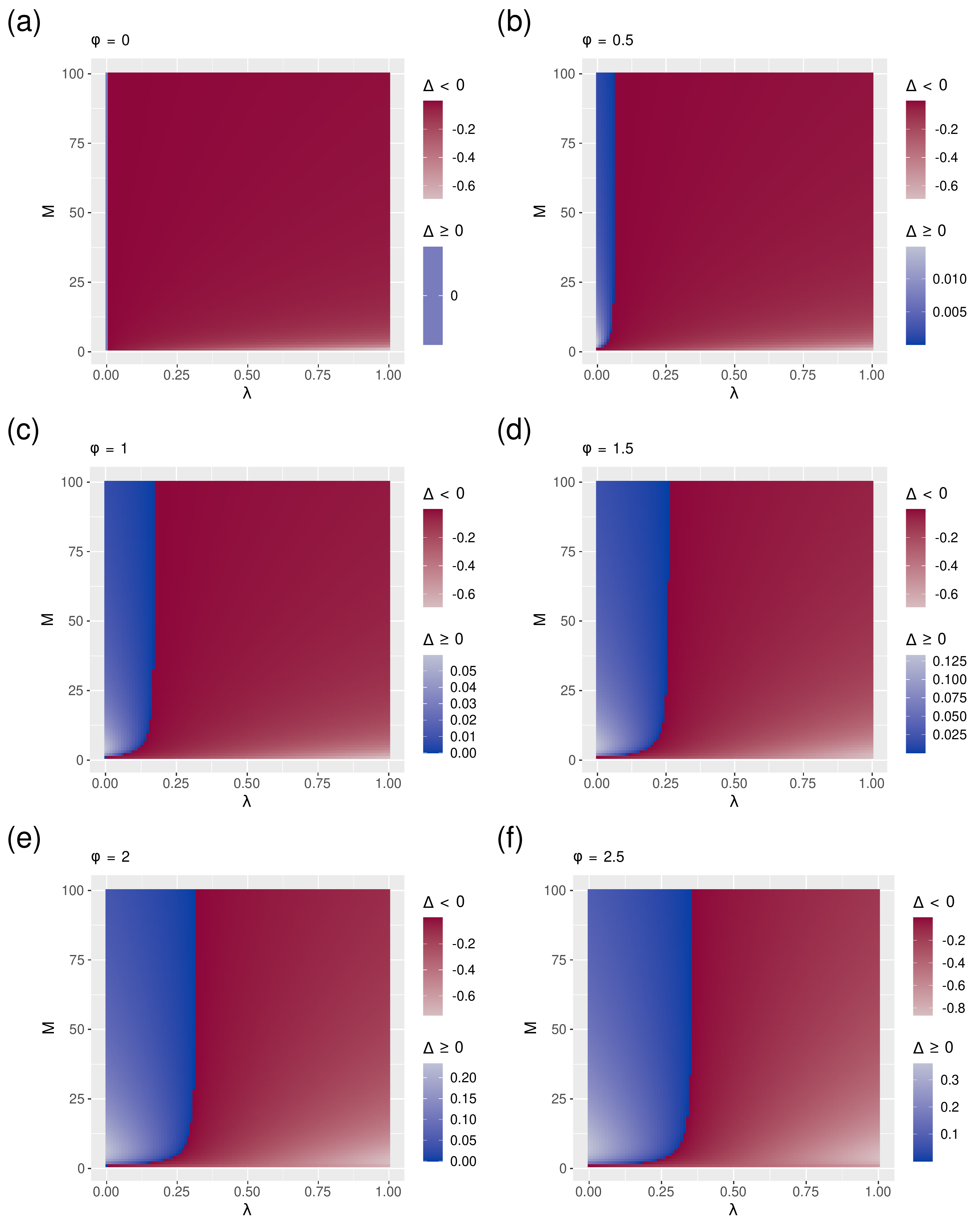}
\caption{\label{heatmap_lambda_vs_M_w_mu_in_0_1_various_phi_figure} $\Delta$, the difference between the cost of strategy $a$ and strategy $b$, as a function of $M$, the number of links and $\lambda$, the parameter that controls the balance between mutual information maximization and entropy minimization, when vertex degrees do not exceed one (Eq. \ref{eq:delta_w_mu_in_0_1}). Red indicates that strategy $a$ is more advantageous while blue indicates that $b$ is more advantageous. The lighter the red, the stronger the bias for strategy $a$. The lighter the blue, the stronger the bias for strategy $b$. (a) $\phi = 0$, (b) $\phi = 0.5$, (c) $\phi = 1$, (d) $\phi = 1.5$, (e) $\phi = 2$ and (f) $\phi = 2.5$. }
\end{figure}

\begin{figure}
\centering
\includegraphics[width = \textwidth]{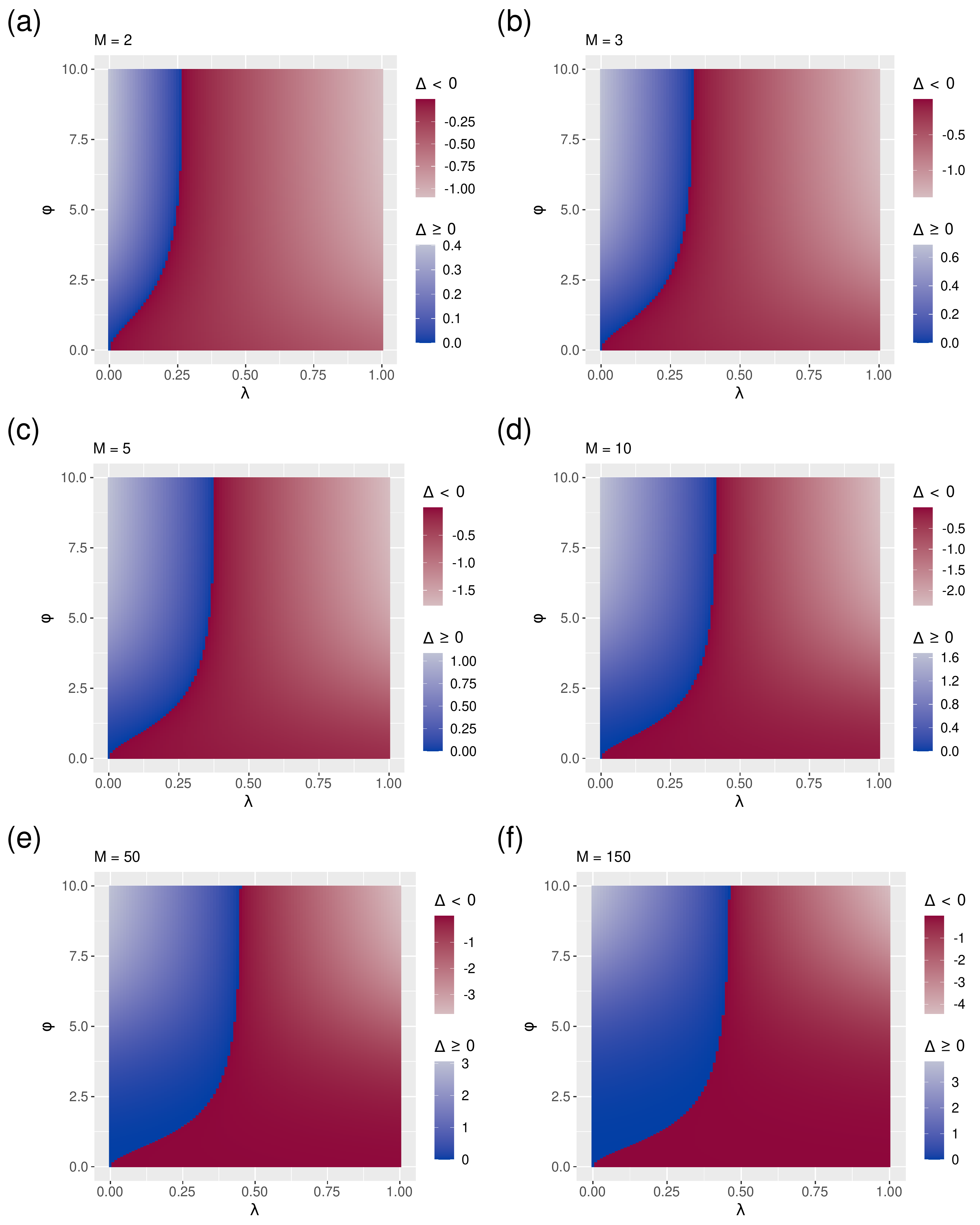}
\caption{\label{heatmap_lambda_vs_phi_w_mu_in_0_1_various_M_figure} $\Delta$, the difference between the cost of strategy $a$ and strategy $b$, as a function of $\phi$, the parameter that defines how the flesh of the model from the skeleton, and $\lambda$, the parameter that controls the balance between mutual information maximization and entropy minimization (Eq. \ref{eq:delta_w_mu_in_0_1}). Red indicates that strategy $a$ is more advantageous while blue indicates that $b$ is more advantageous. The lighter the red, the stronger the bias for strategy $a$. The lighter the blue, the stronger the bias for strategy $b$. (a) $M=2$, (b) $M=3$, (c) $M=5$, (d) $M=10$, (e) $M=50$ and (f) $M=150$. }
\end{figure}

\begin{figure}
\centering
\includegraphics[width = \textwidth]{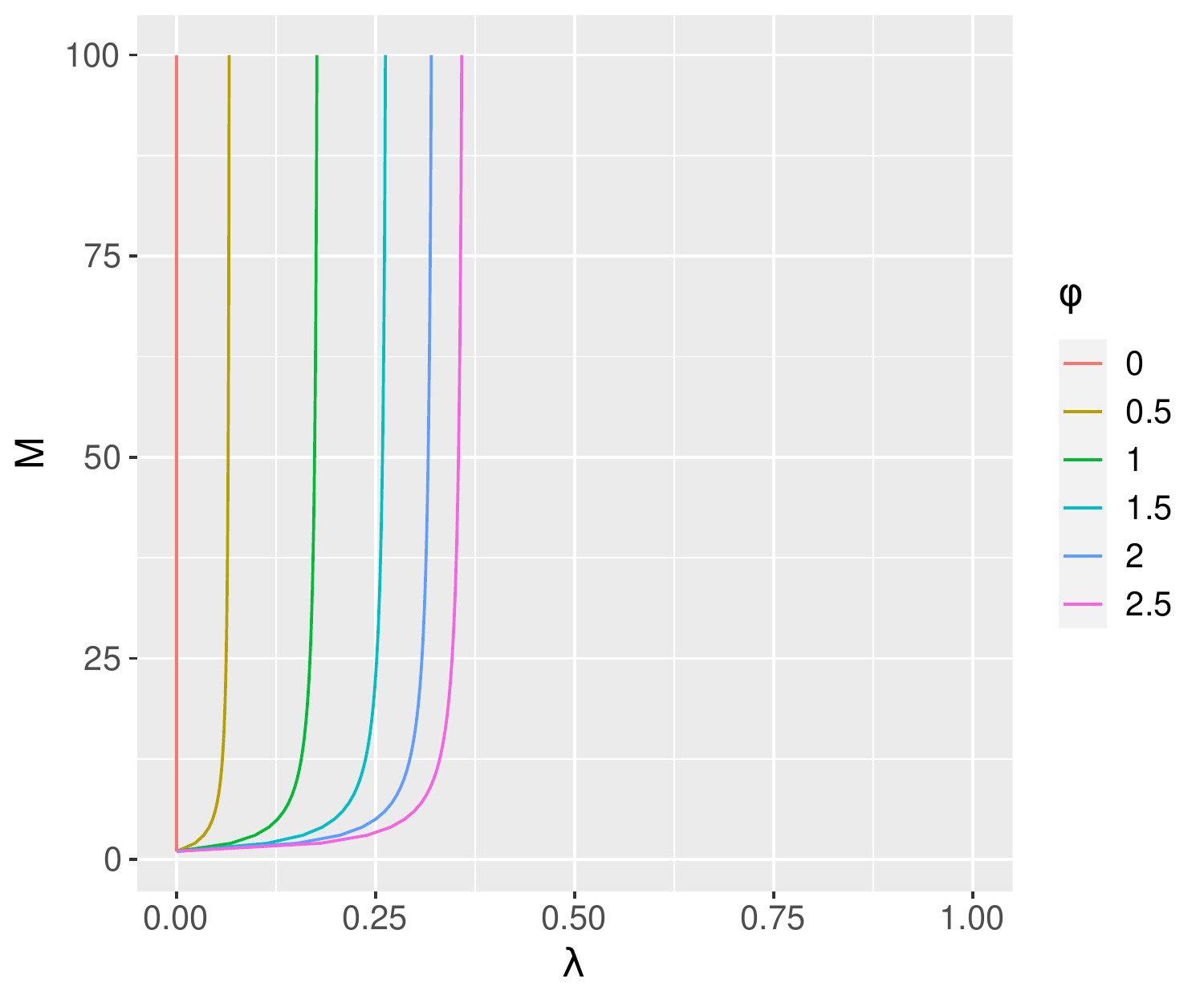}
\caption{\label{curves_delta_0_lambda_vs_M_w_mu_in_0_1_various_phi_figure} Summary of the boundaries between positive and negative values of $\Delta$ when vertex degrees do not exceed one (Fig.  \ref{heatmap_lambda_vs_M_w_mu_in_0_1_various_phi_figure}). Each curve shows the points where $\Delta = 0$ (Eq. \ref{eq:delta_w_in_0_1}) as a function of $\lambda$ and $M$ for distinct values of $\phi$.}
\end{figure}

\begin{figure}
\centering
\includegraphics[width = \textwidth]{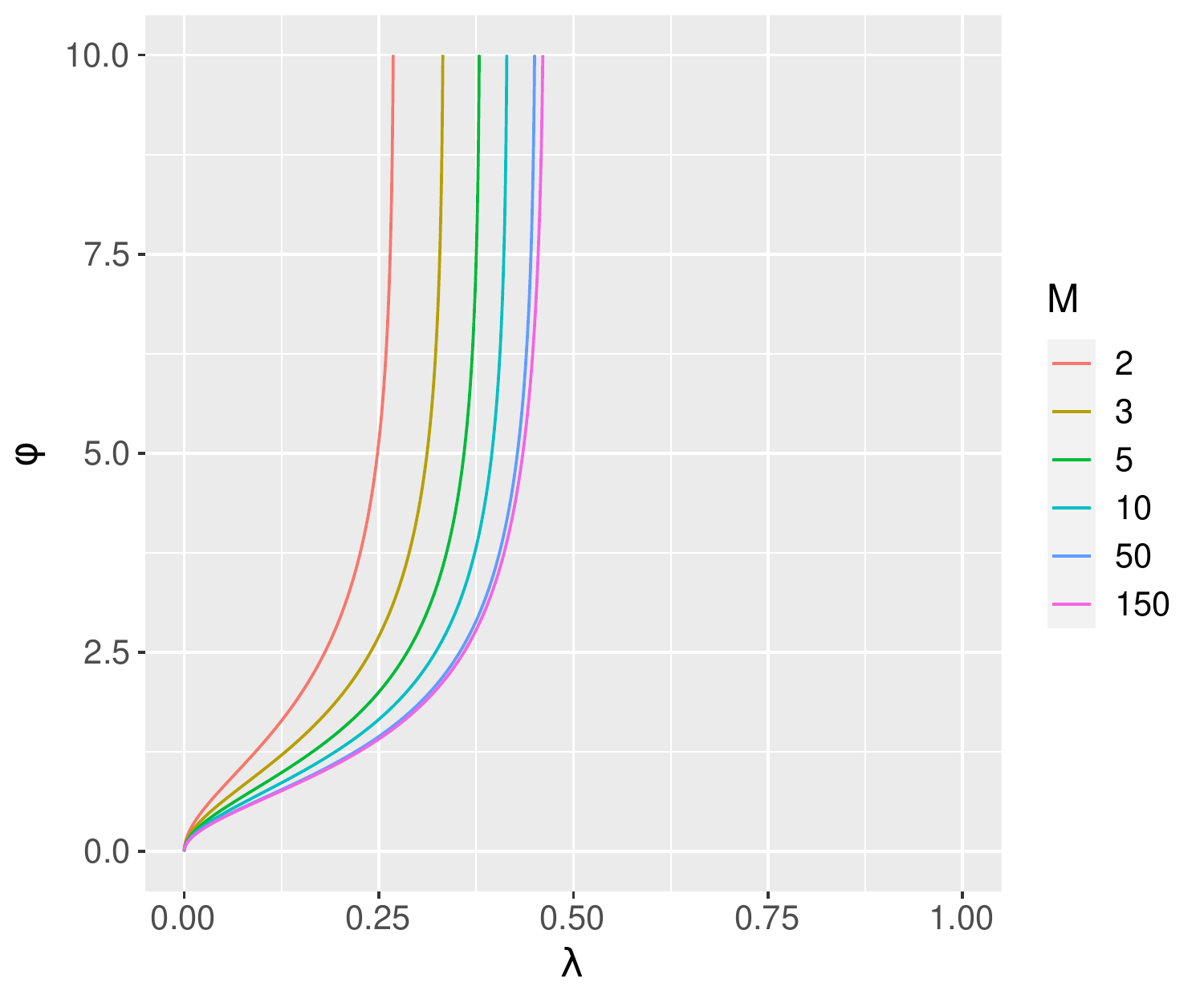}
\caption{\label{curves_delta_0_lambda_vs_phi_w_mu_in_0_1_various_M_figure} Summary of the boundaries between positive and negative values of $\Delta$ when vertex degrees do not exceed one (Fig.  \ref{heatmap_lambda_vs_phi_w_mu_in_0_1_various_M_figure}). Each curve shows the points where $\Delta = 0$ (Eq. \ref{eq:delta_w_in_0_1}) as a function of $\lambda$ and $\phi$ for distinct values of  $M$. }
\end{figure}

Figs. \ref{heatmap_lambda_vs_M_w_mu_in_0_1_various_phi_figure} and \ref{heatmap_lambda_vs_phi_w_mu_in_0_1_various_M_figure} reveal regions where strategy $a$ is more advantageous (red) and regions where $b$ is more advantageous (blue) according to $\Delta$. The extreme situation is found when $\phi=0$ where a single red region covers practically all space except for $\lambda=0$ (Fig. \ref{heatmap_lambda_vs_M_w_mu_in_0_1_various_phi_figure}, top-left) as expected from previous work \citep{Ferrer2013g} and Eq. \ref{eq:delta_lambda_phi_0}. 
Figs. \ref{curves_delta_0_lambda_vs_M_w_mu_in_0_1_various_phi_figure} and \ref{curves_delta_0_lambda_vs_phi_w_mu_in_0_1_various_M_figure} summarize these finding of regions, displaying the curve that 
defines the boundary between strategies $a$ and $b$ ($\Delta=0$).

Figs. \ref{curves_delta_0_lambda_vs_M_w_mu_in_0_1_various_phi_figure} and \ref{curves_delta_0_lambda_vs_phi_w_mu_in_0_1_various_M_figure} show that strategy $b$ is the optimal only if $\lambda$ is sufficiently low, namely when the weight of entropy minimization is sufficiently high compared to that of mutual information maximization. Fig. \ref{curves_delta_0_lambda_vs_M_w_mu_in_0_1_various_phi_figure} shows that the larger the value of $\lambda$ the larger the number of links ($M$) that is required for strategy $b$ to be optimal. Fig. \ref{curves_delta_0_lambda_vs_M_w_mu_in_0_1_various_phi_figure} also indicates that the larger the value of $\phi$, the broader the blue region where $b$ is optimal. From a symmetric perspective, Fig. \ref{curves_delta_0_lambda_vs_phi_w_mu_in_0_1_various_M_figure} shows that the larger the value of $\lambda$ the larger the value of $\phi$ that is required for strategy $b$ to be optimal and also that the larger the number of links ($M$), the broader the blue region where $b$ is optimal.


\subsection{Counterpart degrees do not exceed one}

\begin{figure}
\centering
\includegraphics[width = \textwidth]{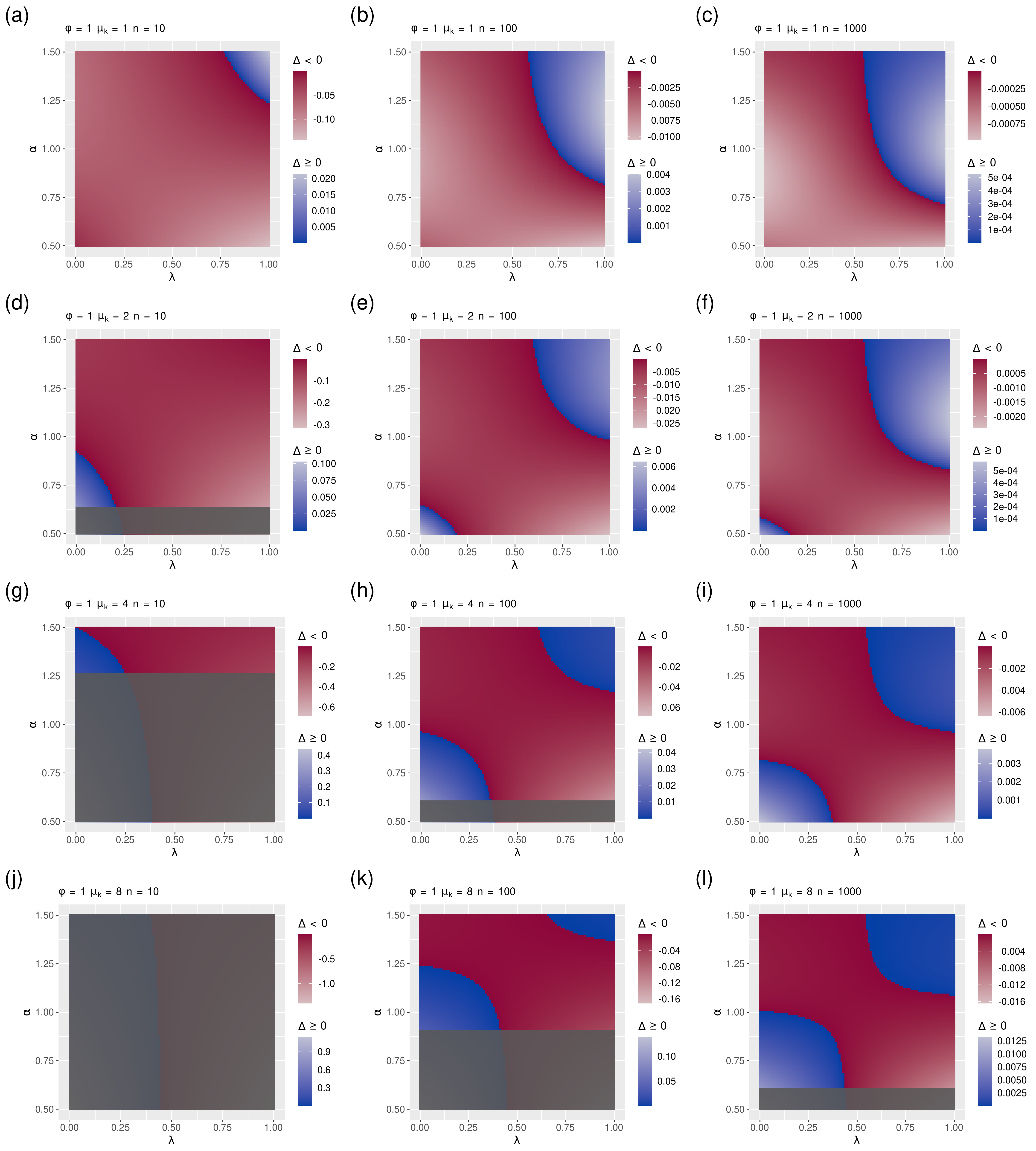}
\caption{\label{heatmap_lambda_vs_alpha_w_in_0_1_mu_continuous_phi_1_figure} 
$\Delta$, the difference between the cost of strategy $a$ and strategy $b$, as a function of $\alpha$, the exponent of the rank-frequency law, and $\lambda$, the parameter that  controls the balance between mutual information maximization and entropy minimization, when the degrees of counterparts do not exceed one (Eq. \ref{eq:delta_w_mu_in_0_1}) and $\phi=1$.  Red indicates that strategy $a$ is more advantageous while blue indicates that $b$ is more advantageous. The lighter the red, the stronger the bias for strategy $a$. The lighter the blue, the stronger the bias for strategy $b$.
Each heatmap corresponds to a distinct combination of $n$ and $\mu_k$. The heatmaps are arranged, from left to right, with $n=10,100,1000$ and, from top to bottom, with $\mu_k=1,2,4,8$. 
Gray indicates regions where $\mu_k$ exceeds the maximum degree according to other parameters (Eq. \ref{maximum_degree_equation}). (a) $\mu_k=1$ and $n=10$, (b) $\mu_k=1$ and $n=100$, (c) $\mu_k=1$ and $n=1000$, (d) $\mu_k=2$ and $n=10$, (e) $\mu_k=2$ and $n=100$, (f) $\mu_k=2$ and $n=1000$, (g) $\mu_k=4$ and $n=10$, (h) $\mu_k=4$ and $n=100$, (i) $\mu_k=4$ and $n=1000$, (j) $\mu_k=8$ and $n=10$, (k) $\mu_k=8$ and $n=100$, (l) $\mu_k=8$ and $n=1000$.
}
\end{figure}

\begin{figure}
\centering
\includegraphics[width = \textwidth]{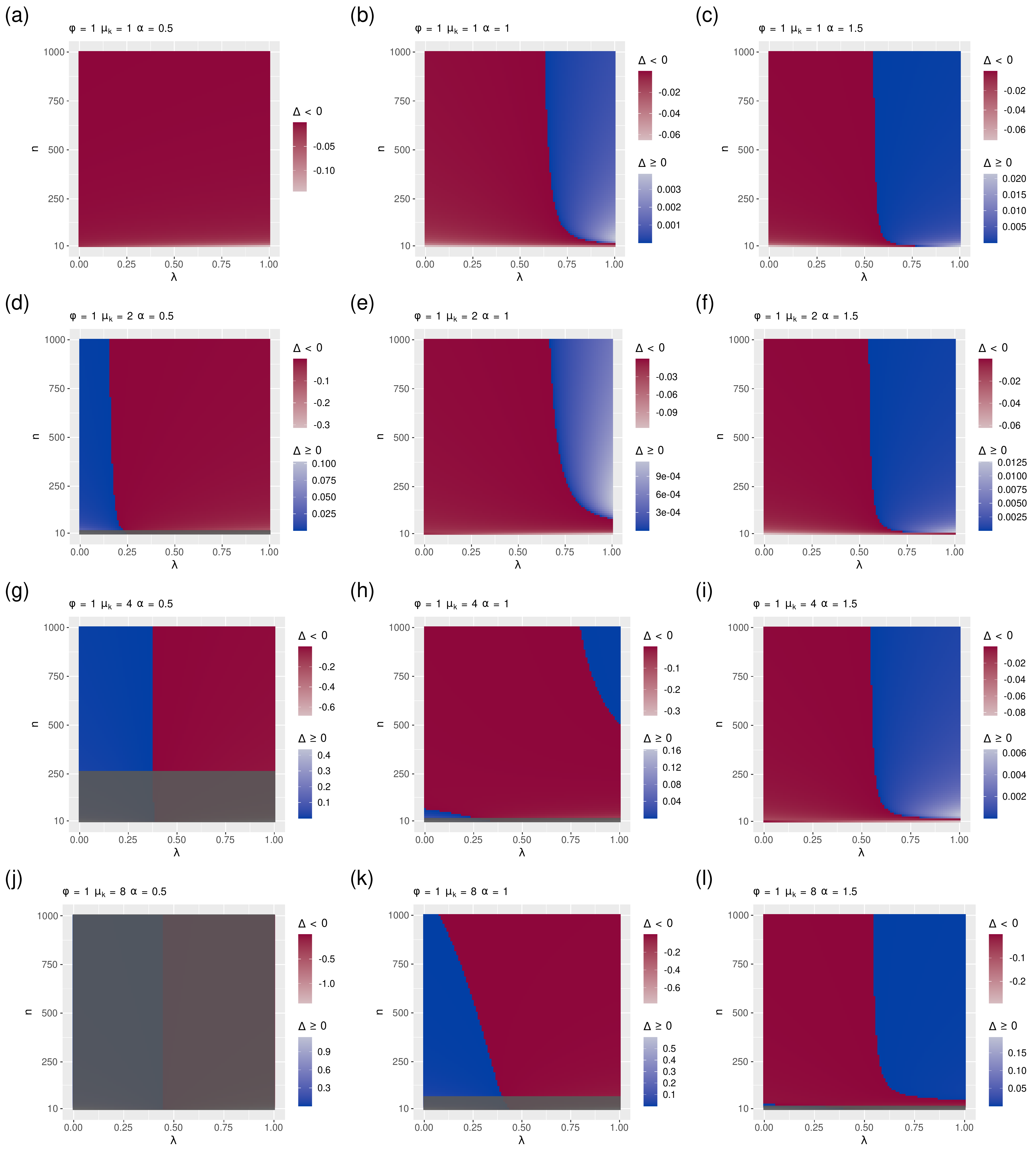}
\caption{\label{heatmap_lambda_vs_n_w_in_0_1_mu_continuous_phi_1_figure} $\Delta$, the difference between the cost of strategy $a$ and strategy $b$, as function of $n$, the number of forms, and $\lambda$, the parameter that controls the balance between mutual information maximization and entropy minimization, when the degrees of counterparts do not exceed one (Eq. \ref{eq:delta_w_mu_in_0_1}) and $\phi=1$. We are taking values of $n$ from 10 onwards (instead of one onwards) to see more clearly the light regions that are reflected on the color scales.  
  Red indicates that strategy $a$ is more advantageous while blue indicates that $b$ is more advantageous. The lighter the red, the stronger the bias for strategy $a$. The lighter the blue, the stronger the bias for strategy $b$. Each heatmap corresponds to a distinct combination of $\mu_k$ and $\alpha$. The heatmaps are arranged, from left to right, with $\alpha=0.5,1,1.5$ and, from top to bottom, with $\mu_k=1,2,4,8$.
Gray indicates regions where $\mu_k$ exceeds the maximum degree according to other parameters (Eq. \ref{maximum_degree_equation}). (a) $\mu_k=1$ and $\alpha=0.5$, (b) $\mu_k=1$ and $\alpha=1$, (c) $\mu_k=1$ and $\alpha=1.5$, (d) $\mu_k=2$ and $\alpha=0.5$, (e) $\mu_k=2$ and $\alpha=1$, (f) $\mu_k=2$ and $\alpha=1.5$, (g) $\mu_k=4$ and $\alpha=0.5$, (h) $\mu_k=4$ and $\alpha=1$, (i) $\mu_k=4$ and $\alpha=1.5$, (j) $\mu_k=8$ and $\alpha=0.5$, (k) $\mu_k=8$ and $\alpha=1$, (l) $\mu_k=8$ and $\alpha=1.5$.
}
\end{figure}

For this class of skeleta, corresponding to older children, we have assumed that word frequencies follow Zipf's rank-frequency law, namely the relationship between the probability of a form (the number of counterparts connected to each form) and its frequency rank follows a right-truncated power-law with exponent $\alpha$ (Section \ref{sec:model}). Then $\Delta$ depends only on $\alpha$ (the exponent of the right-truncated power law), $n$ (the number of forms), $\mu_k$ (the degree of the form linked to the counterpart in strategy $b$ as shown in Fig. \ref{vocabulary_learning_figure}), $\phi$ and $\lambda$.
We will explore the phase space with the help of two-dimensional heatmaps of $\Delta$ where the $x$-axis is always $\lambda$ and the $y$-axis is $\mu_k$, $\alpha$ or $n$.
While in the class where vertex degrees do not exceed one we have found only one blue region (a region where $\Delta>0$ meaning that $b$ is more advantageous), this class yields up to two distinct blue regions located in opposite corners of the heatmap while keeping always a red region as show in Figs. \ref{heatmap_lambda_vs_mu_k_w_in_0_1_mu_continuous_phi_1_figure}, \ref{heatmap_lambda_vs_alpha_w_in_0_1_mu_continuous_phi_1_figure} and \ref{heatmap_lambda_vs_n_w_in_0_1_mu_continuous_phi_1_figure} for $\phi = 1$ from different perspectives. For the sake of brevity, this section only presents heatmaps of $\Delta$ for $\phi=0$ or $\phi=1$ (see Appendix \ref{app:other_values_phi} for the remainder). A summary of exploration of the parameter space follows.  

\begin{figure}
\centering
\includegraphics[width = \textwidth]{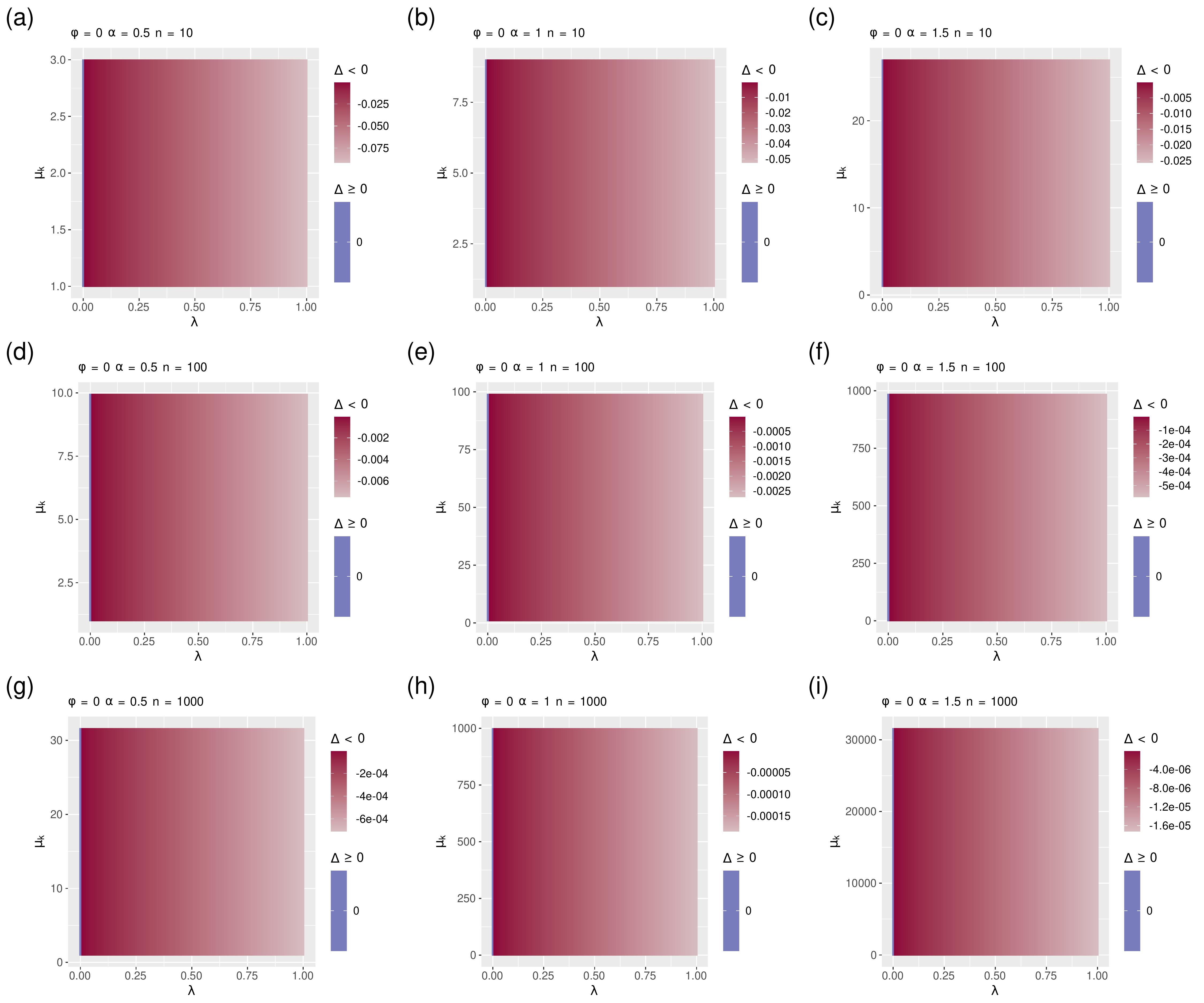}
\caption{\label{heatmap_lambda_vs_mu_k_w_in_0_1_mu_continuous_phi_0_figure} $\Delta$, the difference between the cost of strategy $a$ and strategy $b$, as a function of $\mu_k$, the degree of the form linked to the counterpart in strategy $b$ as shown in Fig. \ref{vocabulary_learning_figure}, the number of links and $\lambda$, the parameter that controls the balance between mutual information maximization and entropy minimization, when the degrees of counterparts do not exceed one (Eq. \ref{eq:delta_w_mu_in_0_1}) and $\phi=0$. Red indicates that strategy $a$ is more advantageous while blue indicates that $b$ is more advantageous. The lighter the red, the stronger the bias for strategy $a$. The lighter the blue, the stronger the bias for strategy $b$.
Each heatmap corresponds to a distinct combination of $n$ and $\alpha$. The heatmaps are arranged, from left to right, with $\alpha=0.5,1,1.5$ and, from top to bottom, with $n=10,100,1000$. (a) $\alpha=0.5$ and $n=10$, (b) $\alpha=1$ and $n=10$, (c) $\alpha=1.5$ and $n=10$, (d) $\alpha=0.5$ and $n=100$, (e) $\alpha=1$ and $n=100$, (f) $\alpha=1.5$ and $n=100$, (g) $\alpha=0.5$ and $n=1000$, (h) $\alpha=1$ and $n=1000$, (i) $\alpha=1.5$ and $n=1000$. }
\end{figure}

\begin{figure}
\centering
\includegraphics[width = \textwidth]{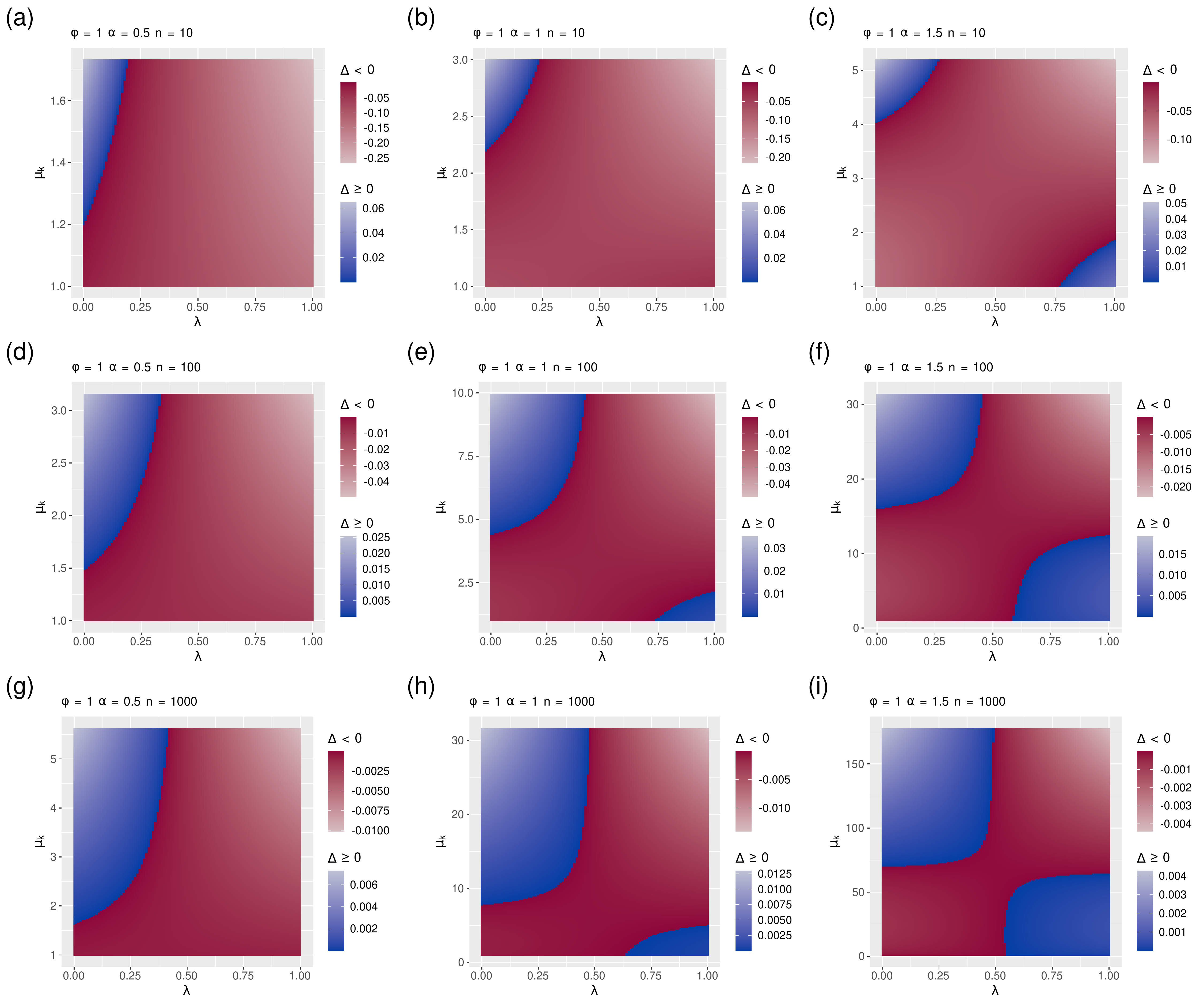}
\caption{\label{heatmap_lambda_vs_mu_k_w_in_0_1_mu_continuous_phi_1_figure} Same as in Fig. \ref{heatmap_lambda_vs_mu_k_w_in_0_1_mu_continuous_phi_0_figure} but with $\phi=1$.
}
\end{figure}

\begin{figure}
\centering
\includegraphics[width = \textwidth]{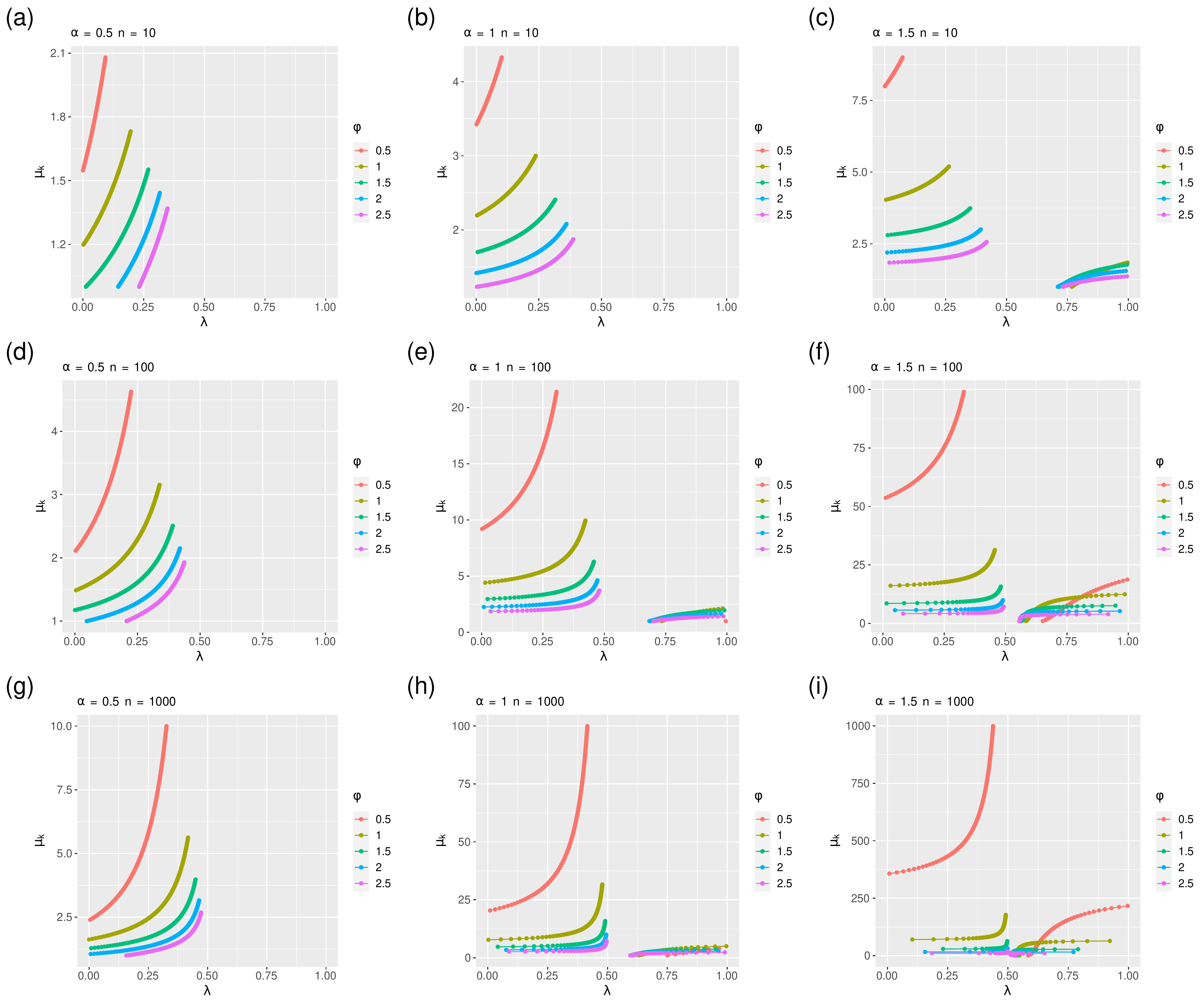}
\caption{\label{curves_delta_0_lambda_vs_mu_k_w_in_0_1_mu_continuous_various_phi_figure} Summary of the boundaries between positive and negative values of $\Delta$ when the degrees of counterparts do not exceed one (figures \ref{heatmap_lambda_vs_mu_k_w_in_0_1_mu_continuous_phi_0_figure}, \ref{heatmap_lambda_vs_mu_k_w_in_0_1_mu_continuous_phi_1_figure}, \ref{heatmap_lambda_vs_mu_k_w_in_0_1_mu_continuous_phi_0.5_figure}, \ref{heatmap_lambda_vs_mu_k_w_in_0_1_mu_continuous_phi_1.5_figure}, \ref{heatmap_lambda_vs_mu_k_w_in_0_1_mu_continuous_phi_2_figure} and \ref{heatmap_lambda_vs_mu_k_w_in_0_1_mu_continuous_phi_2.5_figure}). 
  Each curve shows the points where $\Delta = 0$ (Eq. \ref{eq:delta_w_in_0_1}) as a function of $\lambda$ and $\mu_k$ for distinct values of $\phi$. (a) $\alpha=0.5$ and $n=10$, (b) $\alpha=1$ and $n=10$, (c) $\alpha=1.5$ and $n=10$, (d) $\alpha=0.5$ and $n=100$, (e) $\alpha=1$ and $n=100$, (f) $\alpha=1.5$ and $n=100$, (g) $\alpha=0.5$ and $n=1000$, (h) $\alpha=1$ and $n=1000$, (i) $\alpha=1.5$ and $n=1000$. }
\end{figure}

\paragraph{Heatmaps of $\Delta$ as a function of $\lambda$ and $\mu_k$.}
The heatmaps of $\Delta$ for different combinations of parameters in Figs. \ref{heatmap_lambda_vs_mu_k_w_in_0_1_mu_continuous_phi_0_figure}, \ref{heatmap_lambda_vs_mu_k_w_in_0_1_mu_continuous_phi_1_figure}, \ref{heatmap_lambda_vs_mu_k_w_in_0_1_mu_continuous_phi_0.5_figure}, \ref{heatmap_lambda_vs_mu_k_w_in_0_1_mu_continuous_phi_1.5_figure}, \ref{heatmap_lambda_vs_mu_k_w_in_0_1_mu_continuous_phi_2_figure} and \ref{heatmap_lambda_vs_mu_k_w_in_0_1_mu_continuous_phi_2.5_figure} are summarized in Fig. \ref{curves_delta_0_lambda_vs_mu_k_w_in_0_1_mu_continuous_various_phi_figure}, showing the frontiers between regions where $\Delta=0$. Notice how, for $\phi=0$, strategy $a$ is optimal for all values of $\lambda > 0$, as one would expect from Eq. \ref{eq:delta_lambda_phi_0}. The remainder of the figures show how the shape of the two areas changes with each of the parameters. For small $n$ and $\alpha$, a single blue region indicates that strategy $b$ is more advantageous than $a$ when $\lambda$ is closer to 0 and $\mu_k$ is higher. For higher $n$ or $\alpha$ an additional blue region appears indicating that strategy $b$ is also optimal for high values of $\lambda$ and low values of $\mu_k$.

\begin{figure}
\centering
\includegraphics[width = \textwidth]{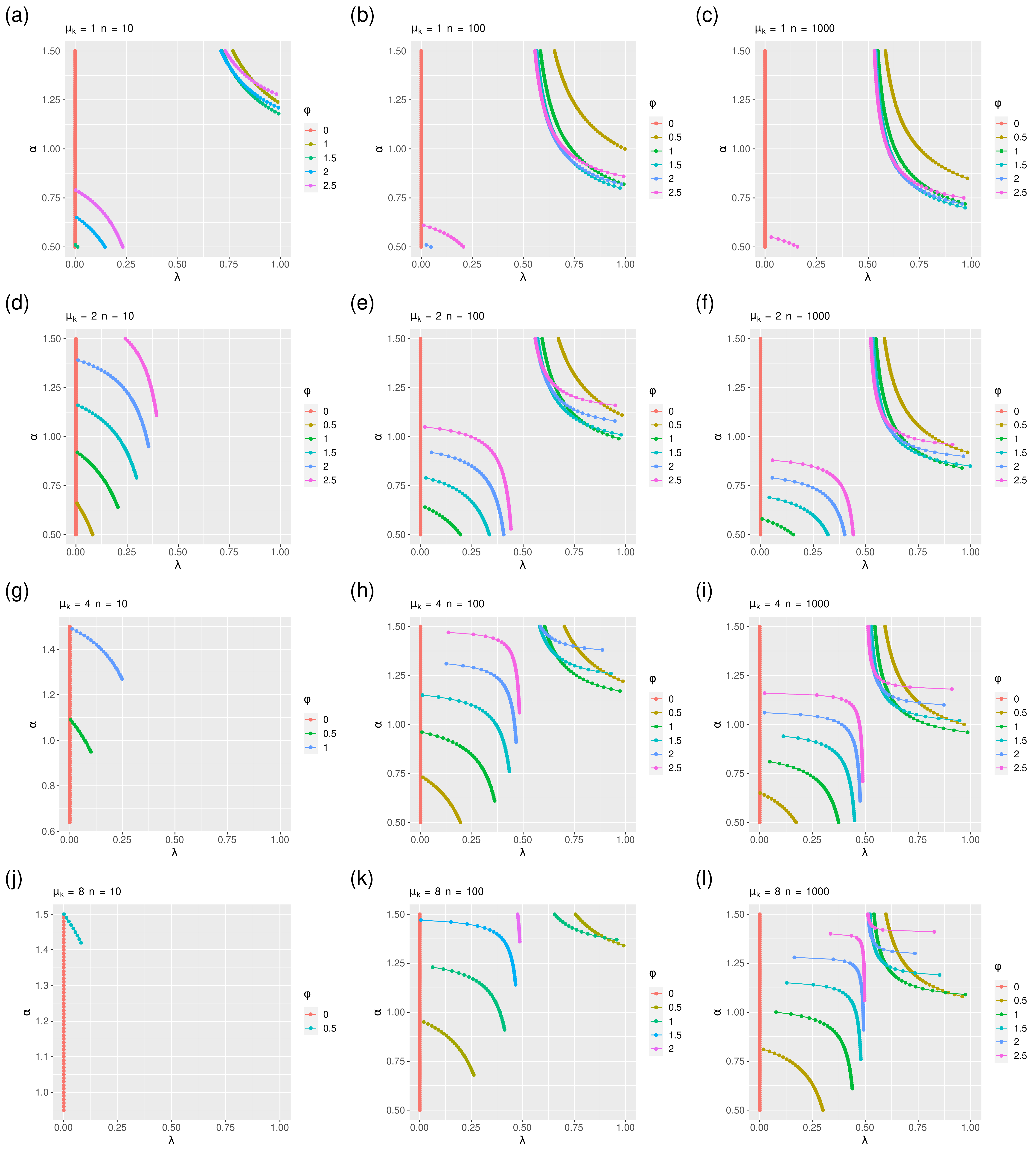}
\caption{\label{curves_delta_0_lambda_vs_alpha_w_in_0_1_mu_continuous_various_phi_figure} 
Summary of the boundaries between positive and negative values of $\Delta$ when the degrees of counterparts do not exceed one (Figs. 
\ref{heatmap_lambda_vs_alpha_w_in_0_1_mu_continuous_phi_1_figure}, \ref{heatmap_lambda_vs_alpha_w_in_0_1_mu_continuous_phi_0.5_figure}, \ref{heatmap_lambda_vs_alpha_w_in_0_1_mu_continuous_phi_1.5_figure}, \ref{heatmap_lambda_vs_alpha_w_in_0_1_mu_continuous_phi_2_figure} and  \ref{heatmap_lambda_vs_alpha_w_in_0_1_mu_continuous_phi_2.5_figure}). 
Each curve shows the points where $\Delta = 0$ (Eq. \ref{eq:delta_w_in_0_1}) as a function of $\lambda$ and $\alpha$ for distinct values of $\phi$.
Points are restricted to combinations of parameters where $\mu_k$ does not exceed the maximum (Eq. \ref{maximum_degree_equation}).
Each distinct heatmap corresponds to a distinct combination of $\mu_k$ and $n$. (a) $\mu_k=1$ and $n=10$, (b) $\mu_k=1$ and $n=100$, (c) $\mu_k=1$ and $n=1000$, (d) $\mu_k=2$ and $n=10$, (e) $\mu_k=2$ and $n=100$, (f) $\mu_k=2$ and $n=1000$, (g) $\mu_k=4$ and $n=10$, (h) $\mu_k=4$ and $n=100$, (i) $\mu_k=4$ and $n=1000$, (j) $\mu_k=8$ and $n=10$, (k) $\mu_k=8$ and $n=100$, (l) $\mu_k=8$ and $n=1000$.}
\end{figure}

\paragraph{Heatmaps of $\Delta$ as a function of $\lambda$ and $\alpha$.}
The heatmaps of $\Delta$ for different combinations of parameters in Figs. 
\ref{heatmap_lambda_vs_alpha_w_in_0_1_mu_continuous_phi_1_figure},
\ref{heatmap_lambda_vs_alpha_w_in_0_1_mu_continuous_phi_0.5_figure}, \ref{heatmap_lambda_vs_alpha_w_in_0_1_mu_continuous_phi_1.5_figure}, \ref{heatmap_lambda_vs_alpha_w_in_0_1_mu_continuous_phi_2_figure} and \ref{heatmap_lambda_vs_alpha_w_in_0_1_mu_continuous_phi_2.5_figure} are summarized in Fig. \ref{curves_delta_0_lambda_vs_alpha_w_in_0_1_mu_continuous_various_phi_figure}, showing the frontiers between regions. 
There is a single region where strategy $b$ is optimal for small values of $\mu_k$ and $\phi$, but for larger values a second blue region appears. 

\begin{figure}
\centering
\includegraphics[width = \textwidth]{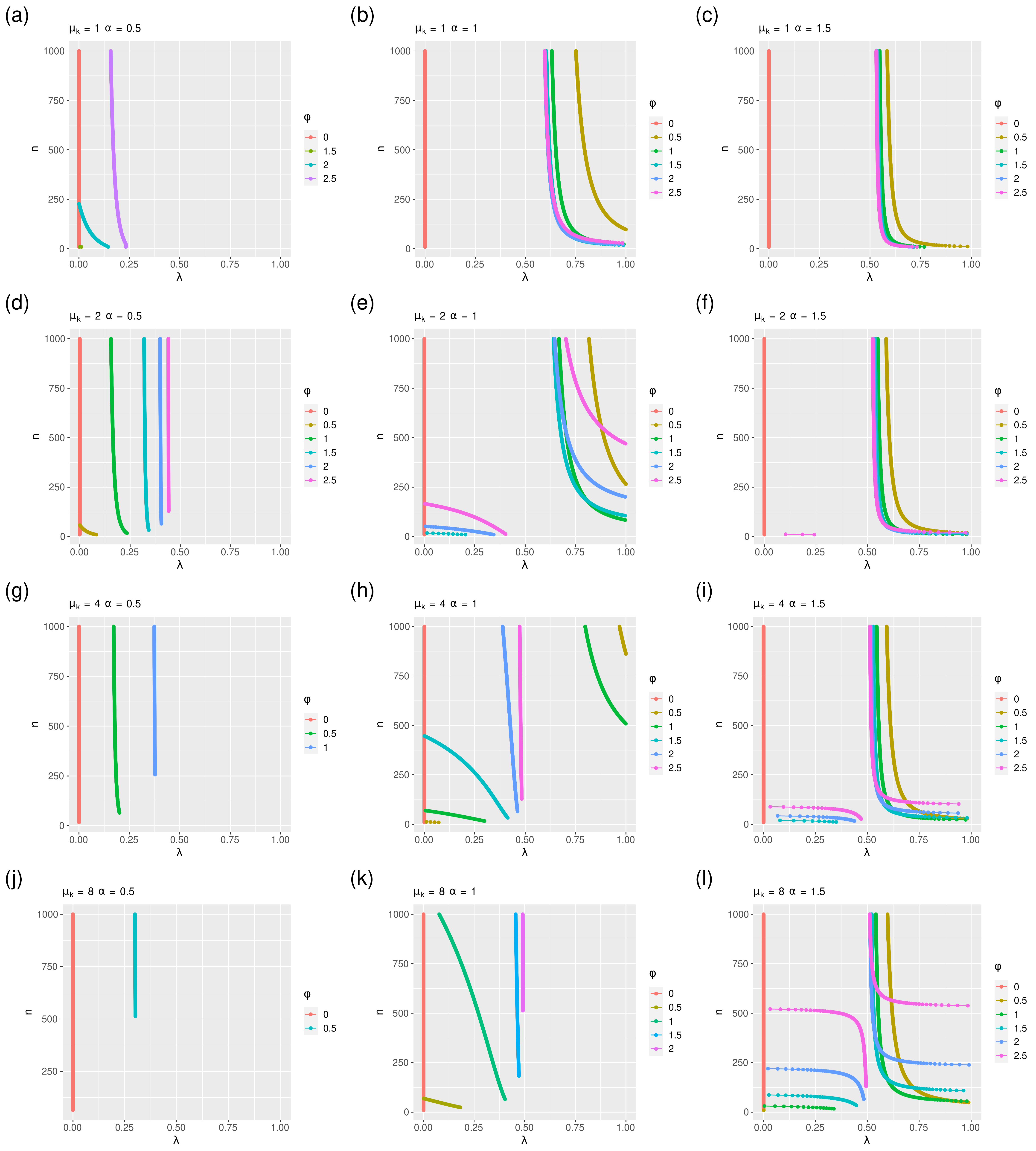}
\caption{\label{curves_delta_0_lambda_vs_n_w_in_0_1_mu_continuous_various_phi_figure} 
Summary of the boundaries between positive and negative values of $\Delta$ when the degrees of counterparts do not exceed one (Figs. \ref{heatmap_lambda_vs_n_w_in_0_1_mu_continuous_phi_1_figure}, \ref{heatmap_lambda_vs_n_w_in_0_1_mu_continuous_phi_0.5_figure}, \ref{heatmap_lambda_vs_n_w_in_0_1_mu_continuous_phi_1.5_figure}, \ref{heatmap_lambda_vs_n_w_in_0_1_mu_continuous_phi_2_figure} and \ref{heatmap_lambda_vs_n_w_in_0_1_mu_continuous_phi_2.5_figure}). 
Each curve shows the points where $\Delta = 0$ (Eq. \ref{eq:delta_w_in_0_1}) as a function of $\lambda$ and $n$ for distinct values of $\phi$.
Points are restricted to combinations of parameters where $\mu_k$ does not exceed the maximum (Eq. \ref{maximum_degree_equation}).
Each distinct heatmap corresponds to a distinct combination of $\mu_k$ and $\alpha$. (a) $\mu_k=1$ and $\alpha=0.5$, (b) $\mu_k=1$ and $\alpha=1$, (c) $\mu_k=1$ and $\alpha=1.5$, (d) $\mu_k=2$ and $\alpha=0.5$, (e) $\mu_k=2$ and $\alpha=1$, (f) $\mu_k=2$ and $\alpha=1.5$, (g) $\mu_k=4$ and $\alpha=0.5$, (h) $\mu_k=4$ and $\alpha=1$, (i) $\mu_k=4$ and $\alpha=1.5$, (j) $\mu_k=8$ and $\alpha=0.5$, (k) $\mu_k=8$ and $\alpha=1$, (l) $\mu_k=8$ and $\alpha=1.5$.
}
\end{figure}

\paragraph{Heatmaps of $\Delta$ as a function of $\lambda$ and $n$.}
The heatmaps of $\Delta$ for different combinations of parameters in Figs. 
\ref{heatmap_lambda_vs_n_w_in_0_1_mu_continuous_phi_1_figure},
\ref{heatmap_lambda_vs_n_w_in_0_1_mu_continuous_phi_0.5_figure}, \ref{heatmap_lambda_vs_n_w_in_0_1_mu_continuous_phi_1.5_figure}, \ref{heatmap_lambda_vs_n_w_in_0_1_mu_continuous_phi_2_figure} and \ref{heatmap_lambda_vs_n_w_in_0_1_mu_continuous_phi_2.5_figure} are summarized in Fig. \ref{curves_delta_0_lambda_vs_n_w_in_0_1_mu_continuous_various_phi_figure}. Again, one or two blue regions appear depending on the combination of parameters. 


See Appendix \ref{app::discrete_degrees} for the impact of using discrete form degrees on the results presented in this section.

%% file: discussion.tex
\subsection{Vocabulary learning}

\label{discussion_on_vocabulary_learning_subsection}

In previous research with $\phi = 0$, we predicted that the vocabulary learning bias (strategy $a$) would be present provided that mutual information minimization is not disabled ($\lambda > 0$) \citep{Ferrer2013g} as show in Eq. \ref{eq:delta_lambda_phi_0}. However, the ``decision'' on whether assigning a new label to a linked or to an unlinked object is influenced by the age of  a child and his/her degree of polylingualism.
As for the effect of the latter, polylingual children tend to pick familiar objects more often than monolingual children, violating mutual exclusivity. This has been found for younger children below two years of age (17-22 months old in one study, 17-18 in another) \citep{Houston-Price2010a,Byers-Heinlein2013a}. 
From three years onward, the difference between polylinguals and monolinguals either vanishes, namely both violate mutual exclusivity similarly \citep{Nicoladis2020a,Frank2002a}, or polylingual children are still more willing to accept lexical overlap \citep{Kalashnikova2015a}. One possible explanation for this phenomenon is the lexicon structure hypothesis \citep{Byers-Heinlein2013a}, which suggests that children that already have many multiple-word-to-single-object mappings may be more willing to suspend mutual exclusivity.

As for the effect of age on monolingual children, the so-called mutual exclusivity bias has been shown to appear at an early age and, as time goes on, it is more easily suspended. 
Starting at 17 months old, children tend to look at a novel object rather than a familiar one when presented with a new word while 16-month-olds do not show a preference \citep{Halberda2003a}. 
Interestingly, in the same study, 14-month-olds systematically look at a familiar object instead of a newer one. 
Reliance on mutual exclusivity is shown to improve between 18 and 30 months \citep{Bion2013a}. Starting at least at 24 months of age, children may suspend mutual exclusivity to learn a second label for an object \citep{Liittschwager1994a}. In a more recent study, it has been shown that three year old children will suspend mutual exclusivity if there are enough social cues present \citep{Yildiz2020a}. Four to five year old children continue to apply mutual exclusivity to learn new words but are able to apply it flexibly, suspending it when given appropriate contextual information \citep{Kalashnikova2016a} in order to associate multiple labels to the same familiar object. As seen before, at 3 years of age both monolingual and polylingual children have similar willingness to suspend mutual exclusivity \citep{Nicoladis2020a,Frank2002a}, although polylinguals may still have a greater tendency to accept multiple labels for the same object \citep{Kalashnikova2015a}.

Here we have made an important contribution with respect to the precursor of the current model \citep{Ferrer2013g}: we have shown that the bias is not theoretically inevitable (even when $\lambda >0$) according a more realistic model. In a more complex setting, research on deep neural networks has shed light on the architectures, learning biases and pragmatic strategies that are required for the vocabulary learning bias to emerge \cite[e.g.][]{Gandhi2019a,Gulordava2020a}. 
In section \ref{sec:results}, we have discovered regions of the space of parameters where strategy $a$ is not advantageous for two classes of skeleta. 
In the restrictive class, where one where vertex degrees do no exceed one, as expected in the earliest stages of vocabulary learning in children, we have unveiled the existence of a region of the phase space where strategy $a$ is not advantageous (Figs. \ref{curves_delta_0_lambda_vs_M_w_mu_in_0_1_various_phi_figure} and \ref{heatmap_lambda_vs_phi_w_mu_in_0_1_various_M_figure}). In the broader class of skeleta where the degree of counterparts does not exceed one we have found up to two distinct regions where $a$ is not advantageous (Figs. \ref{curves_delta_0_lambda_vs_mu_k_w_in_0_1_mu_continuous_various_phi_figure} and \ref{curves_delta_0_lambda_vs_alpha_w_in_0_1_mu_continuous_various_phi_figure}). 

Crucially, our model predicts that the bias should be lost in older children. The argument is as follows. Suppose a child that has not learned a word yet. Then his skeleton belongs to the class where vertex degrees do not exceed one. Then suppose that the child learns a new word. It could be that he/she learns it following strategy $a$ or $b$. If he applies $b$ then the bias is gone at least for this word. Let us suppose that the child learns words adhering to strategy $a$ for as long as possible. By doing this, he/she will increasing the number of links ($M$) of the skeleton keeping as invariant a one-to-one mapping between words and meanings (Figs. \ref{bipartite_graph_figure} (c) and \ref{real_bipartite_graph_figure} (d)), which satisfies that vertex degrees do not exceed one. Then Figs. \ref{curves_delta_0_lambda_vs_M_w_mu_in_0_1_various_phi_figure} and \ref{curves_delta_0_lambda_vs_phi_w_mu_in_0_1_various_M_figure} predict that the longer the time strategy $a$ is kept (when $\phi>0$) the larger the region of the phase space where $a$ is not advantageous. Namely, as times goes on, it will become increasingly more difficult to keep $a$ as the best option. Then it is not surprising that the bias weakens either in older children \citep[e.g.,][]{Yildiz2020a, Kalashnikova2016a}, as they are expected to have more links (larger $M$) because of their continued accretion of new words \citep{Saxton2010a_Chapter6}, or in polylinguals \citep[e.g.,][]{Nicoladis2000a, Greene2013a}, where the mapping of words into meanings combining all their languages, is expected to yield more links than in monolinguals. Polylinguals make use of code-mixing to compensate for lexical gaps, as reported for from one-year-olds onward \citep{Nicoladis2000a} as well as in older children (five year olds) \citep{Greene2013a}. As a result, the bipartite skeleton of a polylingual integrates the words and association in all the languages spoken and thus polylinguals are expected to have a larger value of $M$. Children who know more translation equivalents (words from different languages but with same meaning), adhere to mutual exclusivity less than other children \citep{Byers-Heinlein2013a}. Therefore, our theoretical framework provides an explanation for the lexicon structure hypothesis \citep{Byers-Heinlein2013a}, but shedding light on the possible origin of the mechanism, that is not the fact that there are already synonyms but rather the large number of links (Fig. \ref{curves_delta_0_lambda_vs_phi_w_mu_in_0_1_various_M_figure}) as well as the capacity of words of higher degree to attract more meanings, a consequence of Eq. \ref{joint_probability_equation} with $\phi>0$ in the vocabulary learning process (Fig. \ref{vocabulary_learning_figure}). Recall the stark contrast between Fig. \ref{heatmap_lambda_vs_mu_k_w_in_0_1_mu_continuous_phi_1_figure} for $\phi = 1$ and the Fig. \ref{heatmap_lambda_vs_mu_k_w_in_0_1_mu_continuous_phi_0_figure} with $\phi=0$, where such attraction effect is missing. Our models offer a transparent theoretical tool to understand the failure of deep neural networks to reproduce the vocabulary learning bias \citep{Gandhi2019a}: in its simpler form (vertex degrees do not exceed one), whether it is due to an excessive $\phi$ (Fig. \ref{curves_delta_0_lambda_vs_M_w_mu_in_0_1_various_phi_figure}) or an excessive $M$ (Fig. \ref{curves_delta_0_lambda_vs_phi_w_mu_in_0_1_various_M_figure}).

{We have focused on the loss of the bias in older children. However, there is evidence that the bias is missing initially in children, by the age of 14 months \citep{Halberda2003a}. We speculate that this could be related to very young children having lower values of $\lambda$ or larger values of $\phi$ as suggested by Figs. \ref{curves_delta_0_lambda_vs_M_w_mu_in_0_1_various_phi_figure} and \ref{heatmap_lambda_vs_phi_w_mu_in_0_1_various_M_figure}. This issue should be the subject of future research. Methods to estimate $\phi$ and $\lambda$ in real speakers should be investigated.

Now we turn our attention to skeleta where only the degree of the counterparts does not exceed one, that we believe to be more appropriate for older children.
Whereas $\phi$, $\lambda$ and $M$ sufficed for the exploration of the phase space when vertex degrees do not exceed one, the exploration of that kind of skeleta involved many parameters: $\phi$, $\lambda$, $n$, $\mu_k$ and $\alpha$. 
The more general class exhibits behaviors that we have already seen in the more restrictive class. While an increase in $M$ implies a widening of the region where $a$ is not advantageous in the more restrictive class, the more general class experiences an increase of $M$ when $n$ is increased but $\alpha$ and $\phi$ remain constant (Section \ref{counterpart_degrees_do_not_exceed_one_section}). Consistently with the more restrictive class, such increase of $M$ leads to a growth of the regions where $a$ is not advantageous as it can be seen in Figs. \ref{heatmap_lambda_vs_mu_k_w_in_0_1_mu_continuous_phi_0.5_figure}, \ref{heatmap_lambda_vs_mu_k_w_in_0_1_mu_continuous_phi_1_figure}, \ref{heatmap_lambda_vs_mu_k_w_in_0_1_mu_continuous_phi_1.5_figure}, \ref{heatmap_lambda_vs_mu_k_w_in_0_1_mu_continuous_phi_2_figure} and \ref{heatmap_lambda_vs_mu_k_w_in_0_1_mu_continuous_phi_2.5_figure} when selecting a column (thus fixing $\alpha$ and $\phi$) and moving from the top to the bottom increasing $n$. The challenge is that $\alpha$ may not remain constant in real children as they become older and how to involve the remainder of the parameters in the argument.
In fact, some of these parameters are known to be correlated with child's age:
\begin{itemize}
\item
$n$ tends to increase over time in children, as children are learning new words over time \citep{Saxton2010a_Chapter6}. We assume that the loss of words can be neglected in children.
\item $M$ tends to increase over time in children. In this class of skeleta, the growth of $M$ has two sources: the learning of new words as well as the learning of new meanings for existing words. We assume that the loss of connections can be neglected in children. 
\item
The ambiguity of the words that children learn over time tends to increase over time \citep{Casas2019a}. This does not imply that children are learning all the meanings of the word according to some online dictionary but rather than as times go on, children are able to handle words that have more meanings according to adult standards. 
\item 
$\alpha$ remains stable over time or tends to decrease over time in children depending on the individual \citep[Chapter IV]{Baixeries2012c,Zipf1949a}.
\end{itemize}
For other parameters, we can just speculate on their evolution with child's age. The growth of $M$ and the increase in the learning of ambiguous words over time leads to expect that the maximum value of $\mu_k$ will be larger in older children. It is hard to tell if older children will have a chance to encounter larger values of $\mu_k$. We do not know the value of $\lambda$ in real language but the higher diversity of vocabulary in older children and adults \citep{Baixeries2012c} suggests that $\lambda$ may tend to increase over time, because the lower the value of $\lambda$, the higher the pressure to minimize the entropy of words (Eq. \ref{cost_equation}), namely the higher the force towards unification in Zipf's view \citep{Zipf1949a}. We do not know the real value of $\phi$ but a reasonable choice for adult language is $\phi = 1$ \citep{Ferrer2017b}. 

Given the complexity of the space of parameters in the more general class of skeleta where only the degrees of counterparts cannot exceed one, we cannot make predictions that are as strong as those stemming from the class where vertex degrees cannot exceed one. However, we wish to make some remarks suggesting that a weakening of the vocabulary learning bias is also expected in older children for this class (provided that $\phi>0$). The combination of increasing $n$ and a value of $\alpha$ that is stable over time suggests a weakening of the strategy $a$ over time from different perspectives}
\begin{itemize}
\item 
Children evolve on a column of panels (constant $\alpha$) of the matrix of panels in  
Figs. \ref{heatmap_lambda_vs_mu_k_w_in_0_1_mu_continuous_phi_0.5_figure}, \ref{heatmap_lambda_vs_mu_k_w_in_0_1_mu_continuous_phi_1_figure}, \ref{heatmap_lambda_vs_mu_k_w_in_0_1_mu_continuous_phi_1.5_figure}, \ref{heatmap_lambda_vs_mu_k_w_in_0_1_mu_continuous_phi_2_figure} and \ref{heatmap_lambda_vs_mu_k_w_in_0_1_mu_continuous_phi_2.5_figure}, moving from top (low $n$) to the bottom (large $n$). 
That trajectory implies an increase of the size of the blue region, where strategy $a$ is not advantageous.
\item
We do not know the temporal evolution of $\mu_k$ but once $\mu_k$ is fixed, namely a row of panels is selected in 
Figs. \ref{heatmap_lambda_vs_alpha_w_in_0_1_mu_continuous_phi_0.5_figure}, \ref{heatmap_lambda_vs_alpha_w_in_0_1_mu_continuous_phi_1_figure}, \ref{heatmap_lambda_vs_alpha_w_in_0_1_mu_continuous_phi_1.5_figure}, \ref{heatmap_lambda_vs_alpha_w_in_0_1_mu_continuous_phi_2_figure} and \ref{heatmap_lambda_vs_alpha_w_in_0_1_mu_continuous_phi_2.5_figure}, children evolve from left (lower $n$) to right (higher $n$),
which implies an increase of the size of the blue region where strategy $a$ is not advantageous as children become older.
\item
Within each panel in Figs. \ref{heatmap_lambda_vs_n_w_in_0_1_mu_continuous_phi_0.5_figure}, \ref{heatmap_lambda_vs_n_w_in_0_1_mu_continuous_phi_1_figure}, \ref{heatmap_lambda_vs_n_w_in_0_1_mu_continuous_phi_1.5_figure}, \ref{heatmap_lambda_vs_n_w_in_0_1_mu_continuous_phi_2_figure} and \ref{heatmap_lambda_vs_n_w_in_0_1_mu_continuous_phi_2.5_figure},
an increase of $n$, as a results of vocabulary learning over time, implies a widening of the blue region.  
\end{itemize}
In the preceding analysis we have assumed that $\alpha$ remains stable over time. We wish to speculate on the combination of increasing $n$ and decreasing $\alpha$ as time goes on in certain children. In that case, children would evolve close to the diagonal of the matrix of panels, starting from the right-upper corner (low $n$, high $\alpha$, panel (c)) towards the lower-left corner (high $n$, low $\alpha$, panel (g)) in 
Figs. \ref{heatmap_lambda_vs_mu_k_w_in_0_1_mu_continuous_phi_0.5_figure}, \ref{heatmap_lambda_vs_mu_k_w_in_0_1_mu_continuous_phi_1_figure}, \ref{heatmap_lambda_vs_mu_k_w_in_0_1_mu_continuous_phi_1.5_figure}, \ref{heatmap_lambda_vs_mu_k_w_in_0_1_mu_continuous_phi_2_figure} and \ref{heatmap_lambda_vs_mu_k_w_in_0_1_mu_continuous_phi_2.5_figure}, which implies an increase of the size of the blue region where strategy $a$ is not advantageous. Recall that we have argued that a combined increase of $n$ and decrease of $\alpha$ is likely to lead in the long run to an increase of $M$ (Fig. \ref{number_of_links_versus_alpha_and_n_figure}). We suggest that the behavior "along the diagonal" of the matrix is an extension of the weakening of the bias when $M$ is increased in the more restrictive class (Fig. \ref{curves_delta_0_lambda_vs_phi_w_mu_in_0_1_various_M_figure}).

In our exploration of the phase space for the class of the skeleta where the degrees of counterparts do not exceed one, we assumed a right-truncated power-law with two parameters, $\alpha$ and $n$ as a model for Zipf's rank-frequency law. However, distributions giving a better fit have been considered \citep{Li2010a} and function (distribution) capturing the shape of the law of what Piotrowski called saturated samples \citep{Piotrowski2007a} should be considered in future research. Our exploration of the phase space was limited by a brute force approach neglecting the negative  correlation between $n$ and $\alpha$ that is expected in children where $\alpha$ and time are negatively correlated: as children become older, $n$ increases as a result of word learning \citep{Saxton2010a_Chapter6} but $\alpha$ decreases \citep{Baixeries2012c}. A more powerful exploration of the phase space could be performed with a realistic mathematical relationship of the expected correlation between $n$ and $\alpha$, which invites to empirical research. Finally, there might be deeper and better ways of parameterizing the class of skeleta.


\subsection{Biosemiotics}

Biosemiotics is concerned about building bridges between biology, philosophy, linguistics, and the communication sciences as announced in the front page of this journal \url{https://www.springer.com/journal/12304}. As far as we know, there is little research on the vocabulary learning bias in other species. Its confirmation in a domestic dog suggests that ``{\em the perceptual and cognitive mechanisms that may mediate the comprehension of speech were already in place before early humans began to talk}'' \citep{Kaminski2004a}. We hypothesize that the cost function $\Omega$ captures the essence of these mechanisms. A promising target for future research are ape gestures, where there has been significant progress recently on their meaning \citep{Hobaiter2014a}. As far as we know, there is no research on that bias in other domains that also fall into the scope of biosemiotics, e.g., in unicellular organisms such as bacteria. Our research has established some mathematical foundations for research on the accretion and interpretation of signs across the living world, not only among great apes, a key problem in research program of biosemiotics \citep{Kull2018a}.

The remainder of the discussion section is devoted to examine general challenges that are shared by biosemiotics and quantitative linguistics, a field that, as biosemiotics, aspires to contribute to develop a science beyond human communication.

\subsection{Science and its method}

It has been argued that a problem of research on the rank-frequency is law is the {\em The absence of novel predictions...} which {\em has led to a very peculiar situation in the cognitive sciences, where we have a profusion of theories to explain an empirical phenomenon, yet very little attempt to distinguish those theories using scientific methods.} \citep{Piantadosi2014a}.
As we have already shown the predictive power of a model whose original target was the rank-frequency laws here and in previous research \citep{Ferrer2013g}, we take this criticism as an invitation to reflect on science and its method \citep{Altmann1993,Bunge2001a}.  

\subsubsection{The generality of the patterns for theory construction}

While in psycholinguisics and the cognitive sciences a major source of evidence are often experiments involving restricted tasks or sophisticated statistical analyses covering a handful of languages (typically English and a few other Indo-European languages), quantitative linguistics aims to build theory departing from statistical laws holding in a typologically wide range of languages \citep{Koehler1987,Debowski2020a} as reflected in Fig. \ref{tab:prediction}. In addition, here we have investigated a specific vocabulary learning phenomenon that is, however, supported cross-linguistically (recall Section \ref{sec:introduction}). A recent review on the efficiency of languages, only pays attention to the law of abbreviation \citep{Gibson2019a} in contrast with the body of work that has been developed in the last decades linking laws with optimization principles (Fig. \ref{tab:prediction}), suggesting that this law is the only general pattern of languages that is shaped by efficiency or that linguistic laws are secondary for deep theorizing on efficiency. In other domains of the cognitive sciences, the importance of scaling laws has been recognized \citep{Chater1999a,Kello2010a,Baronchelli2013a}. 

\subsubsection{Novel predictions}

In section \ref{discussion_on_vocabulary_learning_subsection}, we have checked predictions of our information theoretic framework that matches knowledge on the vocabulary learning bias from past research. Our theoretical framework allows the researcher to play the game of science in another direction: use the relevant parameters to guide the design of new experiments with children or adults where more detailed predictions of the theoretical framework can be tested. For children who have about the same $n$ and $\alpha$, and $\phi=1$, our model predicts that strategy $a$ will be discarded if (Fig. \ref{heatmap_lambda_vs_mu_k_w_in_0_1_mu_continuous_phi_1_figure}) 
\begin{enumerate}
\item[(1)]
$\lambda$ is low and $\mu_k$ (Fig.\ref{vocabulary_learning_figure}) is large enough. 
\item[(2)]
$\lambda$ is high and $\mu_k$ is sufficiently low. 
\end{enumerate}
Interestingly, there is a red horizontal band in Fig. \ref{heatmap_lambda_vs_mu_k_w_in_0_1_mu_continuous_phi_1_figure}, and even for other values of $\phi$ such that $\phi \neq 1$ but keeping $\phi > 0$ (Figs. \ref{heatmap_lambda_vs_mu_k_w_in_0_1_mu_continuous_phi_0.5_figure}, \ref{heatmap_lambda_vs_mu_k_w_in_0_1_mu_continuous_phi_1.5_figure}, \ref{heatmap_lambda_vs_mu_k_w_in_0_1_mu_continuous_phi_2_figure}, \ref{heatmap_lambda_vs_mu_k_w_in_0_1_mu_continuous_phi_2.5_figure}), indicating the existence of some value of $\mu_k$ or a range of $\mu_k$ where strategy $a$ is always advantageous (notice however, that when $\phi> 1$, the band may become too narrow for an integer $\mu_k$ to fit as suggested by Figs. \ref{heatmap_lambda_vs_mu_k_w_in_0_1_phi_1.5_figure}, \ref{heatmap_lambda_vs_mu_k_w_in_0_1_phi_2_figure}, \ref{heatmap_lambda_vs_mu_k_w_in_0_1_phi_2.5_figure} in Appendix \ref{app::discrete_degrees}). Therefore the 1st concrete prediction is that, for a given child, there is likely to be some range or value of $\mu_k$ where the bias (strategy $a$) will be observed. 
The 2nd concrete prediction that can be made is on the conditions where the bias will not be observed. Although the true value of $\lambda$ is not known yet, previous theoretical research with $\phi=0$ suggests that $\lambda \leq 1/2$ in real language 
\citep{Ferrer2002a,Ferrer2004e,Ferrer2005e,Ferrer2004a}, which would imply that real speakers should satisfy only (1).
Child or adult language researchers may design experiments where $\mu_k$ is varied. If successful, that would confirm the lexicon structure hypothesis \citep{Byers-Heinlein2013a} but providing a deeper understanding.
These are just examples of experiments that could be carried out. 

\subsubsection{Towards a mathematical theory of language efficiency}

Our past and current research on the efficiency are supported by a cost function and a (analytical or numerical) mathematical procedure that links the minimization of the cost function with the target phenomena, e.g., vocabulary learning, as in research on how pressure for efficiency gives rise to Zipf's rank-frequency law, the law of abbreviation or Menzerath's law \citep{Ferrer2004e,Gustison2016a,Ferrer2019c}. In the cognitive sciences, such a cost function and the mathematical linking argument are sometimes missing \citep[e.g.,][]{Piantadosi2011a} and neglected when reviewing how languages are shaped by efficiency \citep{Gibson2019a}. 
A truly quantitative approach in the context of language efficiency is two-fold: it has to comprise either a quantitative description of the data and a quantitative theorizing, i.e. it has to employ both statistical methods of analysis and mathematical methods to define the cost and the how cost minimization leads to the expected phenomena. Our framework relies on standard information theory \citep{Cover2006a} and its extensions \citep{Ferrer2019c,Debowski2020a}. The psychological foundations of the information theoretic principles postulated in that framework and the relationships between them have already been reviewed \citep{Ferrer2015b}. How the so-called noisy-channel ``theory'' or noisy-channel hypothesis explains the results in \citep{Piantadosi2011a}, others reviewed recently \citep{Gibson2019a} or language laws in a broad sense has not yet shown, to our knowledge, with detailed enough information theory arguments. Furthermore, the major conclusions of the statistical analysis of \citep{Piantadosi2011a} have recently been shown to change substantially after improving the methods: effects attributable to plain compression are stronger than previously reported \citep{Meylan2021a}. Theory is crucial to reduce false positives and replication failures \citep{Stewart2021a}. In addition, higher order compression can explain more parsimoniously phenomena that are central in noisy-channel ``theorizing'' \citep{Ferrer2013f}.

\subsubsection{The trade-off between parsimony and perfect fit. }

Our emphasis is on generality and parsimony over perfect fit. \citet{Piantadosi2014a} makes emphasis on what models of Zipf's rank-frequency law apparently do not explain while our emphasis is on what the models do explain and the many predictions they make (Table \ref{tab:prediction}), in spite of their simple design. It is worth reminding a big lesson from machine learning, i.e. a perfect fit can be obtained simply by overfitting the data and another big lesson from the philosophy of science to machine learning and AI: sophisticated models (specially deep learning ones) are in most cases black boxes that imitate complex behavior but neither explain nor yield understanding. In our theoretical framework, the principle of contrast \citep{Clark1987a} or the mutual exclusivity bias \citep{Markman1988a,Merriman1989a} are not principles {\em per se} (or core principles) but predictions of the principle of mutual information maximization involved in explaining the emergence of Zipf's rank-frequency law \citep{Ferrer2002a,Ferrer2004e} and word order patterns \citep{Ferrer2013f}. Although there are computational models that are able to account for that vocabulary learning bias and other phenomena \citep{Frank2009a,Gulordava2020a}, ours is much simpler, transparent (in opposition to black box modeling) and to the best our knowledge, the first to predict that the bias will weaken over time providing a preliminary understanding of why this could happen. 

%% file: model_appendix.tex
This appendix is organized as follows. Section \ref{appendix_formulae} details the expressions for probabilities and entropies introduced in Section \ref{sec:model}. Section \ref{appendix_dynamic} addresses the general problem of the dynamic calculation of $\Omega$ (Eq. \ref{cost_equation_SR}) when a cell of the adjacency matrix is mutated, deriving the formulae to update these entropies once a single mutation has taken place. Finally, Section \ref{appendix_vocabulary_learning_basic} applies these formulae to derive the expressions for $\Delta$ presented in Section \ref{subsection_delta_formulae}.

\subsection{Probabilities and entropies}
\label{appendix_formulae}

In section \ref{sec:model}, we obtained an expression for the joint probability of a form and a counterpart (Eq. \ref{normalized_joint_probability_equation}) and the corresponding normalization factor, $M_\phi$ (Eq. \ref{normalization_factor_equation}). Notice that $M_0$ is the number of edges of the bipartite graph. i.e. $M = M_0$.
To ease the derivation of the marginal probabilities, we define 
\begin{eqnarray}
  \mu_{\phi,i} & = & \sum_{j=1}^m a_{ij} \omega_j^\phi \label{mu_phi_equation} \\
  \omega_{\phi,i} & = & \sum_{i=1}^n a_{ij} \mu_i^\phi. \label{omega_phi_equation}
\end{eqnarray}
Notice that $\mu_{\phi,i}$ and $\omega_{\phi,j}$ should not be confused with $\mu_i$ and $\omega_i$ (the degree of the form $i$ and of the counterpart $j$ respectively). Indeed, $\mu_i = \mu_{0,i}$ and $\omega_j = \omega_{0,j}$.
From the joint probability (Eq. \ref{normalized_joint_probability_equation}), we obtain the marginal probabilities
\begin{eqnarray}
  p(s_i) & = & \sum_{j=1}^m p(s_i, r_j) \nonumber \\
         & = & \frac{\mu_i^\phi \mu_{\phi,i}}{M_\phi} \label{marginal_probability_form} \\
  p(r_j) & = & \sum_{i=1}^n p(s_i, r_j) \nonumber \\
         & = & \frac{\omega_j^\phi \omega_{\phi,j}}{M_\phi}. \label{marginal_probability_counterpart}
\end{eqnarray}
To obtain expressions for the entropies, we use the rule
\begin{equation}
  - \sum_i \frac{x_i}{T} \log \frac{x_i}{T} = \log T - \frac{1}{T} \sum_i x_i \log x_i,
  \label{general_rule_entropies}
\end{equation}
which holds when $\sum_i x_i = T$.

We can now derive the entropies $H(S,R)$, $H(S)$ and $H(R)$.
Applying Eq. \ref{normalized_joint_probability_equation} to
\begin{equation*}
  H(S,R) = - \sum_{i=1}^n \sum_{j=1}^m p(s_i, r_j) \log p(s_i, r_j),
\end{equation*}
we obtain
\begin{equation*}
  H(S,R) = \log M_\phi - \frac{\phi}{M_\phi} \sum_{i=1}^n \sum_{j=1}^m a_{ij} (\mu_i \omega_j)^\phi \log (\mu_i \omega_j).
\end{equation*}

Applying Eq. \ref{marginal_probability_form} and the rule in Eq. \ref{general_rule_entropies}, 
\begin{equation*}
  H(S) = - \sum_{i=1}^n p(s_i) \log p(s_i)
\end{equation*}
becomes
\begin{equation*}
  H(S) = \log M_\phi - \frac{1}{M_\phi} \sum_{i=1}^n \left(\mu_i^\phi \mu_{\phi, i} \right) \log \left(\mu_i^\phi \mu_{\phi, i} \right).
\end{equation*}
By symmetry, equivalent formulae for $H(R)$ can be derived easily using Eq. \ref{marginal_probability_counterpart}, obtaining
\begin{equation*}
  H(R) = \log M_\phi - \frac{1}{M_\phi} \sum_{j=1}^m \left(\omega_j^\phi \omega_{\phi, j} \right) \log \left(\omega_j^\phi \omega_{\phi, j} \right).
\end{equation*}
Interestingly, when $\phi = 0$, the entropies simplify as
\begin{eqnarray*}
  H(S,R) & = & \log M_0 \\
  H(S)   & = & \log M_0 - \frac{1}{M_0} \sum_{i=1}^n \mu_i \log \mu_i \\
  H(R)   & = & \log M_0 - \frac{1}{M_0} \sum_{j=1}^m \omega_i \log \omega_i
\end{eqnarray*}
as expected from previous work \citep{Ferrer2004e}. Given the formulae for $H(S,R)$, $H(S)$ and $H(R)$ above, the calculation of $\Omega(\lambda)$ (Eq. \ref{cost_function_from_entropies}) is straightforward.

\subsection{Change in entropies after a single mutation in the adjacency matrix}
\label{appendix_dynamic}

Here we investigate a general problem: the change in the entropies needed to calculate $\Omega$ when there is a single mutation in the cell $(i,j)$ of the adjacency matrix, i.e. 
when a link between a form $i$ and a counterpart $j$ is added ($a_{ij}$ becomes $1$) or deleted ($a_{ij}$ becomes $0$). 
The goal of this analysis is to provide the mathematical foundations for research on the evolution of communication and in particular, the problem of learning of a new word, i.e. linking a form that was previously unlinked (Appendix \ref{appendix_vocabulary_learning_basic}), which is a particular case of mutation where $a_{ij}=0$ and $\mu_i=0$ before the mutation ($a_{ij}=1$ and $\mu_i=1$ after the mutation). 

Firstly, we express the entropies compactly as
\begin{eqnarray}
  H(S,R) & = & \log M_\phi - \frac{\phi}{M_\phi} X(S,R) \label{equation_HSR_compacted} \\
  H(S)   & = & \log M_\phi - \frac{1}{M_\phi} X(S) \label{equation_HS_compacted} \\
  H(R)   & = & \log M_\phi - \frac{1}{M_\phi} X(R) \label{equation_HR_compacted}
\end{eqnarray}
with
\begin{eqnarray}
X(S,R)      & = & \sum_{(i,j) \in E} x(s_i, r_j) \nonumber \\
X(S)        & = & \sum_{i=1}^n x(s_i) \nonumber \\
X(R)        & = & \sum_{j=1}^m x(r_j) \label{equation_XR_compacted} \\
x(s_i, r_j) & = & (\mu_i \omega_j)^\phi \log (\mu_i \omega_j) \label{equation_xsr} \\
x(s_i)      & = & \left(\mu_i^\phi \mu_{\phi, i} \right) \log \left(\mu_i^\phi \mu_{\phi, i} \right) \label{equation_xs} \\ 
x(r_j)      & = & \left(\omega_j^\phi \omega_{\phi, j} \right) \log \left(\omega_j^\phi \omega_{\phi, j} \right). \label{equation_xr}
\end{eqnarray}
We will use a prime mark to indicate the new value of a certain measure once a mutation has been produced in the adjacency matrix. 
Suppose that $a_{ij}$ mutates. Then
\begin{eqnarray}
a'_{ij}   & = & 1 - a_{ij} \nonumber \\ 
\mu_i'    & = & \mu_i + (-1)^{a_{ij}} \label{equation_mu_dynamic} \\
\omega_j' & = & \omega_j + (-1)^{a_{ij}}. \label{equation_omega_dynamic}
\end{eqnarray}
We define $\Gamma_{S}(i)$ as the set of neighbors of $s_i$ in the graph and, similarly, $\Gamma_{R}(j)$ as the set of neighbors of $r_j$ in the graph. Then $\mu'_{\phi,k}$ can only change if $k=i$ or $k \in \Gamma_R(j)$ (recall Eq. \ref{mu_phi_equation}) and $\omega'_{\phi,l}$ can only change if $l=j$ or $l \in \Gamma_R(i)$ (Eq. \ref{omega_phi_equation}). Then, for any $k$ such that $1 \leq k \leq n$, we have that 
\begin{equation}
  \mu'_{\phi,k} = \left\{
    \begin{array}{ll}
      \mu_{\phi,k} - a_{ij} \omega_j^\phi + (1 - a_{ij}) {\omega_j'}^\phi & \mbox{if~}k=i \\
      \mu_{\phi,k} - \omega_j^\phi + {\omega_j'}^\phi & \mbox{if~}k \in \Gamma_R(j) \\
      \mu_{\phi,k} & \mbox{otherwise}.
    \end{array} 
  \right.
  \label{equation_mu_phi_dynamic}
\end{equation}
Likewise, for any $l$ such that $1 \leq l \leq m$, we have that 
\begin{equation}
  \omega'_{\phi,l} = \left\{
    \begin{array}{ll}
      \omega_{\phi,l} - a_{ij} \mu_i^\phi + (1 - a_{ij}) {\mu_i'}^\phi & \mbox{if~}l=j \\
      \omega_{\phi,l} - \mu_i^\phi + {\mu_i'}^\phi & \mbox{if~}l \in \Gamma_S(i) \\
      \omega_{\phi,l} & \mbox{otherwise}.
    \end{array} 
  \right.
  \label{equation_omega_phi_dynamic}
\end{equation}
We then aim to calculate $M'_\phi$ and $X'(S,R)$ from $M_\phi$ and $X(S,R)$ (Eq. \ref{normalization_factor_equation} and Eq. \ref{equation_HSR_compacted}) respectively. Accordingly, we focus on the pairs $(s_k, r_l)$, shortly $(k,l)$, such that $\mu'_k\omega'_l = \mu_k\omega_l$ may not hold. These pairs belong to $E(i,j) \cup (i,j)$, where $E(i,j)$ is the set of edges having $s_i$ or $r_j$ at one of the ends. That is, $E(i,j)$ is the set of edges of the form $(i, l)$ where $l \in \Gamma_S(i)$ or $(k, j)$ where $k \in \Gamma_R(j)$. Then the new value of $M_{\phi}$ will be
\begin{equation}
\begin{split}
  M'_{\phi} = M_\phi &- \left[ \sum_{(k,l) \in E(i,j)}(\mu_k \omega_l)^\phi \right]   - a_{ij} (\mu_i \omega_j)^\phi \\
                     &+ \left[ \sum_{(k,l) \in E(i,j)}(\mu'_k \omega'_l)^\phi \right] + (1 - a_{ij}) (\mu'_i \omega'_j)^\phi.
\end{split}
\label{equation_M_phi_dynamic}
\end{equation}
Similarly, the new value of $X(S,R)$ will be
\begin{equation}
\begin{split}
  X'(S,R) = X(S,R) &- \left[ \sum_{(k,l) \in E(i,j)}x(s_k, r_l) \right] - a_{ij} x(s_i, r_j) \\
                   &+ \left[ \sum_{(k,l) \in E(i,j)}x'(s_k, r_l)\right] + (1 - a_{ij}) x'(s_i, r_j).
\end{split}
\label{equation_XSR_dynamic}
\end{equation}
$x'(s_i, r_j)$ can be obtained by applying $\mu'_i$ and $\omega'_j$ (Eqs. \ref{equation_mu_dynamic} and \ref{equation_omega_dynamic}) to $x(s_i, r_j)$ (Eq. \ref{equation_xsr}). The value of $H'(S,R)$ is then obtained applying $M_\phi'$ (Eq. \ref{equation_M_phi_dynamic}) and $X(S,R)'$ (Eq. \ref{equation_XSR_dynamic}) to $H(S,R)$ (Eq. \ref{equation_HSR_compacted}).

As for $H'(S)$, notice that $x'(s_k)$ can only differ from $x(s_k)$ if $\mu'_k$ and $\mu'_{\phi,k}$ change, namely when $k=i$ or $k \in \Gamma_R(j)$. Therefore
\begin{equation}
  X'(S) = X(S) - \left[ \sum_{k \in \Gamma_{R}(j)} x(s_k) \right] - a_{ij} x(s_i) + \left[ \sum_{k \in \Gamma_{R}(j)} x'(s_k) \right] + (1 - a_{ij}) x'(s_i).
  \label{equation_XS_dynamic}
\end{equation}
Similarly, $x'(s_i)$ can be obtained by applying $\mu_i$ (Eq. \ref{equation_mu_dynamic}) and $\mu_{\phi,i}$ (Eq. \ref{equation_mu_phi_dynamic}) to $x(s_i)$ (Eq \ref{equation_xs}). Then $H'(S)$ is obtained by applying $M_\phi$ and $X'(S)$ (Eqs. \ref{equation_M_phi_dynamic} and \ref{equation_XS_dynamic}) to $H(S)$ (Eq. \ref{equation_HS_compacted}).
By symmetry,
\begin{equation}
X'(R) = X(R) - \left[ \sum_{l \in \Gamma_{S}(i)} x(r_l) \right] - a_{ij} x(r_j) + \left[ \sum_{l \in \Gamma_{S}(i)} x'(r_l) \right] + (1 - a_{ij}) x'(r_j),
  \label{equation_XR_dynamic}
\end{equation}
where $x'(r_j)$ and $H'(R)$ are obtained similarly, applying $\omega_j$ (Eq. \ref{equation_omega_phi_dynamic}) $x(r_j)$ (Eq. \ref{equation_xr}) and finally $H(R)$ (Eq. \ref{equation_HR_compacted}).


\subsection{Derivation of $\Delta$}
\label{appendix_vocabulary_learning_basic}

Following from the previous sections, we set off to obtain expressions for $\Delta$ for each of the skeleton classes we set out to study. As before, we denote the value of a variable after applying either strategy with a prime mark, meaning that it is a modified value after a mutation in the adjacency matrix.
We also use a subindex $a$ or $b$ to indicate the vocabulary learning strategy corresponding to the mutation. 
A value without prime mark then denotes the state of that variable before applying either strategy. 

Firstly, we aim to obtain an expression for $\Delta$ that depends on the new values of the entropies after either strategy $a$ or $b$ has been chosen. 
Combining $\Delta(\lambda)$ (Eq. \ref{eq:delta_from_omega_a_b}) with $\Omega(\lambda)$ (Eq. \ref{cost_function_from_entropies}), one obtains
\begin{equation*}
  \Delta(\lambda) = (1 - 2 \lambda) (H'_a(S) - H'_b(S)) - \lambda (H'_a(R) - H'_b(R)) + \lambda (H'_a(S, R) - H'_b(S, R)).
\end{equation*}
The application of $H(S,R)$ (Eq. \ref{equation_HSR_compacted}), $H(S)$ (Eq. \ref{equation_HS_compacted}) and $H(R)$ (Eq. \ref{equation_HR_compacted}), yields
\begin{equation}
  \Delta(\lambda) = (1-2\lambda)\log \frac{{M'_\phi}_a}{{M'_\phi}_b} - \frac{1}{{M'_\phi}_a {M'_\phi}_b} \left[ (1-2\lambda) \Delta_{X(S)} - \lambda \Delta_{X(R)} + \lambda \phi \Delta_{X(S,R)} \right]
  \label{equation_delta_from_M_XS_XR_XSR_ab}
\end{equation}
with
\begin{eqnarray*}
\Delta_{X(S)}   & = & {M'_\phi}_b X'_a(S) - {M'_\phi}_a X'_b(S) \\
\Delta_{X(R)}   & = & {M'_\phi}_b X'_a(R) - {M'_\phi}_a X'_b(R) \\
\Delta_{X(S,R)} & = & {M'_\phi}_b X'_a(S,R) - {M'_\phi}_a X'_b(S,R).
\end{eqnarray*}

Now we find expressions for ${M'_\phi}_a$, $X'_a(S,R)$, $X'_a(S)$, $X'_a(R)$, ${M'_\phi}_b$, $X'_b(S,R)$, $X'_b(S)$, $X'_b(R)$.
To obtain generic expressions for $M'_\phi$, $X'(S,R)$, $X'(S)$ and $X'(R)$ via Eqs. \ref{equation_M_phi_dynamic}, \ref{equation_XSR_dynamic}, \ref{equation_XS_dynamic} and \ref{equation_XR_dynamic},
we define mathematically the state of the bipartite matrix before and after applying either strategy $a$ or $b$ with the following restrictions
\begin{itemize}
\item
  ${a_{ij}}_a = {a_{ij}}_b = 0$. Form $i$ and counterpart $j$ are initially unconnected.
\item 
  ${\mu_i}_a = {\mu_i}_b = 0$. Form $i$ has initially no connections.
\item 
  ${\mu'_i}_a = {\mu'_i}_b = 1$. Form $i$ will have one connection afterwards.
\item 
  ${\omega_j}_a = 0$. In case $a$, counterpart $j$ is initially disconnected.
\item 
  ${\omega_j}_b = \omega_j > 0$. In case $b$, counterpart $j$ has initially at least one connection.
\item 
  ${\omega'_j}_a = 1$. In case $a$, counterpart $j$ will have one connection afterwards.
\item 
  ${\omega'_j}_b = \omega_j + 1$. In case $b$, counterpart $j$ will have one more connection afterwards.
\item 
  ${\Gamma_S}_a(i) = {\Gamma_S}_b(i) = \varnothing$. Form $i$ has initially no neighbors.
\item 
  ${\Gamma_R}_a(j) = \varnothing$. In case $a$, counterpart $j$ has initially no neighbors.
\item 
  ${\Gamma_R}_b(j) \neq \varnothing$. In case $b$, counterpart $j$ has initially some neighbors.
\item 
  $E_a(i,j) = \varnothing$. In case $a$, there are no links with $i$ or $j$ at one of their ends.
\item 
  $E_b(i,j) = \left\{ (k,j) | k \in \Gamma_R(j) \right\}$. In case $b$, there are no links with $i$ at one of their ends, only with $j$.
\end{itemize}

We can apply these restrictions to $x(s_i, r_j)$, $x(s_i)$ and $x(r_j)$ (Eqs. \ref{equation_xsr}, \ref{equation_xs} and \ref{equation_xr}) to obtain expressions of $x_a'(s_i)$, $x_b'(s_i)$, $x_b'(r_j)$ and $x_b'(s_i, r_j)$ that depend only on the initial values of $\omega_j$ and $\omega_{\phi,j}$
\begin{eqnarray}
x'(s_i) & = & \phi {\omega'_j}^\phi \log \omega'_j \nonumber \\
x'_a(s_i) & = & 0 \label{equation_xsi_dynamic_a} \\
x'_a(r_j) & = & 0 \label{equation_xrj_dynamic_a} \\
x'_b(s_i) & = & \phi (\omega_j + 1)^\phi \log (\omega_j + 1) \label{equation_xsi_dynamic_b} \\
x'_b(r_j) & = & (\omega_j + 1)^\phi (\omega_{\phi,j} + 1) \log ((\omega_j + 1)^\phi (\omega_{\phi,j} + 1)) \label{equation_xrj_dynamic_b} \\
x'_a(s_i, r_J) & = & 0 \label{equation_xsirj_dynamic_a} \\
x'_b(s_i, r_j) & = & (\omega_j + 1)^\phi \log (\omega_j + 1). \label{equation_xsirj_dynamic_b}
\end{eqnarray}
Additionally, for any forms $s_k$ such that $k \in {\Gamma_R}_b(j)$ (that is, for every form that counterpart $j$ is connected to), we can also obtain expressions that depend only on the initial values of $\omega_j$, $\omega_{\phi,j}$, $\mu_k$ and $\mu_{\phi,k}$ using the same restrictions and equations
\begin{align}
  x_b(s_k, r_j) &= \omega_j^\phi (\mu_k^\phi \log \mu_k) + (\omega_j^\phi \log \omega_j) \mu_k^\phi \label{equation_xskrj_b} \\
  x'_b(s_k, r_j) &= (\omega_j + 1)^\phi (\mu_k^\phi \log \mu_k) + \left[(\omega_j + 1)^\phi \log (\omega_j + 1)\right] \mu_k^\phi \label{equation_xskrj_dynamic_b} \\
  \begin{split}
    x'_b(s_k) &= \left\{\mu_k^\phi \mu_{\phi,k} + \mu_k^\phi \left[-\omega_j^\phi + (\omega_j + 1)^\phi\right]\right\} \\
    &\qquad \cdot \log\left[(\mu_k^\phi \mu_{\phi,k}) \left(\frac{\mu_{\phi,k} - \omega_j^\phi +(\omega_j+1)^\phi}{\mu_{\phi,k}}\right)\right]
  \end{split} \nonumber \\
  \begin{split}
    &= s_b(s_k) + \left[(\omega_j+1)^\phi-\omega_j^\phi\right] \mu_k^\phi \log\left\{\mu_k^\phi\left[\mu_{\phi,k}-\omega_j^\phi+(\omega_j+1)^\phi\right]\right\} \\
    &\qquad + \mu_k^\phi\mu_{\phi,k}\log\left(\frac{\mu_{\phi,k}-\omega_j^\phi+(\omega_j+1)^\phi}{\mu_{\phi,k}}\right).
  \end{split} \label{equation_xsk_dynamic}
\end{align}
Applying the restrictions to $M'_\phi$ (Eq. \ref{equation_M_phi_dynamic}), we can also obtain an expression that depends only on some initial values
\begin{eqnarray}
  {M'_\phi}_a & = & M_\phi + 1 \label{equation_M_phi_dynamic_a} \\
  {M'_\phi}_b & = & M_\phi + \left[(\omega_j + 1)^\phi - \omega_j^\phi\right] \omega_{\phi,j} + (\omega_j + 1)^\phi. \label{equation_M_phi_dynamic_b}
\end{eqnarray}
Applying now the expressions for $x'_a(s_i, r_j)$ (Eq. \ref{equation_xsirj_dynamic_a}), $x'_b(s_i, r_j)$ (Eq. \ref{equation_xsirj_dynamic_b}), $x_b(s_k, r_j)$ (Eq. \ref{equation_xskrj_b}) and $x'_b(s_k, r_j)$ (Eq. \ref{equation_xskrj_dynamic_b}) to $X'(S,R)$ (Eq. \ref{equation_XSR_dynamic}), along with the restrictions, we obtain
\begin{align}
  X'_a(S, R) &= X(S, R) \label{equation_XSR_dynamic_a} \\
  \begin{split}
  X'_b(S, R) &= X(S, R) + \left[(\omega_j+1)^\phi - \omega_j^\phi\right] \sum_{k \in \Gamma_R(j)} \mu_k^\phi \log \mu_k \\
             &\qquad + \omega_{\phi,j} \left[(\omega_j+1)^\phi \log(\omega_j+1) - \omega_j^\phi \log(\omega_j)\right] \\
             &\qquad + (\omega_j + 1)^\phi \log(\omega_j + 1).
  \end{split} \label{equation_XSR_dynamic_b}
\end{align}
Similarly, we apply $x'_a(s_i)$ (Eq. \ref{equation_xsi_dynamic_a}), $x'_b(s_i)$ (Eq. \ref{equation_xsi_dynamic_b}) and $x'_b(s_k)$ (Eq. \ref{equation_xsk_dynamic}) to $X'(S)$ (Eq. \ref{equation_XS_dynamic}) as well as the restrictions and obtain
\begin{align}
  X'_a(S) &= X(S) \label{equation_XS_dynamic_a} \\
  \begin{split}
    X'_b(S) &= X(S) + \phi (\omega_j+1)^\phi \log(\omega_j+1) \\
    &\qquad + \left[(\omega_j+1)^\phi - \omega_j^\phi\right] \sum_{k \in \Gamma_R(j)} \mu_k^\phi \log\left\{\mu_k^\phi \left[\mu_{\phi,k} - \omega_j^\phi + (\omega_j + 1)^\phi\right]\right\} \\
    &\qquad + \sum_{k \in \Gamma_R(j)} \mu_k^\phi \mu_{\phi,k} \log\left(\frac{\mu_{\phi,k} - \omega_j^\phi + (\omega_j + 1)^\phi}{\mu_{\phi,k}}\right).
  \end{split} \label{equation_XS_dynamic_b}
\end{align}
We apply $x'_a(r_j)$ (Eq. \ref{equation_xrj_dynamic_a}) and $x'_b(r_j)$ (Eq. \ref{equation_xrj_dynamic_b}) to $X'(R)$ (Eq. \ref{equation_XR_dynamic}) along with the restrictions and obtain
\begin{align}
  X'_a(R) &= X(R) \label{equation_XR_dynamic_a} \\
  \begin{split}
    X'_b(R) &= X(R) - \omega_j^\phi \omega_{\phi,j} \log(\omega_j^\phi \omega_{\phi,j}) \\
    &\qquad + (\omega_j + 1)^\phi (\omega_{\phi,j} + 1) \log \left[(\omega_j + 1)^\phi (\omega_{\phi,j} + 1)\right].
  \end{split} \label{equation_XR_dynamic_b}
\end{align}
At this point we could attempt to build an expression for $\Delta$ for the most general case. However, this expression would be extremely complex. Instead, we study the expression of $\Delta$ in three simplifying conditions: the case $\phi=0$ and the two classes of skeleta.

\subsubsection{The case $\phi = 0$}
\label{vocabulary_learning_phi_0}

The condition $\phi=0$ corresponds to a model that is a precursor of the current model \cite{Ferrer2013g}, and that we use to ensure our that our general expressions are correct.
We apply $\phi=0$ to the expressions in Section \ref{appendix_vocabulary_learning_basic}.
$M'_a$ and $M'_b$ (Eqs. \ref{equation_M_phi_dynamic_a} and \ref{equation_M_phi_dynamic_b}) both simplify as
\begin{equation}
  M'_a = M'_b = M + 1. \label{equation_M_phi_dynamic_ab_phi0}
\end{equation}
$X'_a(S, R)$ and $X'_b(S, R)$ (Eqs. \ref{equation_XSR_dynamic_a} and \ref{equation_XSR_dynamic_b}) simplify as
\begin{eqnarray}
  X'_a(S, R) & = & X(S, R) \label{equation_XSR_dynamic_a_phi0} \\
  X'_b(S, R) & = & X(S, R) + (\omega_j+1)\log(\omega_j+1) - \omega_j\log(\omega_j). \label{equation_XSR_dynamic_b_phi0}
\end{eqnarray}
$X'_a(S)$ and $X'_b(S)$ (Eqs. \ref{equation_XS_dynamic_a} and \ref{equation_XS_dynamic_b}) both simplify as
\begin{equation}
  X'_a(S) = X'_b(S) = X(S). \label{equation_XS_dynamic_ab_phi0}
\end{equation}
$X'_a(R)$ and $X'_b(R)$ (Eqs. \ref{equation_XR_dynamic_a} and \ref{equation_XR_dynamic_b}) simplify as
\begin{eqnarray}
  X'_a(R) & = & X(R) \label{equation_XR_dynamic_a_phi0} \\
  X'_b(R) & = & X(R) - \omega_j \log(\omega_j) + (\omega_j + 1) \log (\omega_j + 1). \label{equation_XR_dynamic_b_phi0}
\end{eqnarray}
The application of Eqs. \ref{equation_M_phi_dynamic_ab_phi0}, \ref{equation_XSR_dynamic_a_phi0}, \ref{equation_XSR_dynamic_b_phi0}, \ref{equation_XS_dynamic_ab_phi0}, \ref{equation_XR_dynamic_a_phi0} and \ref{equation_XR_dynamic_b_phi0} into the expression of $\Delta$ (Eq. \ref{equation_delta_from_M_XS_XR_XSR_ab}) results in the expression for $\Delta$ (Eq. \ref{eq:delta_lambda_phi_0}) presented in Section \ref{sec:introduction}.
\subsubsection{Counterpart degrees do not exceed one}
\label{vocabulary_learning_wj_in_01}

In this case we assume that $\omega_j \in \{0,1\}$ for every $r_j$ and further simplify the expressions from \ref{appendix_vocabulary_learning_basic} under this assumption. This is the most relaxed of the conditions and so these expressions remain fairly complex.

${M'_\phi}_a$ and ${M'_\phi}_b$ (Eqs. \ref{equation_M_phi_dynamic_a} and \ref{equation_M_phi_dynamic_b}) simplify as
\begin{eqnarray}
  {M'_\phi}_a & = & M_\phi + 1 \label{equation_M_phi_dynamic_counterparts_one_a} \\
  {M'_\phi}_b & = & M_\phi + (2^\phi - 1) \mu_k^\phi + 2^\phi \label{equation_M_phi_dynamic_counterparts_one_b}
\end{eqnarray}
with
\begin{equation*}
  M_\phi = \sum_{i=1}^n \mu_i^{\phi+1},
\end{equation*}
$X'_a(S, R)$ and $X'_b(S, R)$ (Eqs. \ref{equation_XSR_dynamic_a} and \ref{equation_XSR_dynamic_b}) simplify as
\begin{eqnarray}
  X'_a(S, R) & = & X(S, R) \label{equation_XSR_dynamic_counterparts_one_a} \\
  X'_b(S, R) & = & X(S, R) + (2^\phi - 1) \mu_k^\phi \log \mu_k + (\mu_k^\phi + 1) 2^\phi \log 2 \label{equation_XSR_dynamic_counterparts_one_b}
\end{eqnarray}
with
\begin{equation}
  X(S,R) = \sum_{i=1}^n \mu_i^{\phi+1} \log \mu_i.
  \label{equation_XSR_counterparts_not_exceed_one}
\end{equation}
$X'_a(S)$ and $X'_b(S)$ (Eqs. \ref{equation_XS_dynamic_a} and \ref{equation_XS_dynamic_b}) simplify as
\begin{align}
  X'_a(S) &= X(S) \label{equation_XS_dynamic_counterparts_one_a} \\
  \begin{split}
    X'_b(S) &= X(S) + (2^\phi - 1) \mu_k^\phi \log\left[\mu_k^\phi (\mu_k - 1 + 2^\phi)\right] \\
    &\qquad + \mu_k^{\phi+1} \log \left( \frac{\mu_k - 1 + 2^\phi}{\mu_k} \right) + \phi 2^\phi \log(2)
  \end{split} \label{equation_XS_dynamic_counterparts_one_b}
\end{align}
with
\begin{eqnarray*}
  X(S) & = & \sum_{i=1}^n \mu_i^\phi \mu_{\phi,i} \log(\mu_i^\phi \mu_{\phi,i}) \\
       & = & \sum_{i=1}^n \mu_i^\phi \mu_i \log(\mu_i^\phi \mu_i) \\
       & = & (\phi+1) \sum_{i=1}^n \mu_i^{\phi+1} \log \mu_i \\
       & = & (\phi+1) X(S,R).
\end{eqnarray*}
$X'_a(R)$ and $X'_b(R)$ (Eqs. \ref{equation_XR_dynamic_a} and \ref{equation_XR_dynamic_b}) simplify as
\begin{eqnarray}
  X'_a(R) & = & X(R) \label{equation_XR_dynamic_counterparts_one_a} \\
  X'_b(R) & = & X(R) - \phi \mu_k^\phi \log(\mu_k) + 2^\phi (\mu_k^\phi + 1) \log\left[2^\phi (\mu_k^\phi + 1)\right] \label{equation_XR_dynamic_counterparts_one_b}
\end{eqnarray}
with
\begin{equation}
  X(R) = \phi X(S,R).
  \label{equation_XR_counterpart_degrees_one}
\end{equation}
The previous result on $X(R)$ deserves a brief explanation as it is not straightforward. Firstly, we apply the definition of $x(r_j)$ (Eq. \ref{equation_xr}) to that of $X(R)$ (Eq. \ref{equation_XR_compacted})
\begin{equation*}
  X(R) = \sum_{j=1}^m \omega_j^\phi \omega_{\phi,j} \log(\omega_j^\phi \omega_{\phi,j}).
\end{equation*}
As counterpart degrees are one, $\omega_j = 1 $ and $\omega_{\phi,j} = \mu_{i \leftarrow j}^\phi$, where $i \leftarrow j$ is used to indicate that we refer to the form $i$ that the counterpart $j$ is connected to (see Eq. \ref{omega_phi_equation}). That leads to
\begin{equation*}
  X(R) = \phi \sum_{j=1}^m \mu_{i \leftarrow j}^\phi \log(\mu_{i \leftarrow j}).
\end{equation*}
In order to change the summation over each $j$ (every counterpart) to a summation over each $i$ (every form) we must take into account that when summing over $j$, we accounted for each form $i$ a total of $\mu_i$ times. Therefore we need to multiply by $\mu_i$ in order for the summations to be equivalent, as otherwise we would be accounting for each form $i$ only once. This leads to 
\begin{equation*}
  X(R) = \phi \sum_{i=1}^n \mu_i^{\phi+1} \log \mu_i
\end{equation*}
and eventually Eq. \ref{equation_XR_counterpart_degrees_one} thanks to Eq. \ref{equation_XSR_counterparts_not_exceed_one}.

The application of Eqs. \ref{equation_M_phi_dynamic_counterparts_one_a}, \ref{equation_M_phi_dynamic_counterparts_one_b}, \ref{equation_XSR_dynamic_counterparts_one_a}, \ref{equation_XSR_dynamic_counterparts_one_b}, \ref{equation_XS_dynamic_counterparts_one_a}, \ref{equation_XS_dynamic_counterparts_one_b}, \ref{equation_XR_dynamic_counterparts_one_a} and \ref{equation_XR_dynamic_counterparts_one_b}
into the expression of $\Delta$ (Eq. \ref{equation_delta_from_M_XS_XR_XSR_ab}) results in the expression for $\Delta$ (Eq. \ref{eq:delta_w_in_0_1}) presented in Section \ref{sec:introduction}.
If we apply the two extreme values of $\lambda$, i.e. $\lambda=0$ and $\lambda=1$, to that equation, we obtain the following expressions
\begin{eqnarray*}
    \Delta(0) & = & \log \left( \frac{M_\phi+1}{M_\phi+(2^\phi-1)\mu_k^\phi+2^\phi} \right) \\
              &   & \qquad + \frac{1}{M_\phi+(2^\phi-1)\mu_k^\phi+2^\phi} \Bigg\{ \phi 2^\phi \mu_k^\phi \log(\mu_k) \\
              &   & \qquad - \Big[ (\phi + 1) \frac{X(S,R) (2^\phi - 1) (\mu_k^\phi + 1)}{M_\phi + 1} - \phi 2^\phi \log(2) \\
              &   & \qquad + \mu_k^\phi \left[ \log(\mu_k) (\mu_k + \phi) - (\mu_k - 1 + 2^\phi) \log(\mu_k - 1 + 2^\phi) \right] \Big] \Bigg\}
\end{eqnarray*}
\begin{eqnarray*}
    \Delta(1) & = & -\log \left( \frac{M_\phi+1}{M_\phi+(2^\phi-1)\mu_k^\phi+2^\phi} \right) \\
              &   & \qquad - \frac{1}{M_\phi+(2^\phi-1)\mu_k^\phi+2^\phi} \Bigg\{ (\mu_k^\phi + 1) 2^\phi \log(\mu_k^\phi + 1) \\
              &   & \qquad - \Big[ (\phi + 1) \frac{X(S,R) (2^\phi - 1) (\mu_k^\phi + 1)}{M_\phi + 1} - \phi 2^\phi \log(2) \\
              &   & \qquad + \mu_k^\phi \left[ \log(\mu_k) (\mu_k + \phi) - (\mu_k - 1 + 2^\phi) \log(\mu_k - 1 + 2^\phi) \right] \Big] \Bigg\}.
\end{eqnarray*}
\subsubsection{Vertex degrees do not exceed one}
\label{vocabulary_learning_wj_mui_in_01}

As seen in Section \ref{subsection_delta_formulae}, for this class we are working under the two conditions that $\omega_j \in \{0,1\}$ for every $r_j$ and $\mu_i \in \{0,1\}$ for every $s_i$. We can simplify the expressions from \ref{appendix_vocabulary_learning_basic}.
${M'_\phi}_a$ and ${M'_\phi}_b$ (Eqs. \ref{equation_M_phi_dynamic_counterparts_one_a} and \ref{equation_M_phi_dynamic_counterparts_one_b}) simplify as
\begin{eqnarray}
  {M'_\phi}_a & = & M_\phi + 1 \label{equation_M_phi_dynamic_vertex_one_a} \\
  {M'_\phi}_b & = & M_\phi + 2^{\phi+1} - 1, \label{equation_M_phi_dynamic_vertex_one_b}
\end{eqnarray}
where $M_\phi = M_0 = M$, the number of edges in the bipartite graph.
$X'_a(S, R)$ and $X'_b(S, R)$ (Eqs. \ref{equation_XSR_dynamic_counterparts_one_a} and \ref{equation_XSR_dynamic_counterparts_one_b}) simplify as
\begin{eqnarray}
  X'_a(S, R) & = & 0 \label{equation_XSR_dynamic_vertex_one_a} \\
  X'_b(S, R) & = & 2^{\phi+1} \log 2. \label{equation_XSR_dynamic_vertex_one_b}
\end{eqnarray}
$X'_a(S)$ and $X'_b(S)$ (Eqs. \ref{equation_XS_dynamic_counterparts_one_a} and \ref{equation_XS_dynamic_counterparts_one_b}) simplify as
\begin{eqnarray}
  X'_a(S) & = & 0 \label{equation_XS_dynamic_vertex_one_a} \\
  X'_b(S) & = & \phi 2^{\phi + 1} \log 2. \label{equation_XS_dynamic_vertex_one_b}
\end{eqnarray}
$X'_a(R)$ and $X'_b(R)$ (Eqs. \ref{equation_XR_dynamic_counterparts_one_a} and \ref{equation_XR_dynamic_counterparts_one_b}) simplify as
\begin{eqnarray}
  X'_a(R) & = & 0 \label{equation_XR_dynamic_vertex_one_a} \\
  X'_b(R) & = & (\phi+1) 2^{\phi+1} \log 2. \label{equation_XR_dynamic_vertex_one_b}
\end{eqnarray}
Combining Eqs. \ref{equation_M_phi_dynamic_vertex_one_a}, \ref{equation_M_phi_dynamic_vertex_one_b}, \ref{equation_XSR_dynamic_vertex_one_a}, \ref{equation_XSR_dynamic_vertex_one_b}, \ref{equation_XS_dynamic_vertex_one_a}, \ref{equation_XS_dynamic_vertex_one_b}, \ref{equation_XR_dynamic_vertex_one_a}, \ref{equation_XR_dynamic_vertex_one_b} into the equation for $\Delta$ (Eq. \ref{equation_delta_from_M_XS_XR_XSR_ab}) results in the expression for $\Delta$ (Eq.\ref{eq:delta_w_mu_in_0_1}) presented in Section \ref{sec:introduction}.
When the extreme values, i.e. $\lambda=0$ and $\lambda=1$, are applied to this equation, we obtain the following expressions
\begin{eqnarray*}
  \Delta(0) & = & -\log \left( 1 + \frac{2(2^\phi - 1)}{M + 1} \right)  + \frac{2^{\phi+1} \phi \log(2)}{M + 2^{\phi+1} - 1} \\
  \Delta(1) & = & \log \left( 1 + \frac{2(2^\phi - 1)}{M + 1} \right)  - \frac{2^{\phi+1} (\phi + 1) \log(2)}{M + 2^{\phi+1} - 1}.
\end{eqnarray*}

%% file: form_degrees_appendix.tex
Here we develop the implications of Eq. \ref{degree_versus_rank_equation} with $\mu_{n-1} = 1$ and $\mu_{n} = 0$. 
Imposing $\mu_{n-1} = 1$, 
we get
\begin{equation*} 
c = (n - 1)^\tau. 
\end{equation*}
Inserting the previous results into the definition of $p(s_i)$ when $\omega_j \leq 1$, we have that
\begin{eqnarray*}
p(s_i) & = & \frac{1}{M_\phi}\mu_i^{\phi+1} \\
       & = & c'i^{-\alpha},
\end{eqnarray*}
with
\begin{eqnarray*}
  \alpha = \tau(\phi+1) \\
  c' = \frac{(n - 1)^\alpha}{M_\phi}.
\end{eqnarray*}

A continuous approximation to vertex degrees and the number of edges gives
\begin{eqnarray*} 
M & = & \sum_{i=1}^n \mu_i \\
  & = & c \sum_{i=1}^{n-1} i^{-\tau} \\
  & = & (n-1)^\tau \sum_{i=1}^{n-1} i^{-\tau}. 
\end{eqnarray*}
Thanks to well-known integral bounds \citep[pp. 50-51]{Cormen1990_summations}, we have that
\begin{equation*}
\int_{1}^n i^{-\tau} di \leq \sum_{i=1}^{n-1} i^{-\tau} \leq 1 + \int_{1}^{n-1} i^{-\tau} di.
\end{equation*}
as $\tau \geq 0$ by definition. 
When $\tau=1$, one obtains
\begin{equation*}
\log n \leq \sum_{i=1}^{n-1} i^{-1} \leq 1 + \log(n-1).
\end{equation*}
When $\tau \neq 1$, one obtains
\begin{equation*}
\frac{1}{\tau - 1}\left[1 - n^{1 -\tau}\right] \leq \sum_{i=1}^{n-1} i^{-\tau} \leq 1 + \frac{1}{\tau - 1}\left[1 - (n-1)^{1 -\tau}\right].
\end{equation*}
Combining the results above, one obtains
\begin{equation*}
(n-1) \log n \leq M \leq (n-1)[1 + \log(n-1)]
\end{equation*}
for $\tau = 1$ and
\begin{equation*}
(n-1)^\tau\left[\frac{1}{\tau - 1}\left[1 - n^{1 -\tau}\right]\right] \leq M \leq (n-1)^\tau\left[1 + \frac{1}{\tau - 1}\left[1 - (n-1)^{1 -\tau}\right]\right]
\end{equation*}
for $\tau \neq 1$.

%% file: other_values_phi_appendix.tex
In Section \ref{sec:results}, heatmaps were used to analyze $\Delta$ takes for distinct sets of parameters. For the class of skeleta where counterpart degrees do not exceed one, only heatmaps corresponding to $\phi=0$ (Fig. \ref{heatmap_lambda_vs_mu_k_w_in_0_1_mu_continuous_phi_0_figure}) and $\phi=1$ (Figs. \ref{heatmap_lambda_vs_mu_k_w_in_0_1_mu_continuous_phi_1_figure}, \ref{heatmap_lambda_vs_alpha_w_in_0_1_mu_continuous_phi_1_figure} and \ref{heatmap_lambda_vs_n_w_in_0_1_mu_continuous_phi_1_figure}) were presented. The summary figures presented in that same section (Figs. \ref{curves_delta_0_lambda_vs_mu_k_w_in_0_1_mu_continuous_various_phi_figure}, \ref{curves_delta_0_lambda_vs_alpha_w_in_0_1_mu_continuous_various_phi_figure} and \ref{curves_delta_0_lambda_vs_n_w_in_0_1_mu_continuous_various_phi_figure}) already displayed the boundaries between positive and negative values of $\Delta$ for the whole range of values of $\phi$. Heatmaps for the remainder of values of $\phi$ are presented next.

\begin{figure}
\centering
\includegraphics[width = \textwidth]{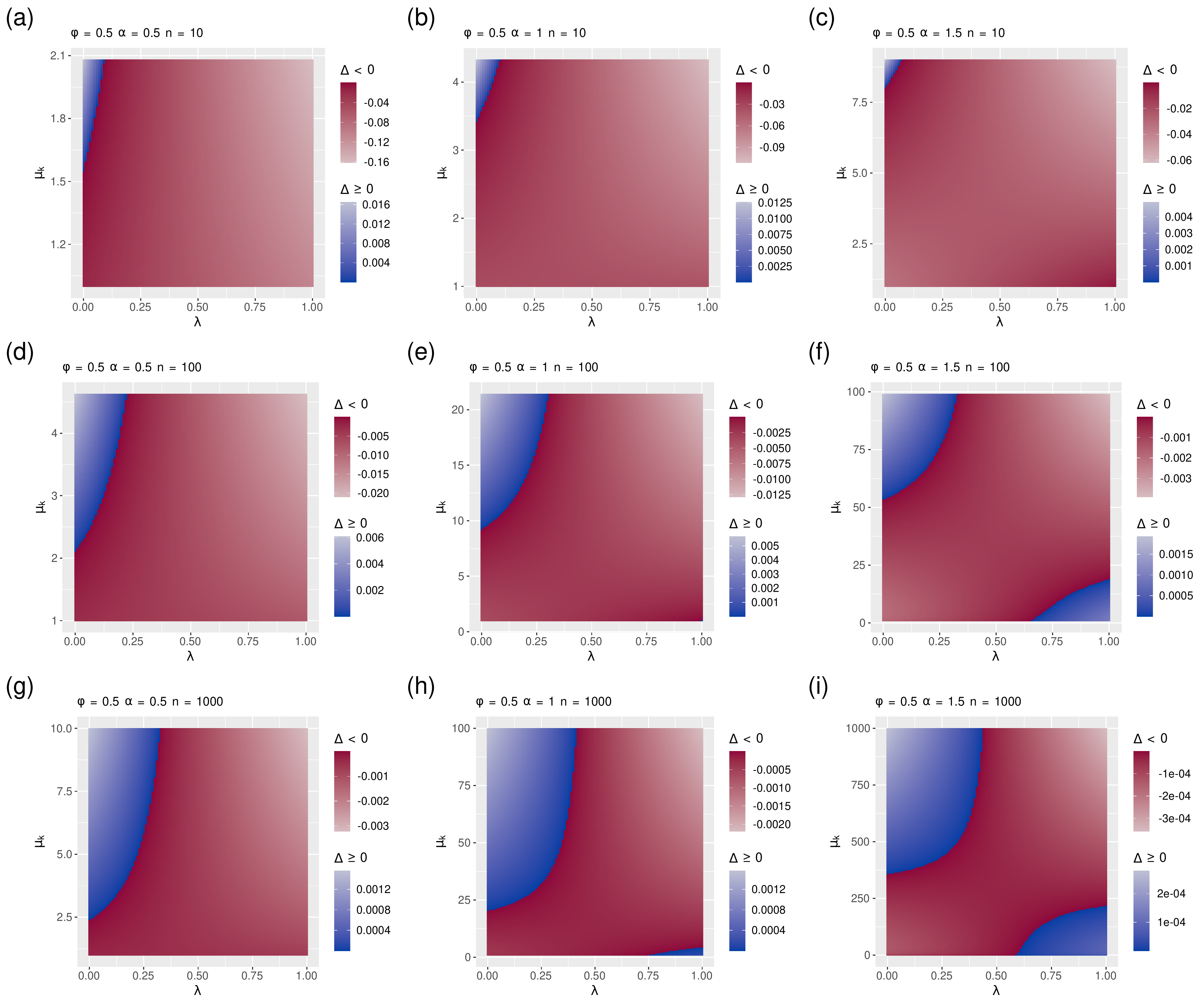}
\caption{\label{heatmap_lambda_vs_mu_k_w_in_0_1_mu_continuous_phi_0.5_figure} Same as in Fig. \ref{heatmap_lambda_vs_mu_k_w_in_0_1_mu_continuous_phi_0_figure} but with $\phi=0.5$.
}
\end{figure}

\begin{figure}
\centering
\includegraphics[width = \textwidth]{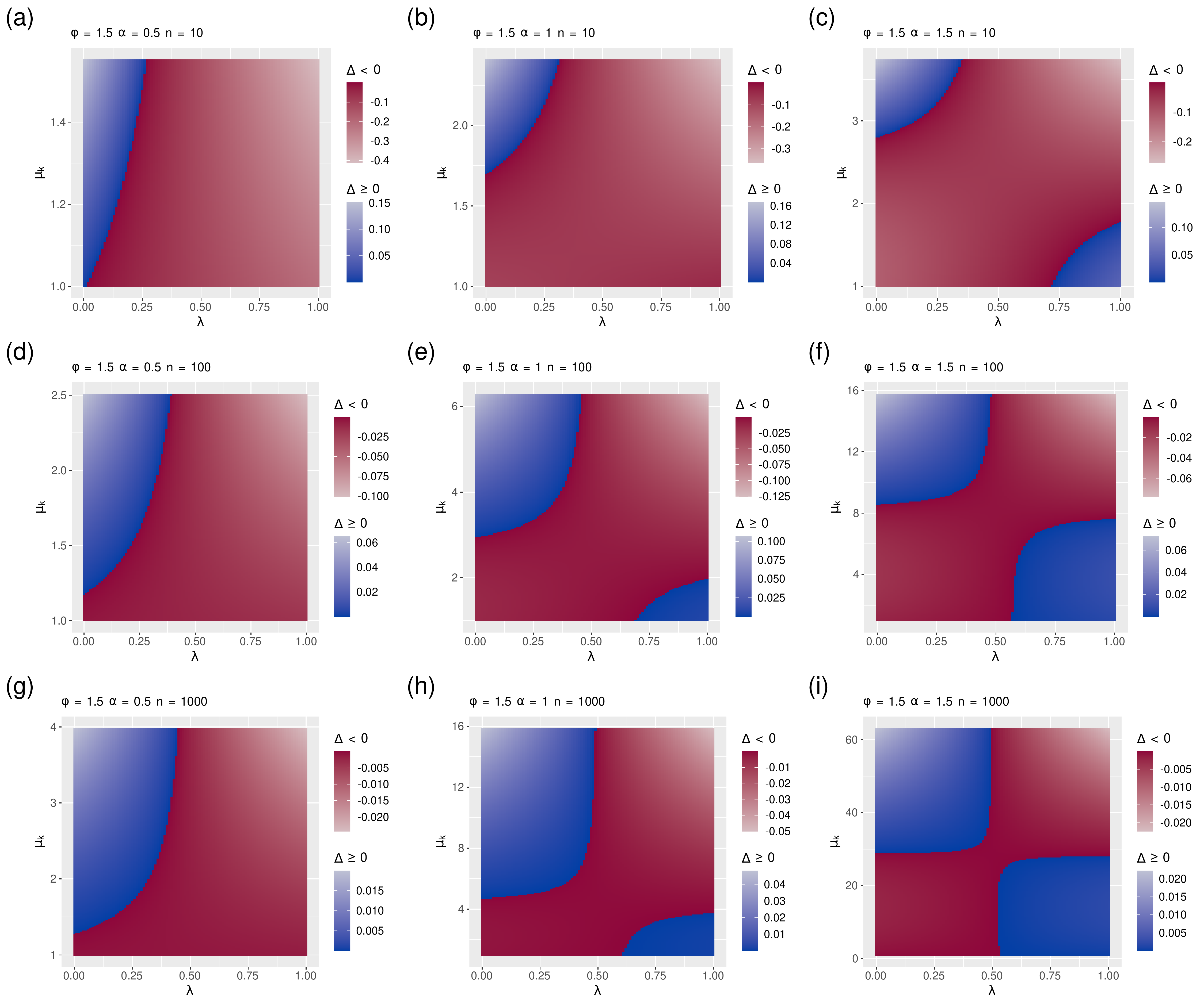}
\caption{\label{heatmap_lambda_vs_mu_k_w_in_0_1_mu_continuous_phi_1.5_figure} Same as in Fig. \ref{heatmap_lambda_vs_mu_k_w_in_0_1_mu_continuous_phi_0_figure} but with $\phi=1.5$.
}
\end{figure}

\begin{figure}
\centering
\includegraphics[width = \textwidth]{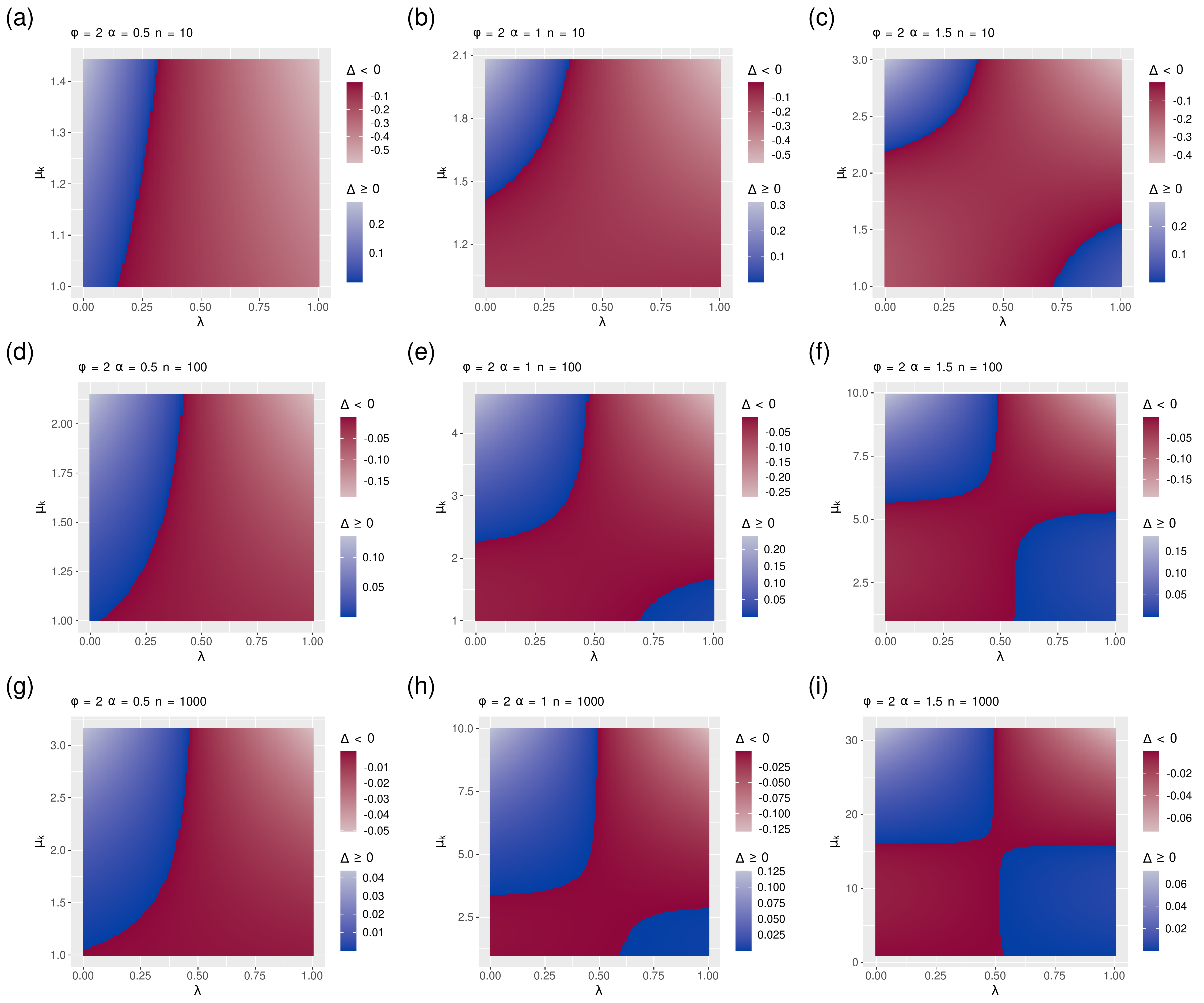}
\caption{\label{heatmap_lambda_vs_mu_k_w_in_0_1_mu_continuous_phi_2_figure} Same as in Fig. \ref{heatmap_lambda_vs_mu_k_w_in_0_1_mu_continuous_phi_0_figure} but with $\phi=2$.
}
\end{figure}

\begin{figure}
\centering
\includegraphics[width = \textwidth]{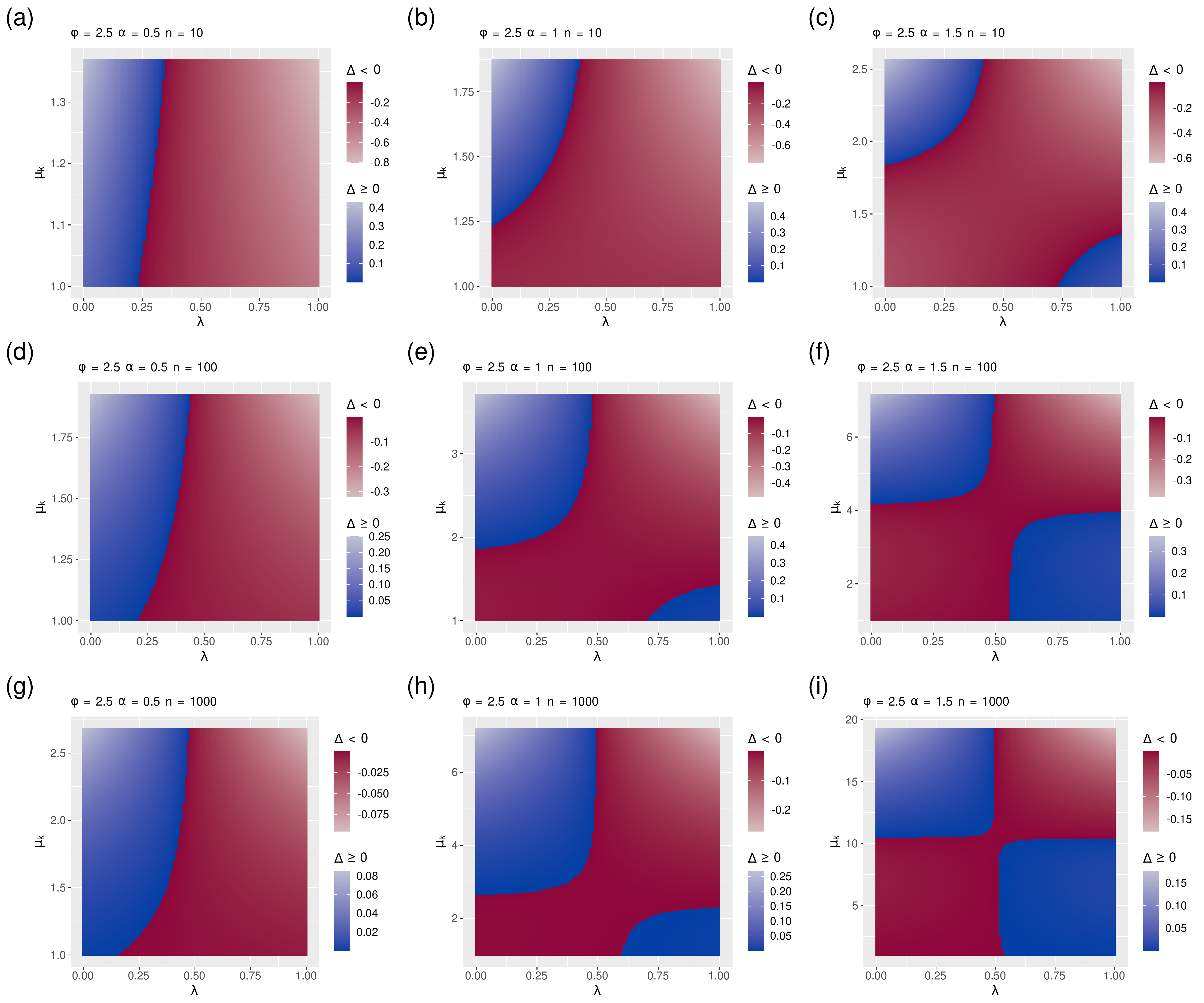}
\caption{\label{heatmap_lambda_vs_mu_k_w_in_0_1_mu_continuous_phi_2.5_figure} Same as in Fig. \ref{heatmap_lambda_vs_mu_k_w_in_0_1_mu_continuous_phi_0_figure} but with $\phi=2.5$.
}
\end{figure}

\paragraph{Heatmaps of $\Delta$ as a function of $\lambda$ and $\mu_k$}
Figures \ref{heatmap_lambda_vs_mu_k_w_in_0_1_mu_continuous_phi_0.5_figure}, \ref{heatmap_lambda_vs_mu_k_w_in_0_1_mu_continuous_phi_1.5_figure}, \ref{heatmap_lambda_vs_mu_k_w_in_0_1_mu_continuous_phi_2_figure} and \ref{heatmap_lambda_vs_mu_k_w_in_0_1_mu_continuous_phi_2.5_figure} vary $\mu_k$ on the $y$-axis (while keeping $\lambda$ on the $x$-axis, as with all others) and correspond to values of $\phi=0.5$, $\phi=1.5$, $\phi=2$ and $\phi=2.5$ respectively.

\begin{figure}
\centering
\includegraphics[width = \textwidth]{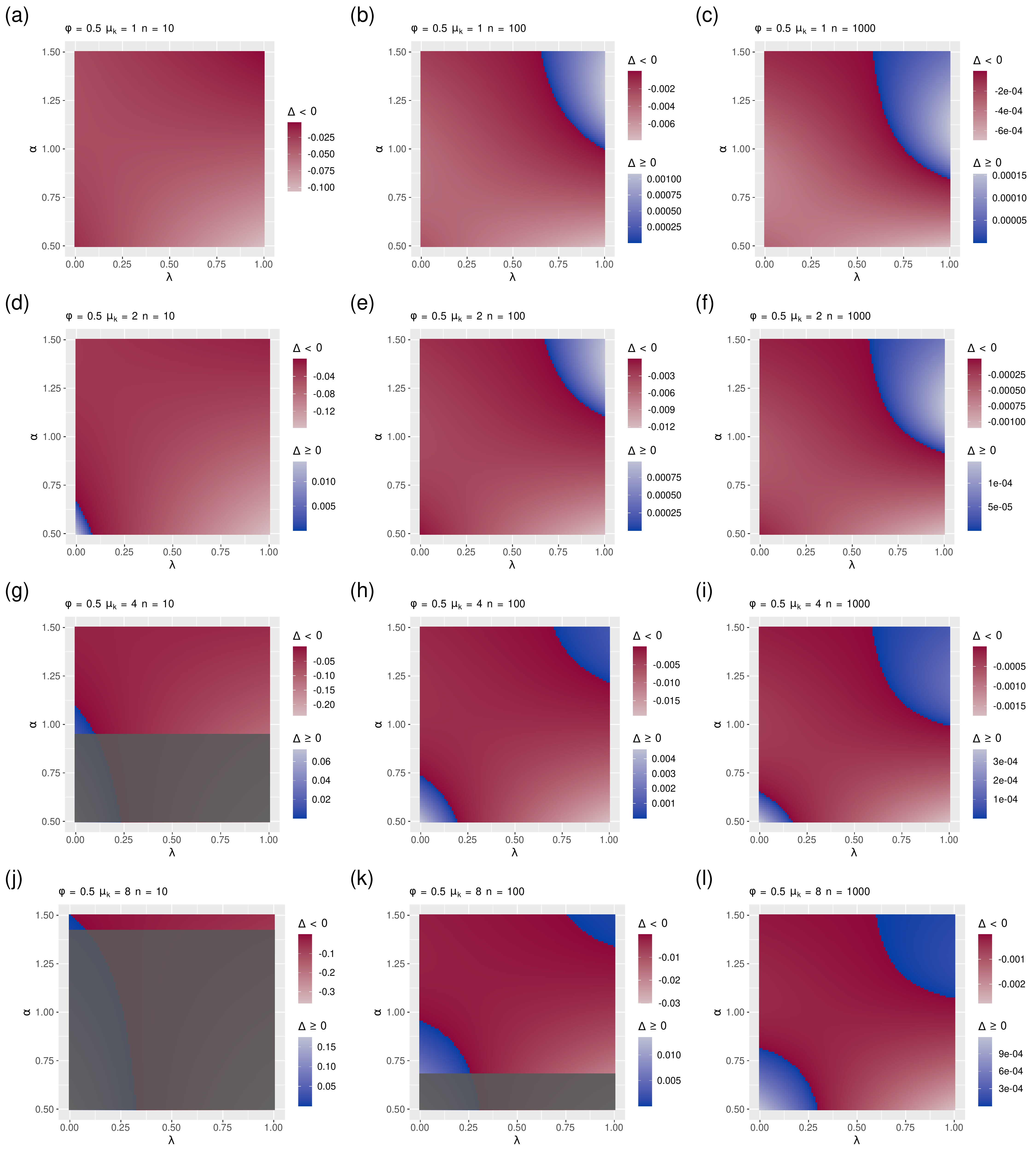}
\caption{\label{heatmap_lambda_vs_alpha_w_in_0_1_mu_continuous_phi_0.5_figure} The same as in Fig. \ref{heatmap_lambda_vs_alpha_w_in_0_1_mu_continuous_phi_1_figure} but with $\phi=0.5$.
}
\end{figure}

\begin{figure}
\centering
\includegraphics[width = \textwidth]{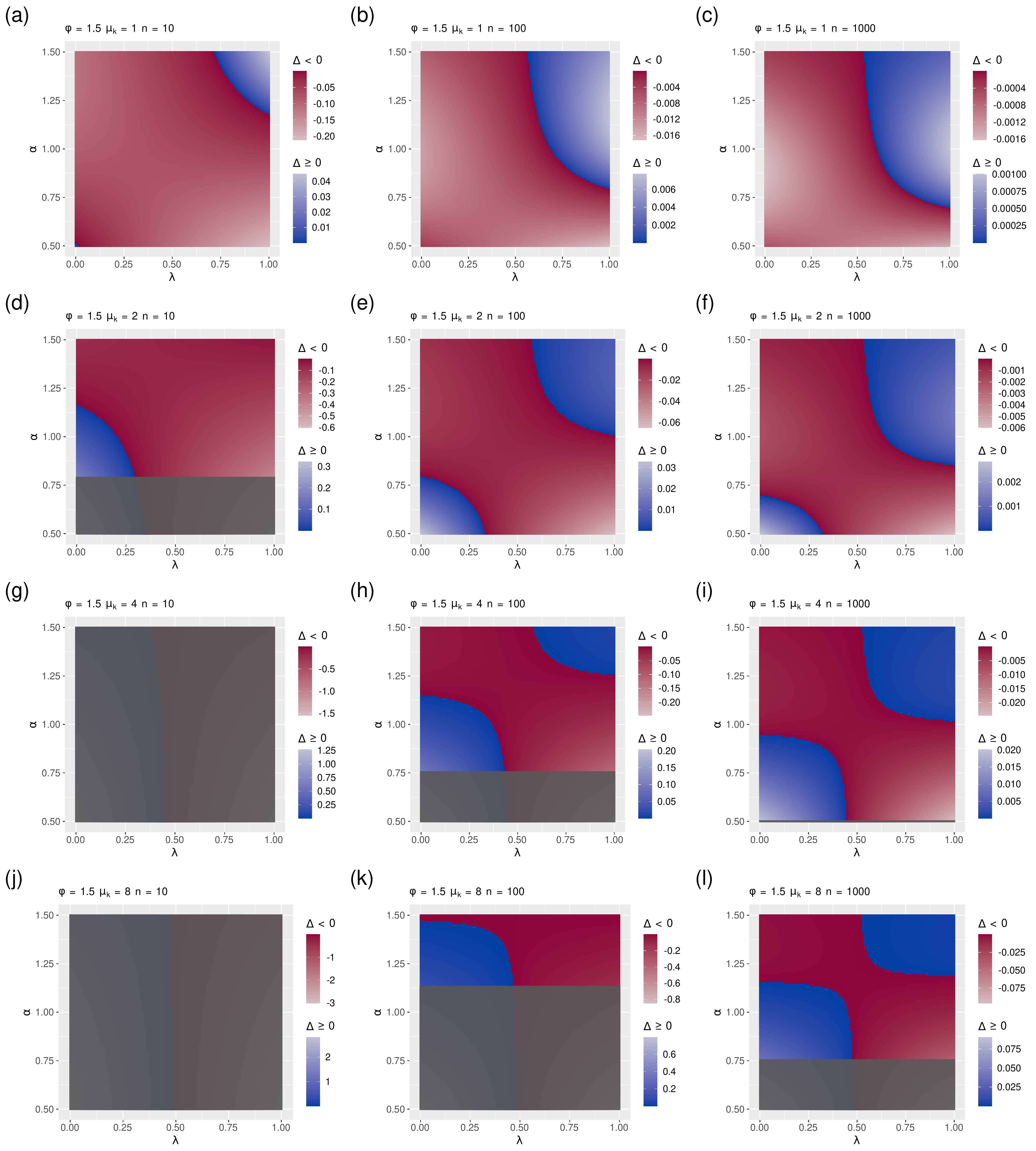}
\caption{\label{heatmap_lambda_vs_alpha_w_in_0_1_mu_continuous_phi_1.5_figure} The same as in Fig. \ref{heatmap_lambda_vs_alpha_w_in_0_1_mu_continuous_phi_1_figure} but with $\phi=1.5$.
}
\end{figure}

\begin{figure}
\centering
\includegraphics[width = \textwidth]{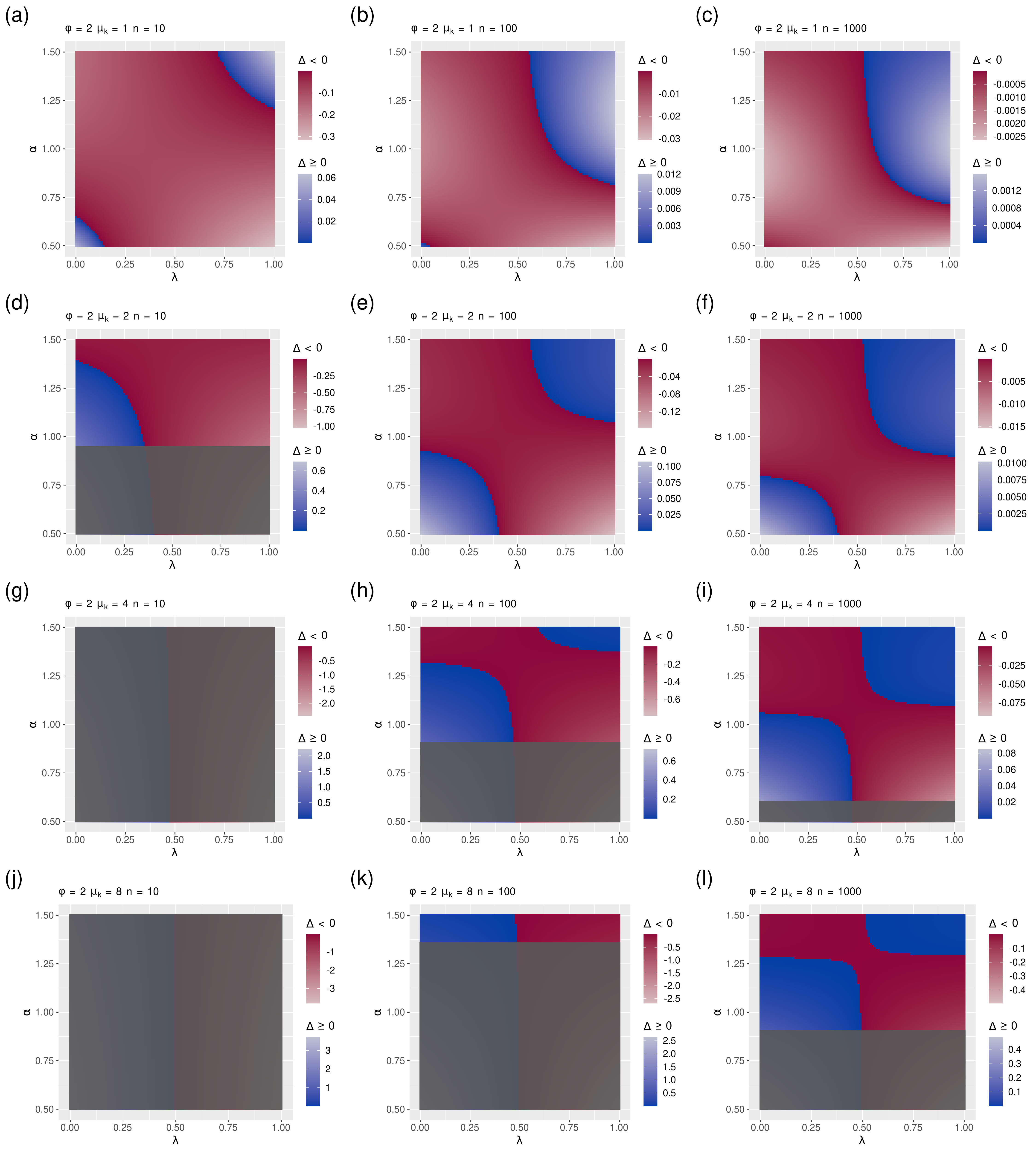}
\caption{\label{heatmap_lambda_vs_alpha_w_in_0_1_mu_continuous_phi_2_figure} The same as in Fig. \ref{heatmap_lambda_vs_alpha_w_in_0_1_mu_continuous_phi_1_figure} but with $\phi=2$.
}
\end{figure}

\begin{figure}
\centering
\includegraphics[width = \textwidth]{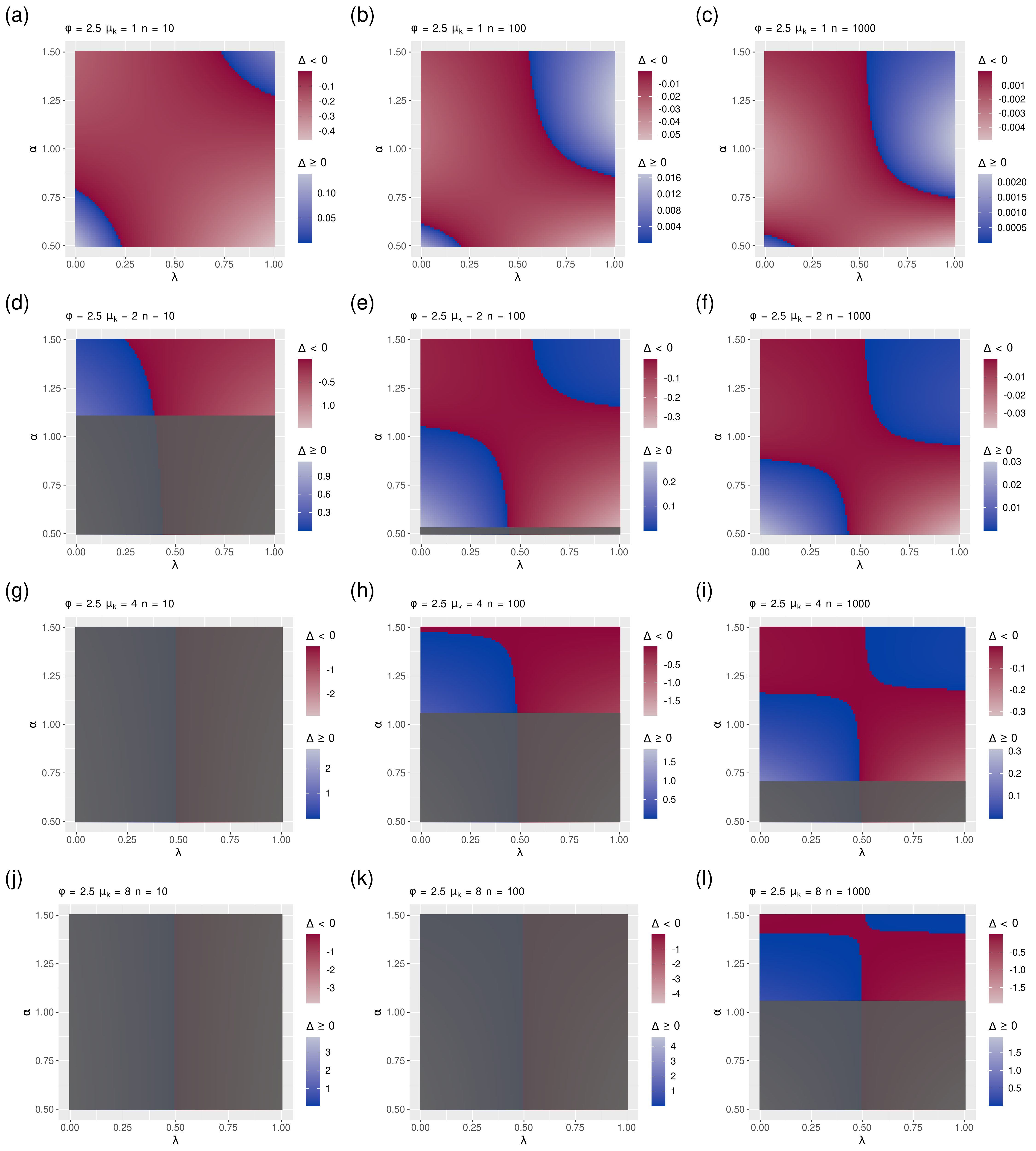}
\caption{\label{heatmap_lambda_vs_alpha_w_in_0_1_mu_continuous_phi_2.5_figure} The same as in Fig. \ref{heatmap_lambda_vs_alpha_w_in_0_1_mu_continuous_phi_1_figure} but with $\phi=2.5$.
}
\end{figure}

\paragraph{Heatmaps of $\Delta$ as a function of $\lambda$ and $\alpha$}
Figures \ref{heatmap_lambda_vs_alpha_w_in_0_1_mu_continuous_phi_0.5_figure}, \ref{heatmap_lambda_vs_alpha_w_in_0_1_mu_continuous_phi_1.5_figure}, \ref{heatmap_lambda_vs_alpha_w_in_0_1_mu_continuous_phi_2_figure} and \ref{heatmap_lambda_vs_alpha_w_in_0_1_mu_continuous_phi_2.5_figure} vary $\alpha$ on the $y$-axis and correspond to values of $\phi=0.5$, $\phi=1.5$, $\phi=2$ and $\phi=2.5$ respectively.

\begin{figure}
\centering
\includegraphics[width = \textwidth]{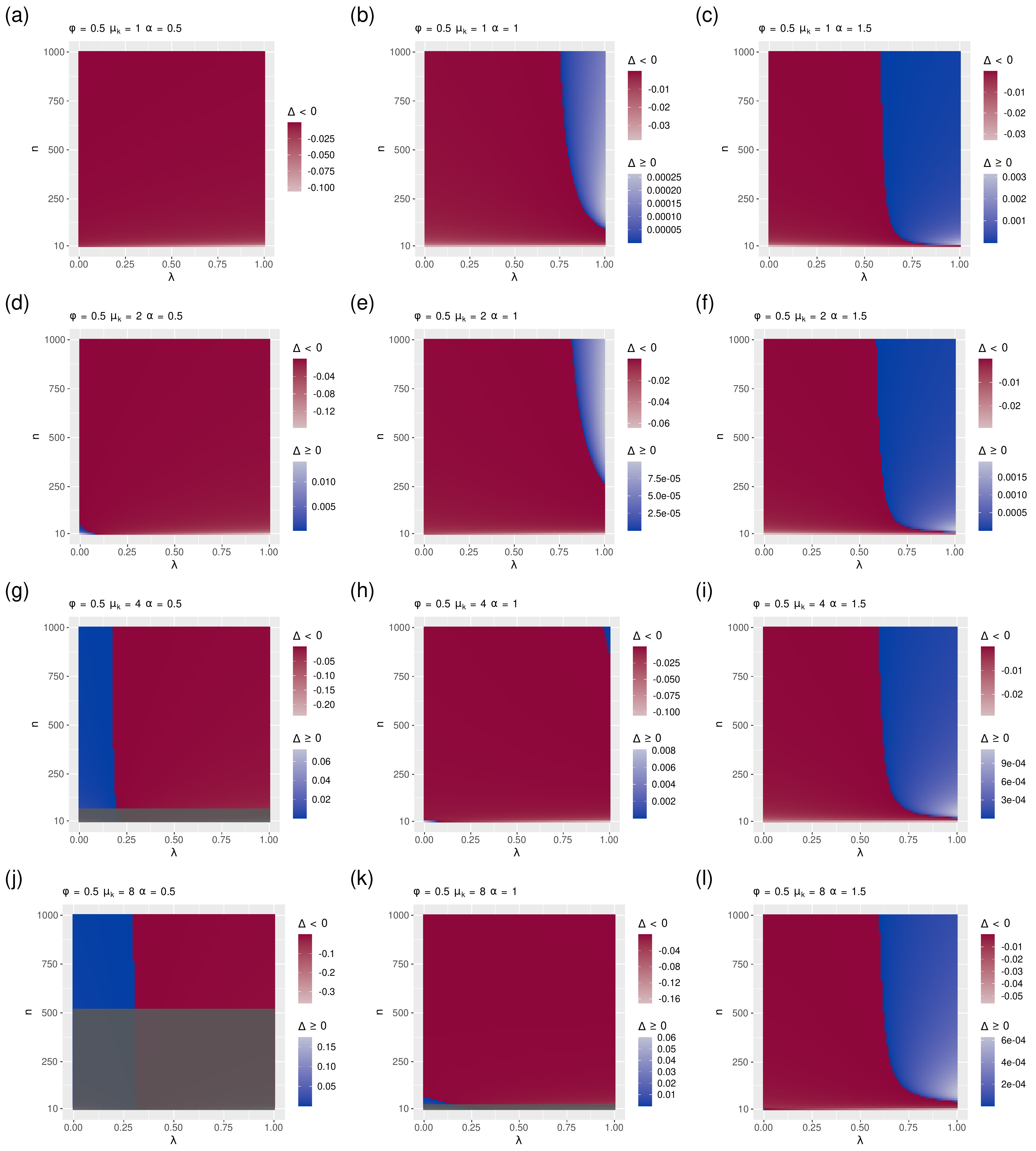}
\caption{\label{heatmap_lambda_vs_n_w_in_0_1_mu_continuous_phi_0.5_figure} The same as in Fig. \ref{heatmap_lambda_vs_n_w_in_0_1_mu_continuous_phi_1_figure} but with $\phi=0.5$.
}
\end{figure}

\begin{figure}
\centering
\includegraphics[width = \textwidth]{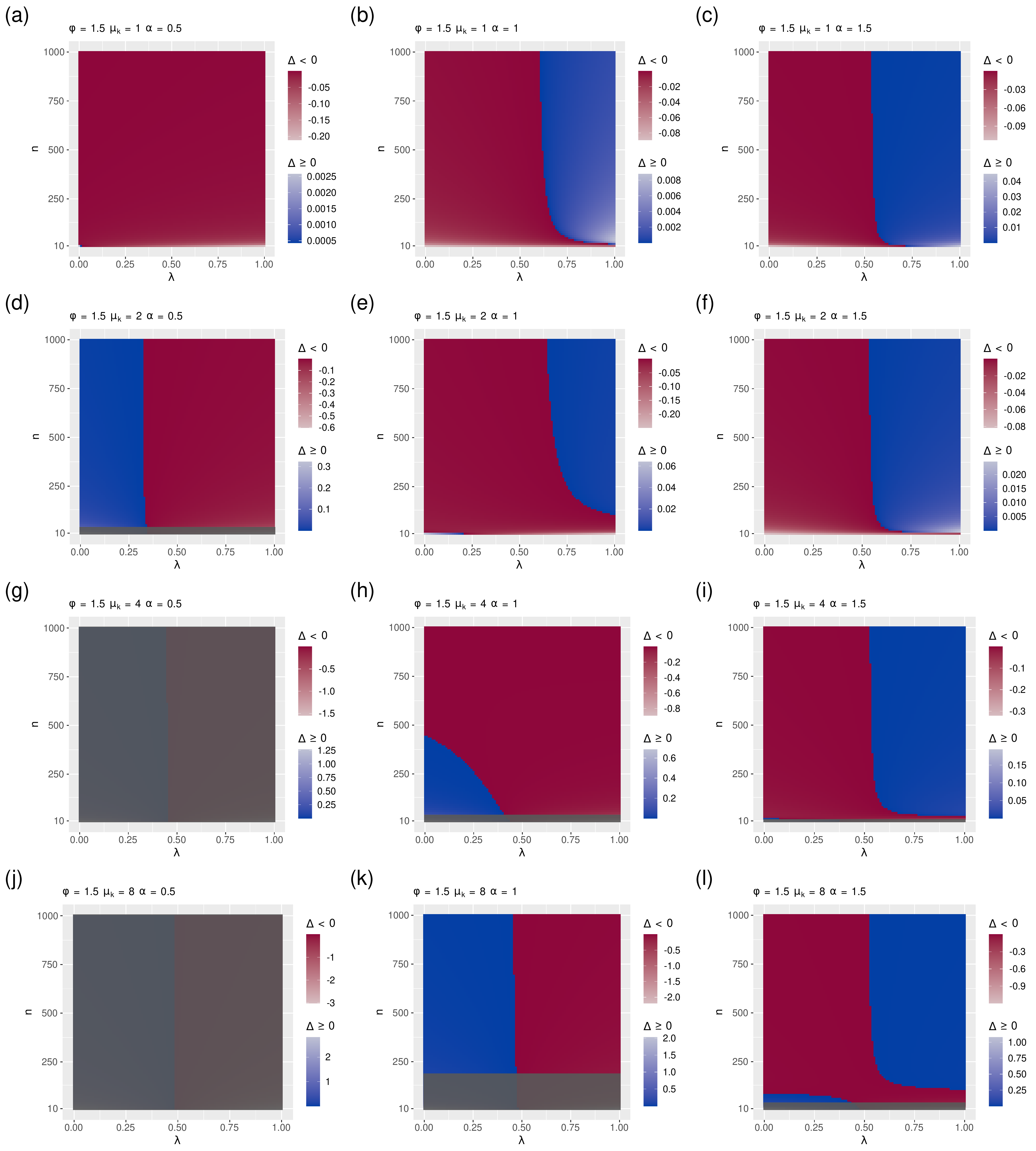}
\caption{\label{heatmap_lambda_vs_n_w_in_0_1_mu_continuous_phi_1.5_figure} The same as in Fig. \ref{heatmap_lambda_vs_n_w_in_0_1_mu_continuous_phi_1_figure} but with $\phi=1.5$.
}
\end{figure}

\begin{figure}
\centering
\includegraphics[width = \textwidth]{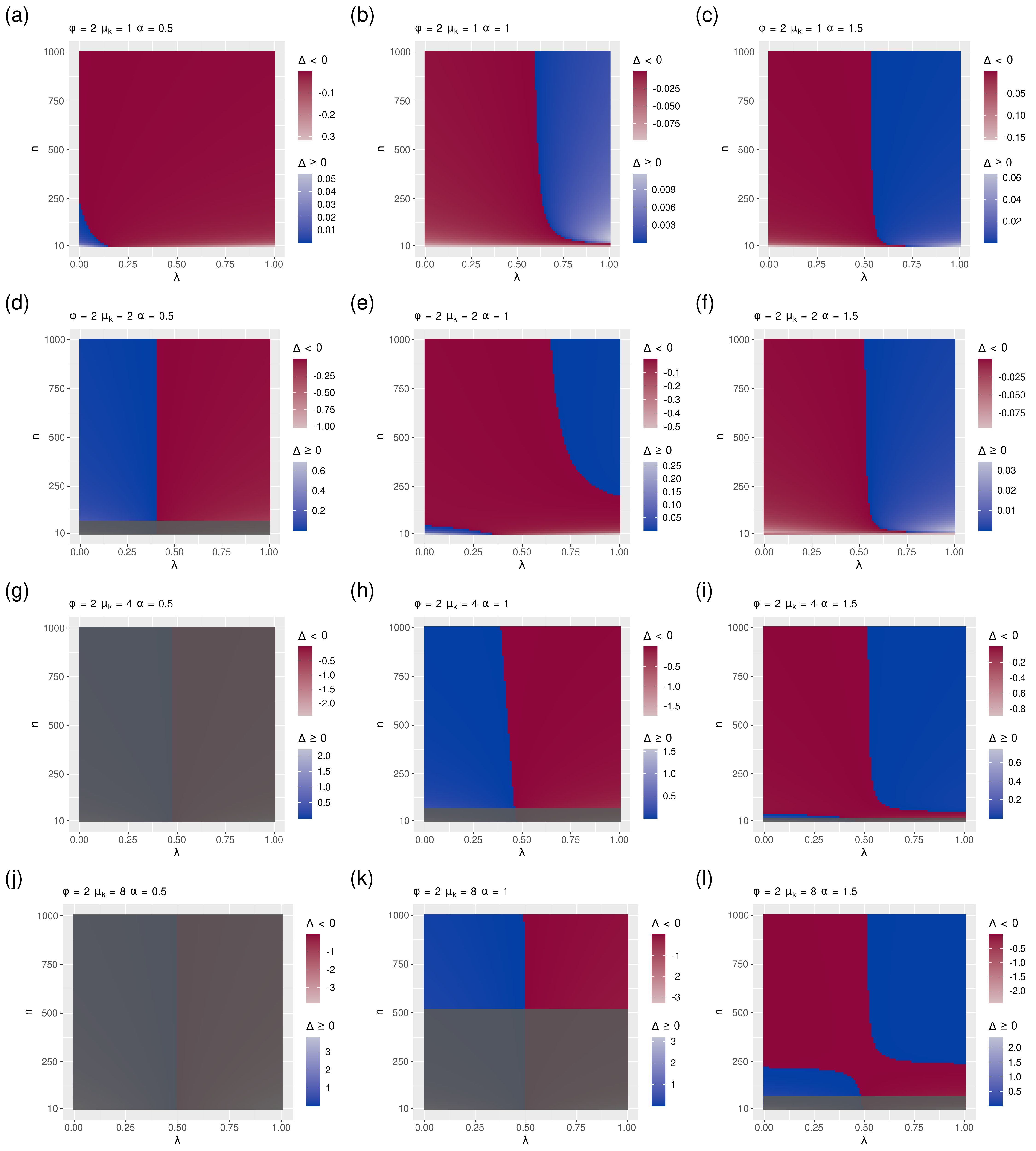}
\caption{\label{heatmap_lambda_vs_n_w_in_0_1_mu_continuous_phi_2_figure} The same as in Fig. \ref{heatmap_lambda_vs_n_w_in_0_1_mu_continuous_phi_1_figure} but with $\phi=2$.
}
\end{figure}

\begin{figure}
\centering
\includegraphics[width = \textwidth]{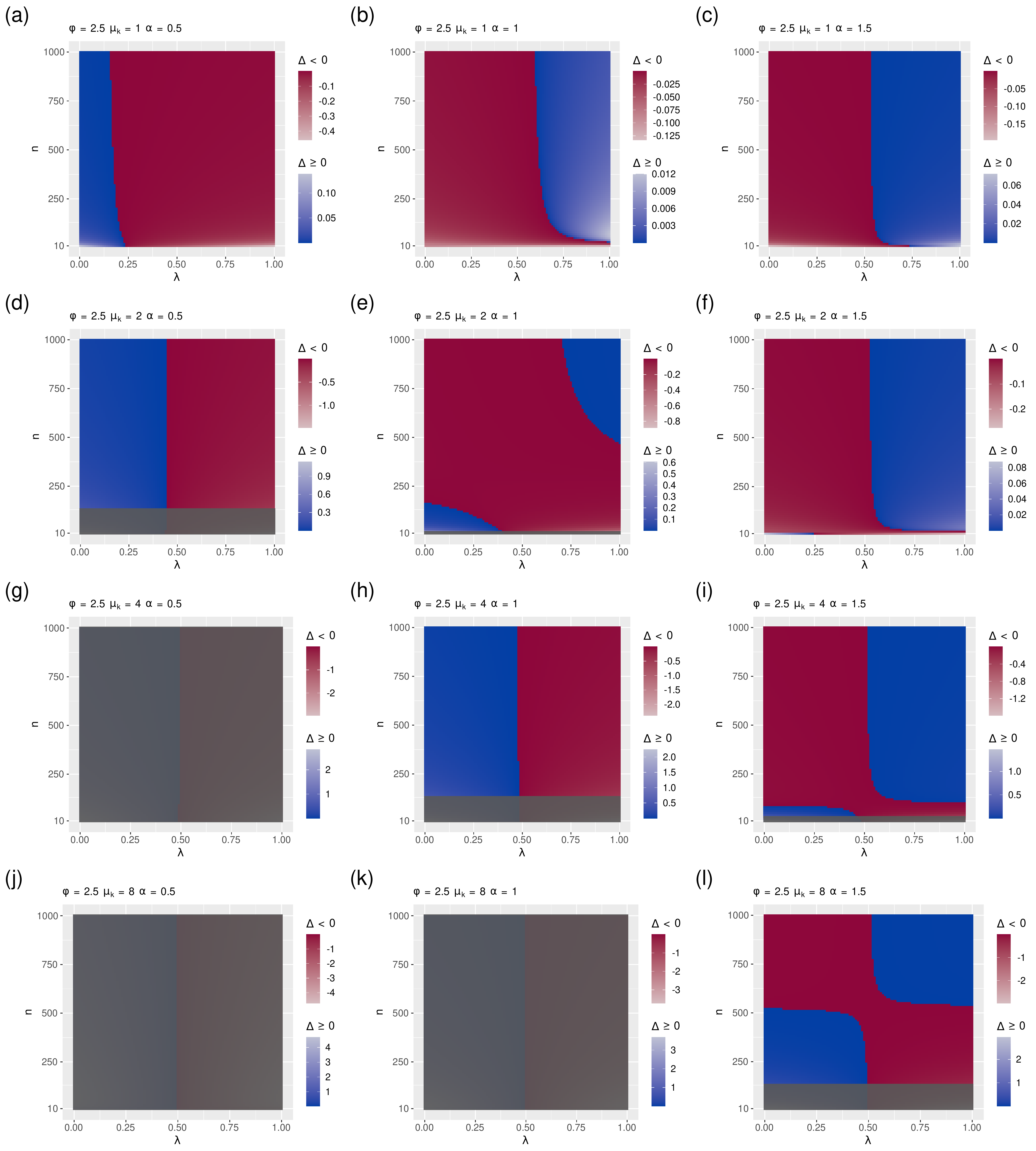}
\caption{\label{heatmap_lambda_vs_n_w_in_0_1_mu_continuous_phi_2.5_figure} The same as in Fig. \ref{heatmap_lambda_vs_n_w_in_0_1_mu_continuous_phi_1_figure} but with $\phi=2.5$.
}
\end{figure}

\paragraph{Heatmaps of $\Delta$ as a function of $\lambda$ and $n$}
Figures \ref{heatmap_lambda_vs_n_w_in_0_1_mu_continuous_phi_0.5_figure}, \ref{heatmap_lambda_vs_n_w_in_0_1_mu_continuous_phi_1.5_figure}, \ref{heatmap_lambda_vs_n_w_in_0_1_mu_continuous_phi_2_figure} and \ref{heatmap_lambda_vs_n_w_in_0_1_mu_continuous_phi_2.5_figure} vary $n$ on the $y$-axis and correspond to values of $\phi=0.5$, $\phi=1.5$, $\phi=2$ and $\phi=2.5$ respectively.


%% file: discrete_degrees_appendix.tex
To investigate the class of skeleta such that the degree of counterparts does not exceed one, we have assumed that the relationship between the degree of a vertex and its rank follows a power-law (Eq. \ref{degree_versus_rank_equation}). For the plots of the regions where strategy $a$ is advantageous, we have assumed, for simplicity, that the degree of a form is a continuous variable. As form degrees are actually discrete in the model, here we show the impact of rounding form degrees defined by Eq. \ref{degree_versus_rank_equation} to the nearest integer in previous figures.

\begin{figure}
\centering
\includegraphics[width = \textwidth]{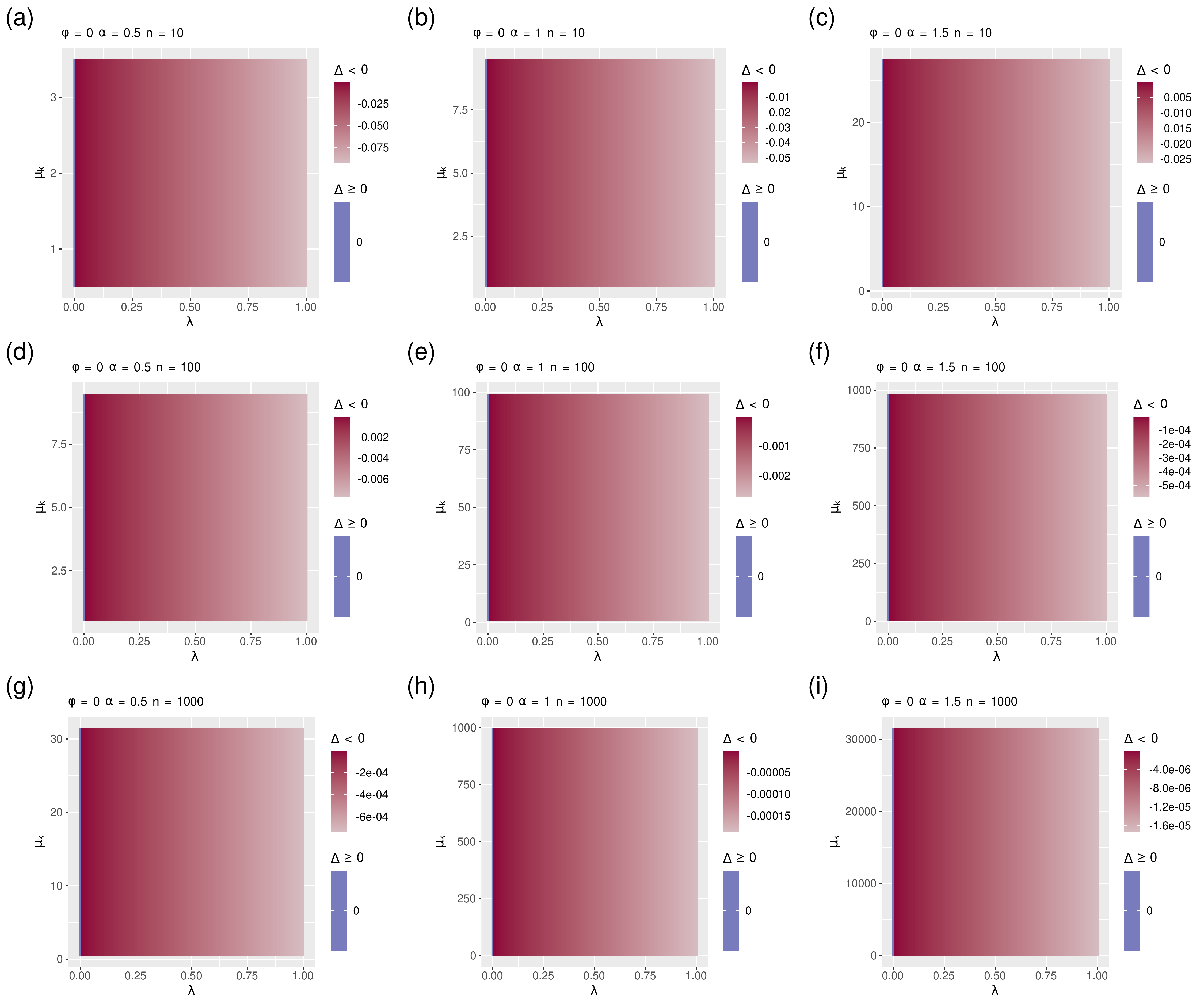}
\caption{\label{heatmap_lambda_vs_mu_k_w_in_0_1_phi_0_figure} Figure equivalent to Fig. \ref{heatmap_lambda_vs_mu_k_w_in_0_1_mu_continuous_phi_0_figure} after discretization of the $\mu_i'$s.
}
\end{figure}

\begin{figure}
\centering
\includegraphics[width = \textwidth]{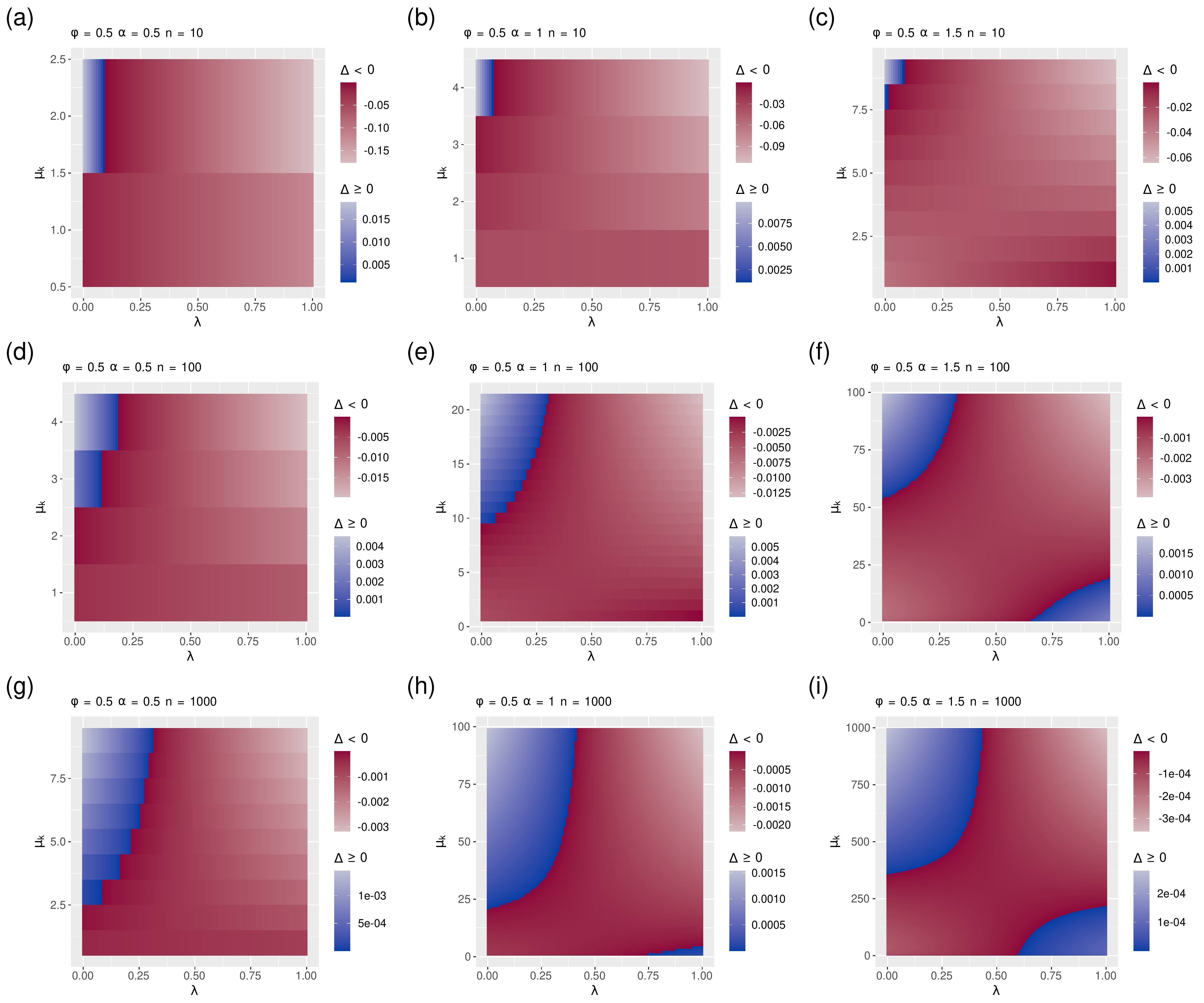}
\caption{\label{heatmap_lambda_vs_mu_k_w_in_0_1_phi_0.5_figure} Figure equivalent to Fig. \ref{heatmap_lambda_vs_mu_k_w_in_0_1_mu_continuous_phi_0.5_figure} after discretization of the $\mu_i'$s.
}
\end{figure}

\begin{figure}
\centering
\includegraphics[width = \textwidth]{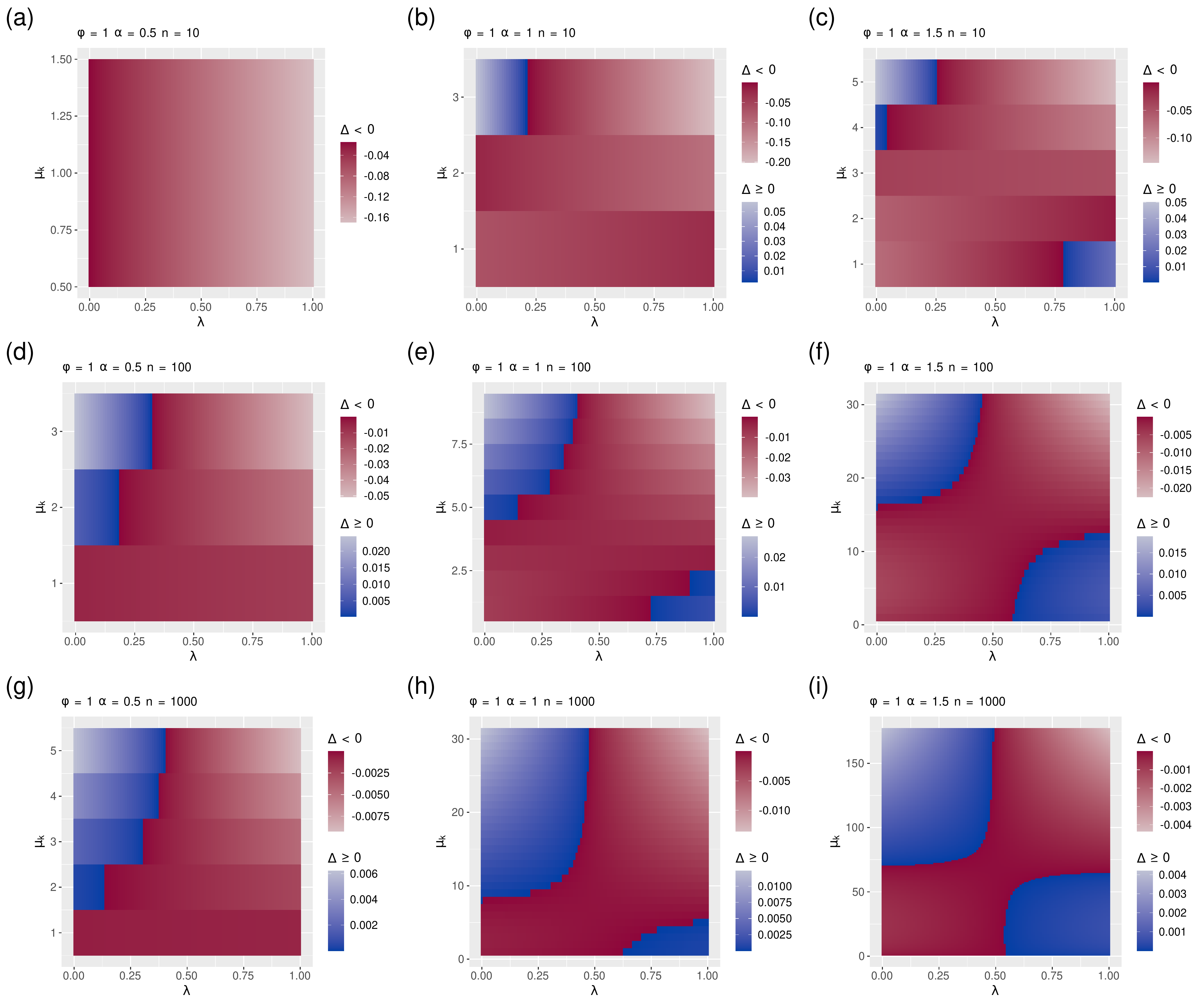}
\caption{\label{heatmap_lambda_vs_mu_k_w_in_0_1_phi_1_figure} Figure equivalent to Fig. \ref{heatmap_lambda_vs_mu_k_w_in_0_1_mu_continuous_phi_1_figure} after discretization of the $\mu_i'$s.
}
\end{figure}

\begin{figure}
\centering
\includegraphics[width = \textwidth]{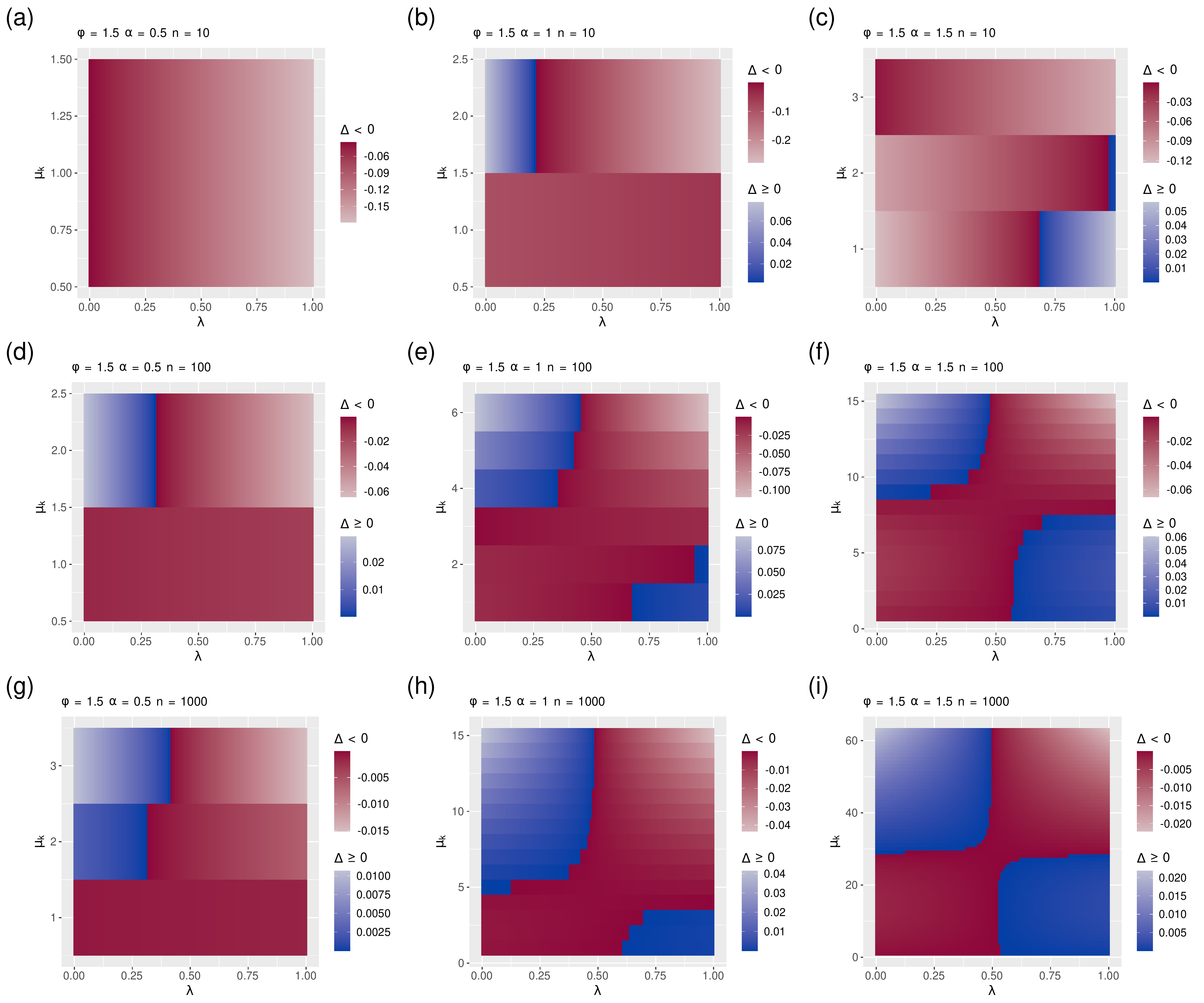}
\caption{\label{heatmap_lambda_vs_mu_k_w_in_0_1_phi_1.5_figure} Figure equivalent to Fig. \ref{heatmap_lambda_vs_mu_k_w_in_0_1_mu_continuous_phi_1.5_figure} after discretization of the $\mu_i'$s.
}
\end{figure}

\begin{figure}
\centering
\includegraphics[width = \textwidth]{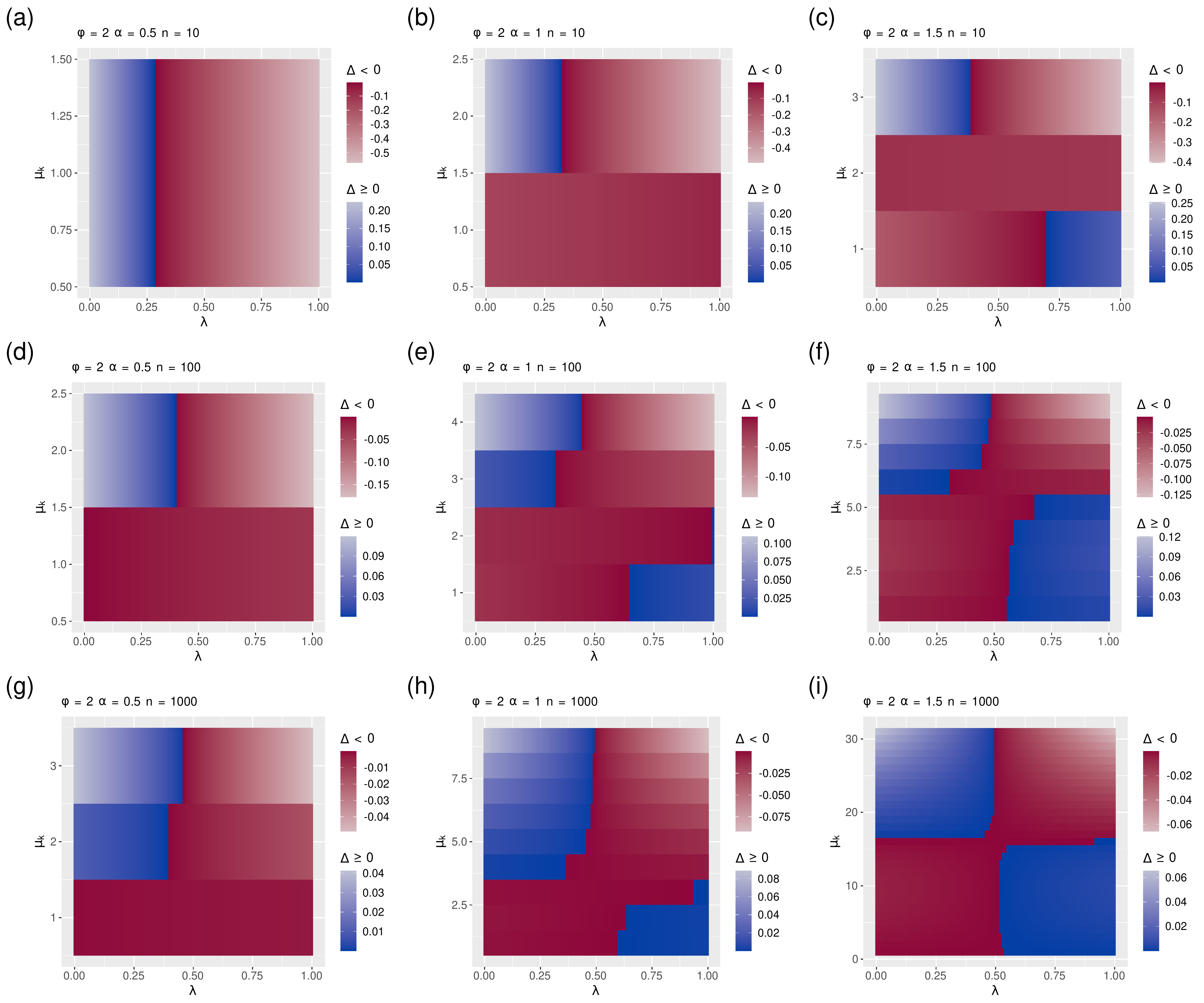}
\caption{\label{heatmap_lambda_vs_mu_k_w_in_0_1_phi_2_figure} Figure equivalent to Fig. \ref{heatmap_lambda_vs_mu_k_w_in_0_1_mu_continuous_phi_2_figure} after discretization of the $\mu_i'$s.
}
\end{figure}

\begin{figure}
\centering
\includegraphics[width = \textwidth]{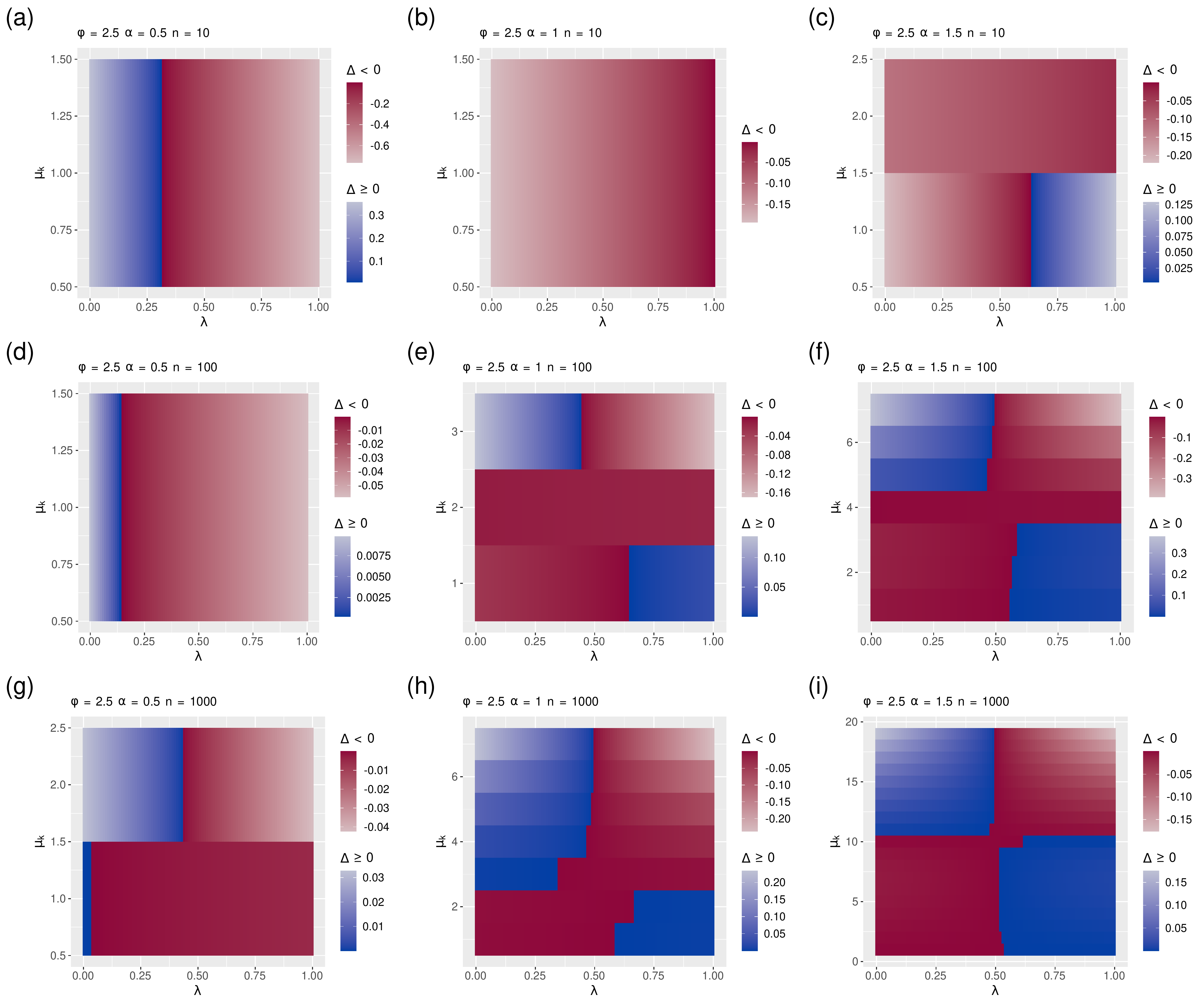}
\caption{\label{heatmap_lambda_vs_mu_k_w_in_0_1_phi_2.5_figure} Figure equivalent to Fig. \ref{heatmap_lambda_vs_mu_k_w_in_0_1_mu_continuous_phi_2.5_figure} after discretization of the $\mu_i'$s.
}
\end{figure}

\begin{figure}
\centering
\includegraphics[width = \textwidth]{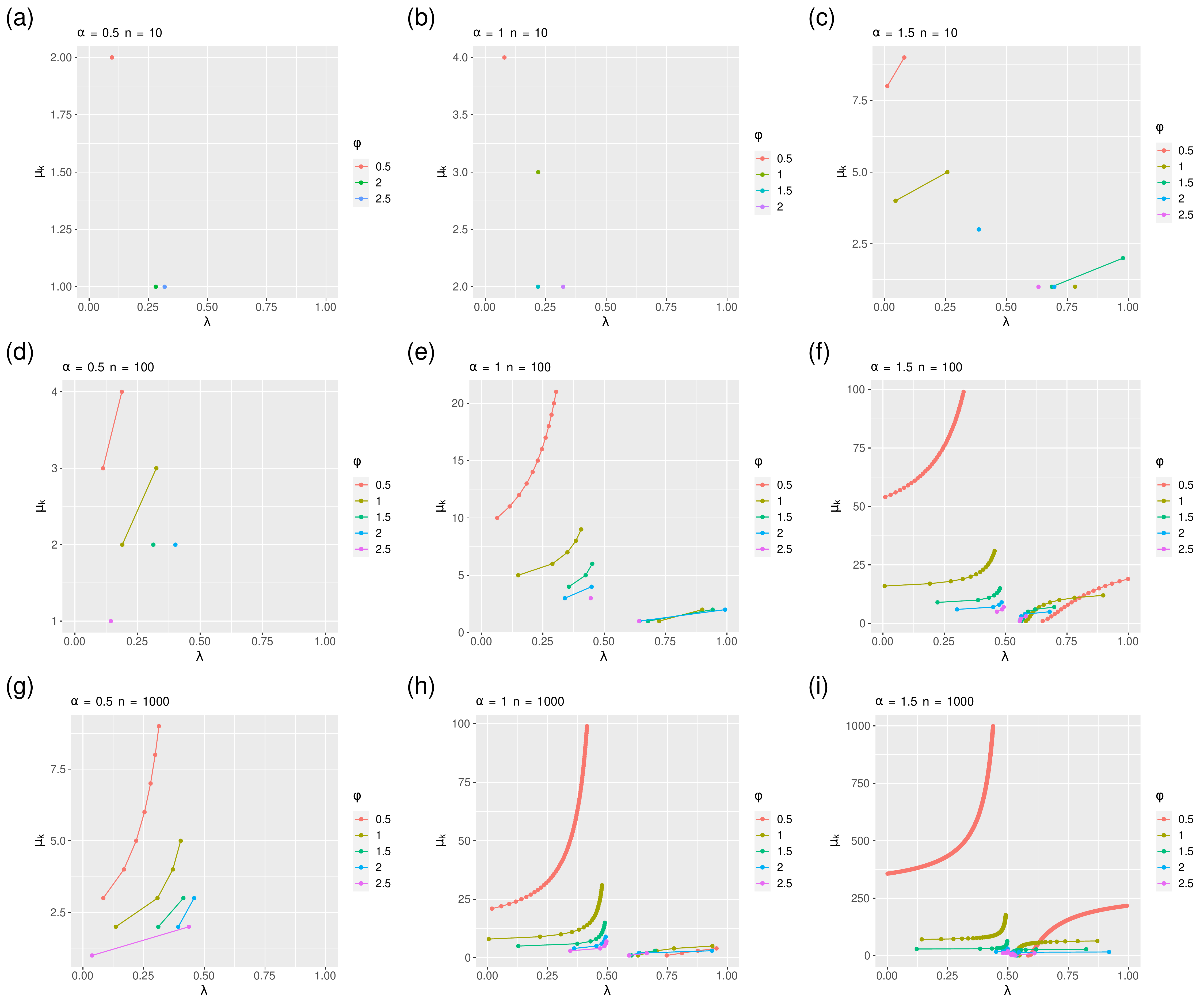}
\caption{\label{curves_delta_0_lambda_vs_mu_k_w_in_0_1_various_phi_figure} Figure equivalent to Fig. \ref{curves_delta_0_lambda_vs_mu_k_w_in_0_1_mu_continuous_various_phi_figure} after discretization of the $\mu_i'$s. It summarizes Figs.
\ref{heatmap_lambda_vs_mu_k_w_in_0_1_phi_0.5_figure}, 
\ref{heatmap_lambda_vs_mu_k_w_in_0_1_phi_1_figure}, 
\ref{heatmap_lambda_vs_mu_k_w_in_0_1_phi_1.5_figure}, 
\ref{heatmap_lambda_vs_mu_k_w_in_0_1_phi_2_figure} and  \ref{heatmap_lambda_vs_mu_k_w_in_0_1_phi_2.5_figure}.}
\end{figure}

\begin{figure}
\centering
\includegraphics[width = \textwidth]{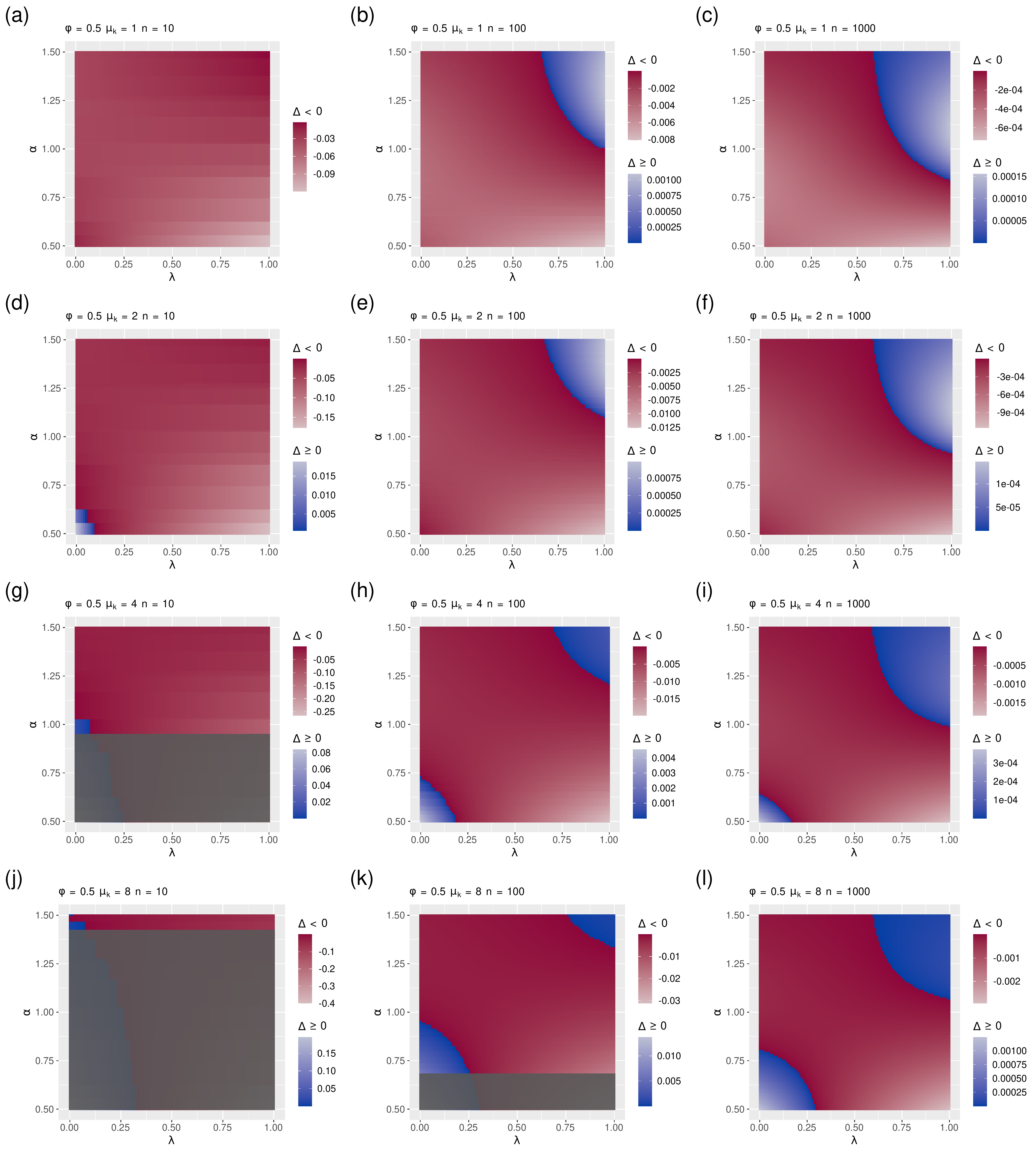}
\caption{\label{heatmap_lambda_vs_alpha_w_in_0_1_phi_0.5_figure} Figure equivalent to Fig. \ref{heatmap_lambda_vs_alpha_w_in_0_1_mu_continuous_phi_0.5_figure} after discretization of the $\mu_i'$s.
}
\end{figure}

\begin{figure}
\centering
\includegraphics[width = \textwidth]{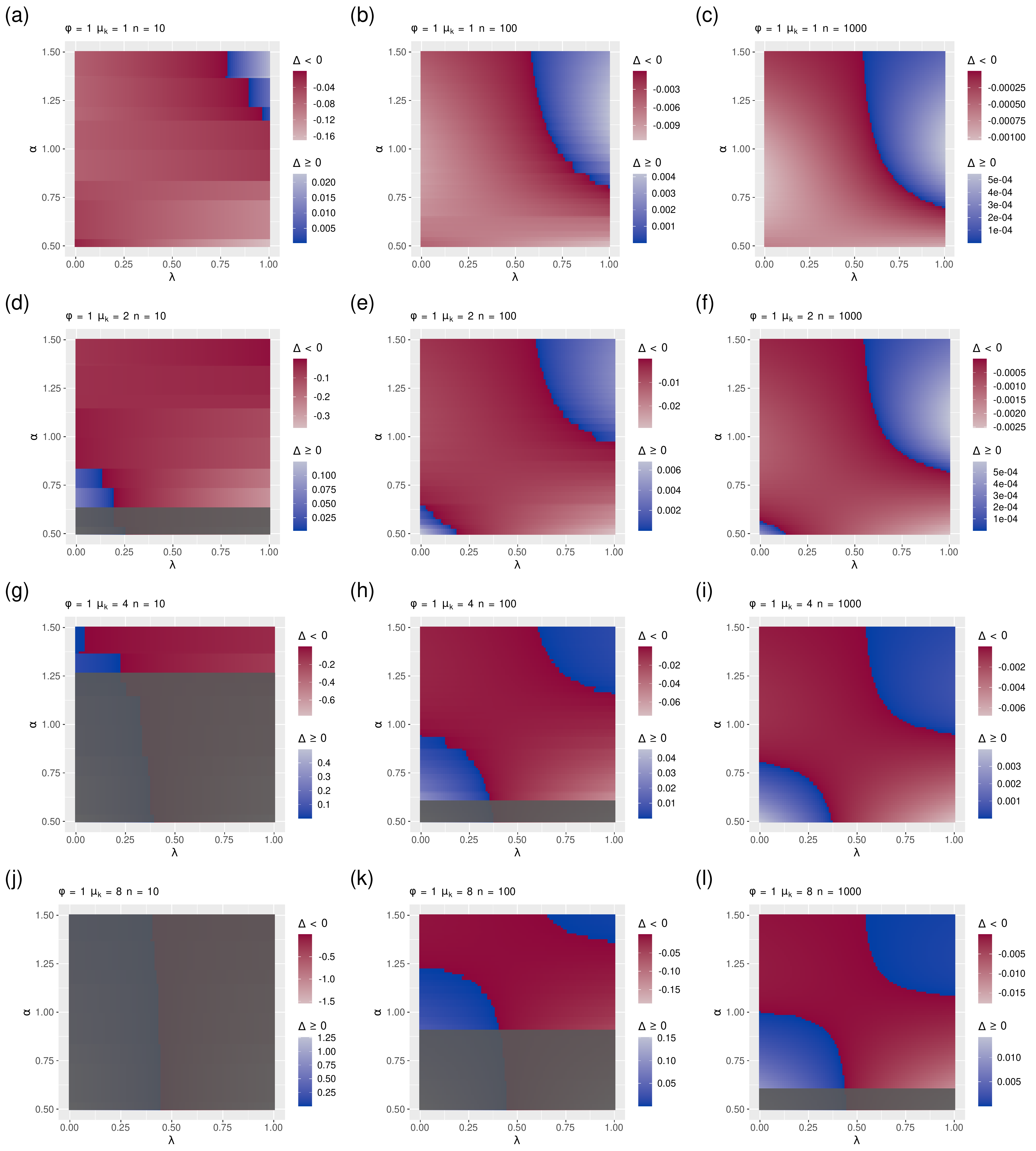}
\caption{\label{heatmap_lambda_vs_alpha_w_in_0_1_phi_1_figure} Figure equivalent to Fig. \ref{heatmap_lambda_vs_alpha_w_in_0_1_mu_continuous_phi_1_figure} after discretization of the $\mu_i'$s.
}
\end{figure}

\begin{figure}
\centering
\includegraphics[width = \textwidth]{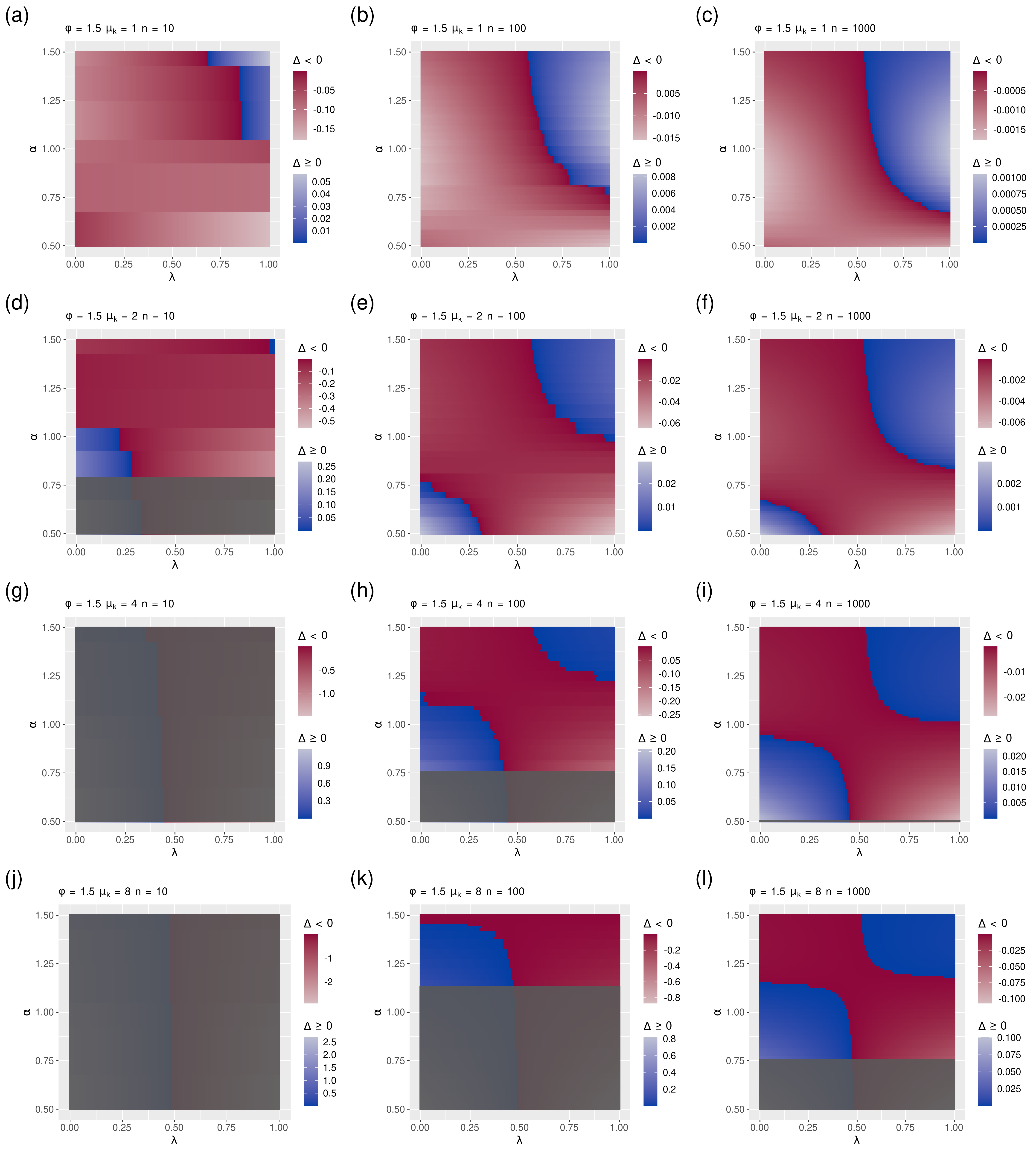}
\caption{\label{heatmap_lambda_vs_alpha_w_in_0_1_phi_1.5_figure} Figure equivalent to Fig. \ref{heatmap_lambda_vs_alpha_w_in_0_1_mu_continuous_phi_1.5_figure} after discretization of the $\mu_i'$s.
}
\end{figure}

\begin{figure}
\centering
\includegraphics[width = \textwidth]{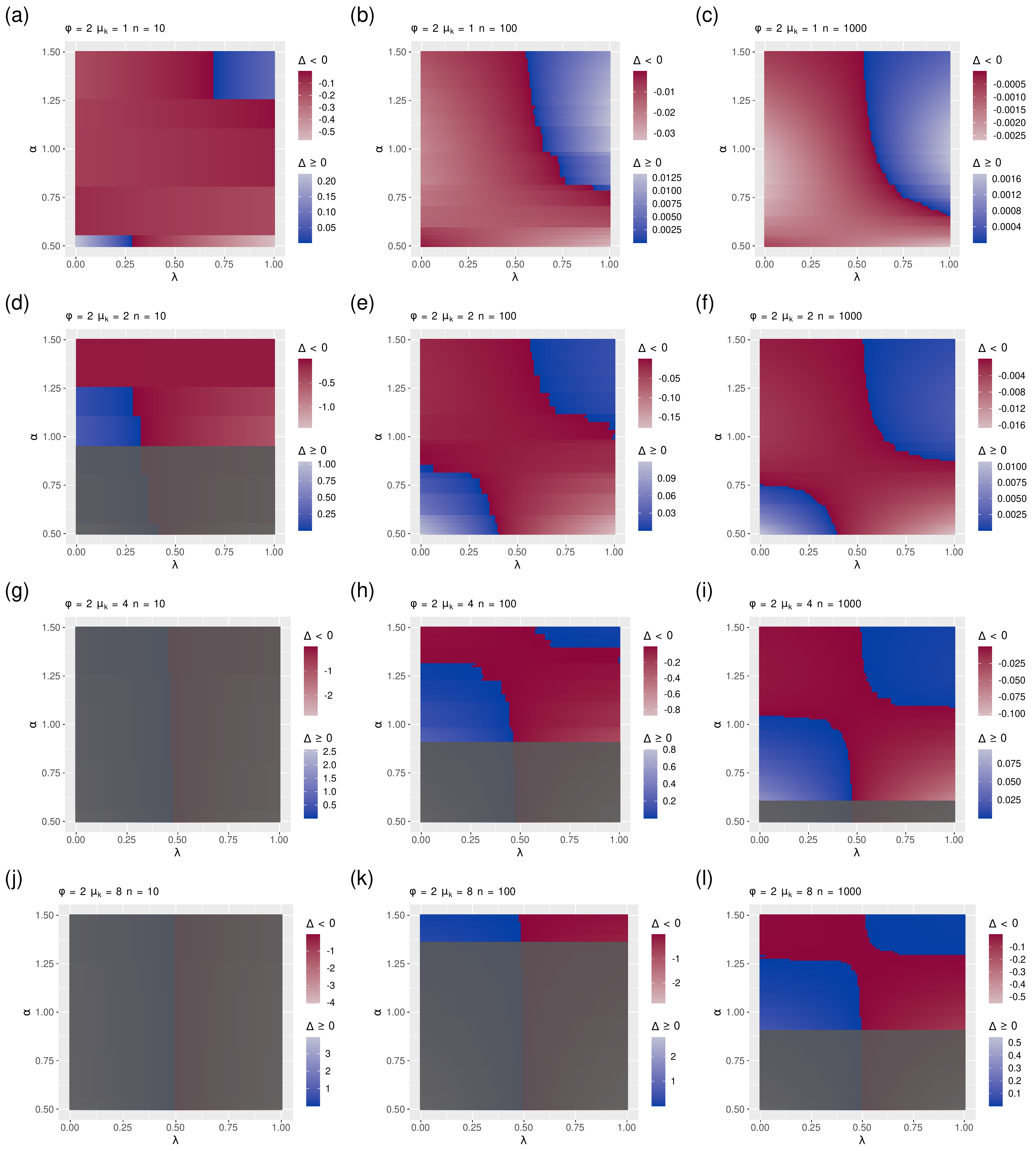}
\caption{\label{heatmap_lambda_vs_alpha_w_in_0_1_phi_2_figure} Figure equivalent to Fig. \ref{heatmap_lambda_vs_alpha_w_in_0_1_mu_continuous_phi_2_figure} after discretization of the $\mu_i'$s.
}
\end{figure}

\begin{figure}
\centering
\includegraphics[width = \textwidth]{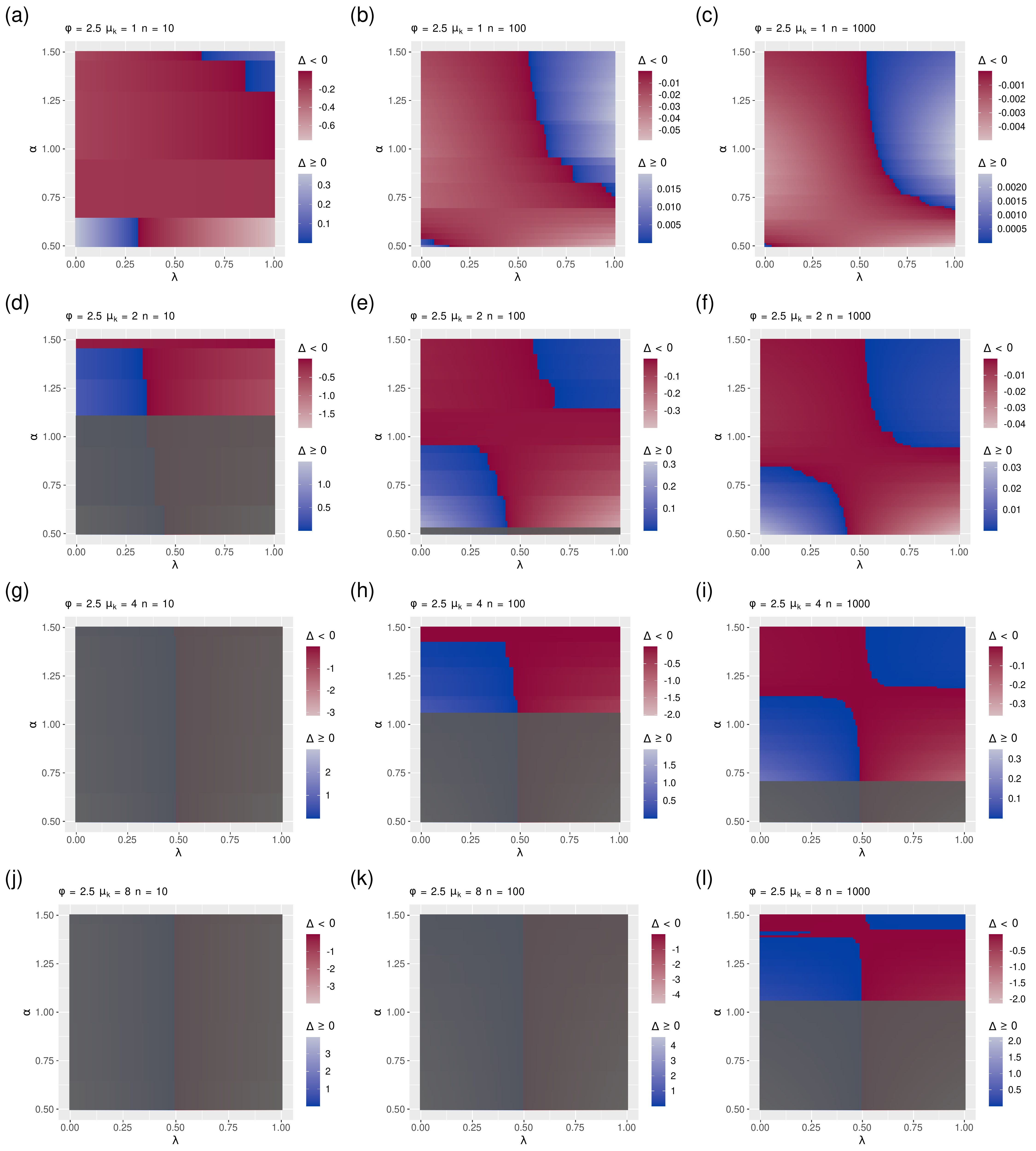}
\caption{\label{heatmap_lambda_vs_alpha_w_in_0_1_phi_2.5_figure} Figure equivalent to Fig. \ref{heatmap_lambda_vs_alpha_w_in_0_1_mu_continuous_phi_2.5_figure} after discretization of the $\mu_i'$s.
}
\end{figure}

\begin{figure}
\centering
\includegraphics[width = \textwidth]{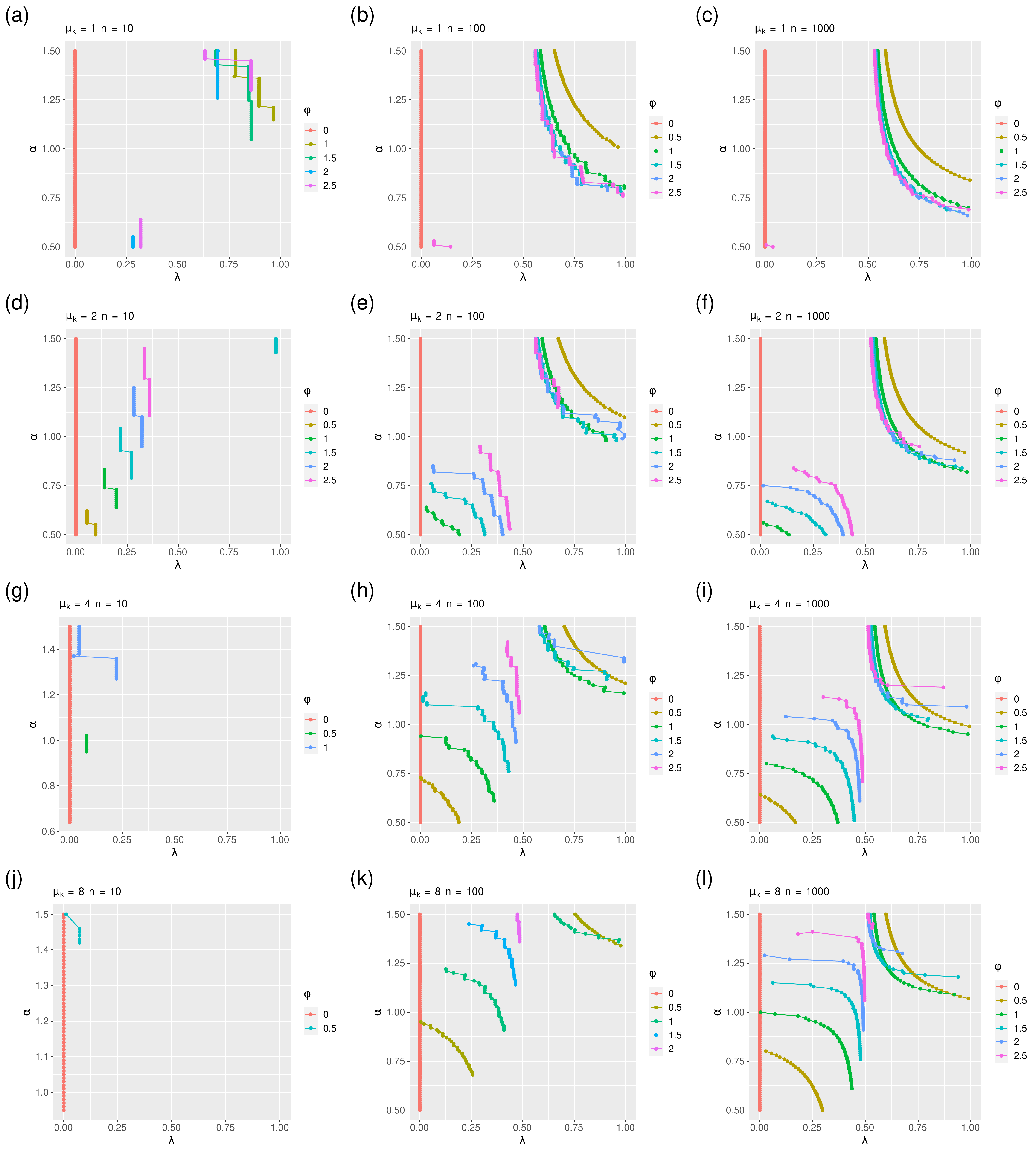}
\caption{\label{curves_delta_0_lambda_vs_alpha_w_in_0_1_various_phi_figure} Figure equivalent to Fig. \ref{curves_delta_0_lambda_vs_alpha_w_in_0_1_mu_continuous_various_phi_figure} after discretization of the $\mu_i'$s. It summarizes Figs. \ref{heatmap_lambda_vs_alpha_w_in_0_1_phi_0.5_figure},  \ref{heatmap_lambda_vs_alpha_w_in_0_1_phi_1_figure},  \ref{heatmap_lambda_vs_alpha_w_in_0_1_phi_1.5_figure},  \ref{heatmap_lambda_vs_alpha_w_in_0_1_phi_2_figure} and \ref{heatmap_lambda_vs_alpha_w_in_0_1_phi_2.5_figure}.
}
\end{figure}

\begin{figure}
\centering
\includegraphics[width = \textwidth]{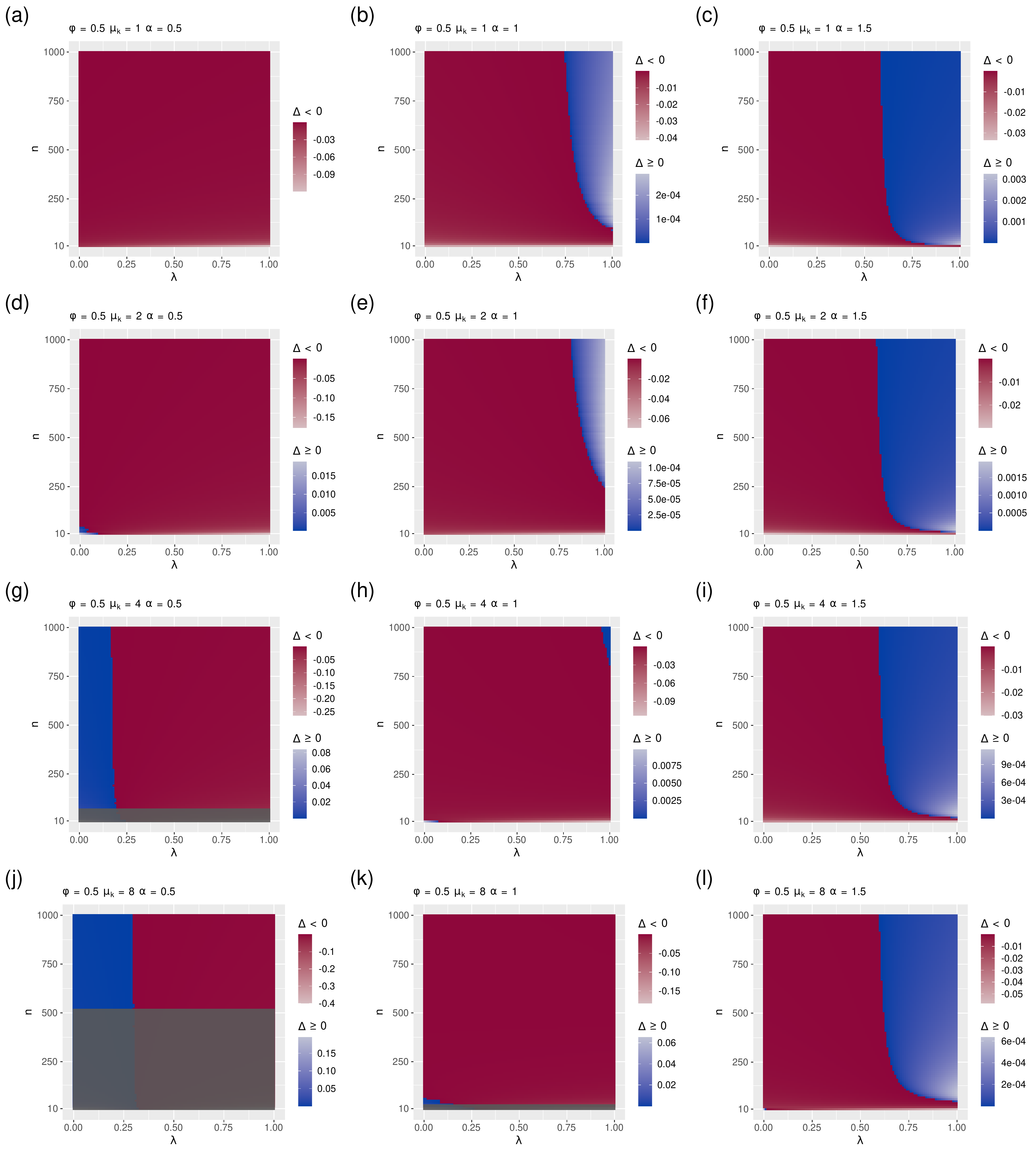}
\caption{\label{heatmap_lambda_vs_n_w_in_0_1_phi_0.5_figure} Figure equivalent to Fig. \ref{heatmap_lambda_vs_n_w_in_0_1_mu_continuous_phi_0.5_figure} after discretization of the $\mu_i'$s.
}
\end{figure}

\begin{figure}
\centering
\includegraphics[width = \textwidth]{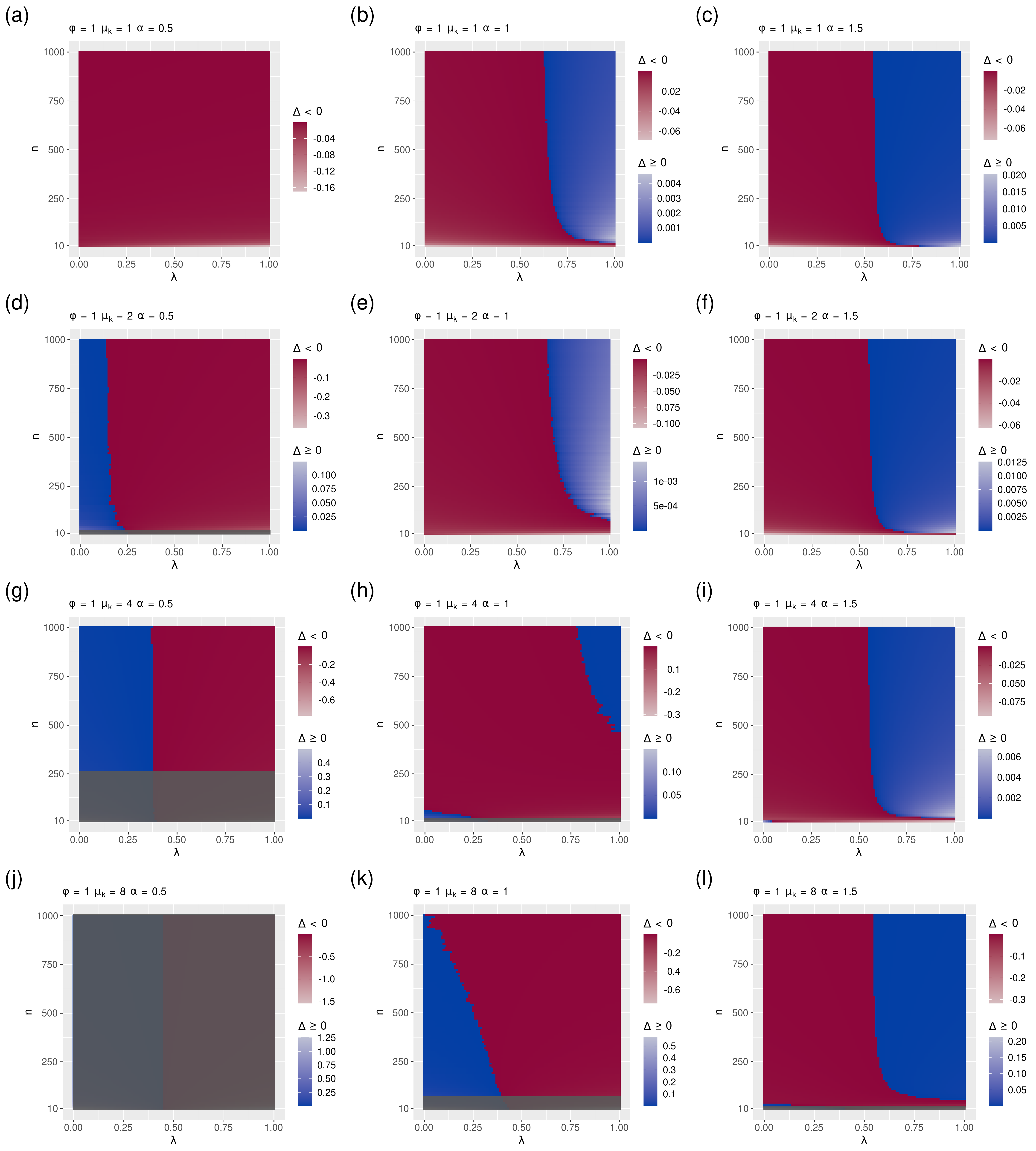}
\caption{\label{heatmap_lambda_vs_n_w_in_0_1_phi_1_figure} Figure equivalent to Fig. \ref{heatmap_lambda_vs_n_w_in_0_1_mu_continuous_phi_1_figure} after discretization of the $\mu_i'$s.
}
\end{figure}

\begin{figure}
\centering
\includegraphics[width = \textwidth]{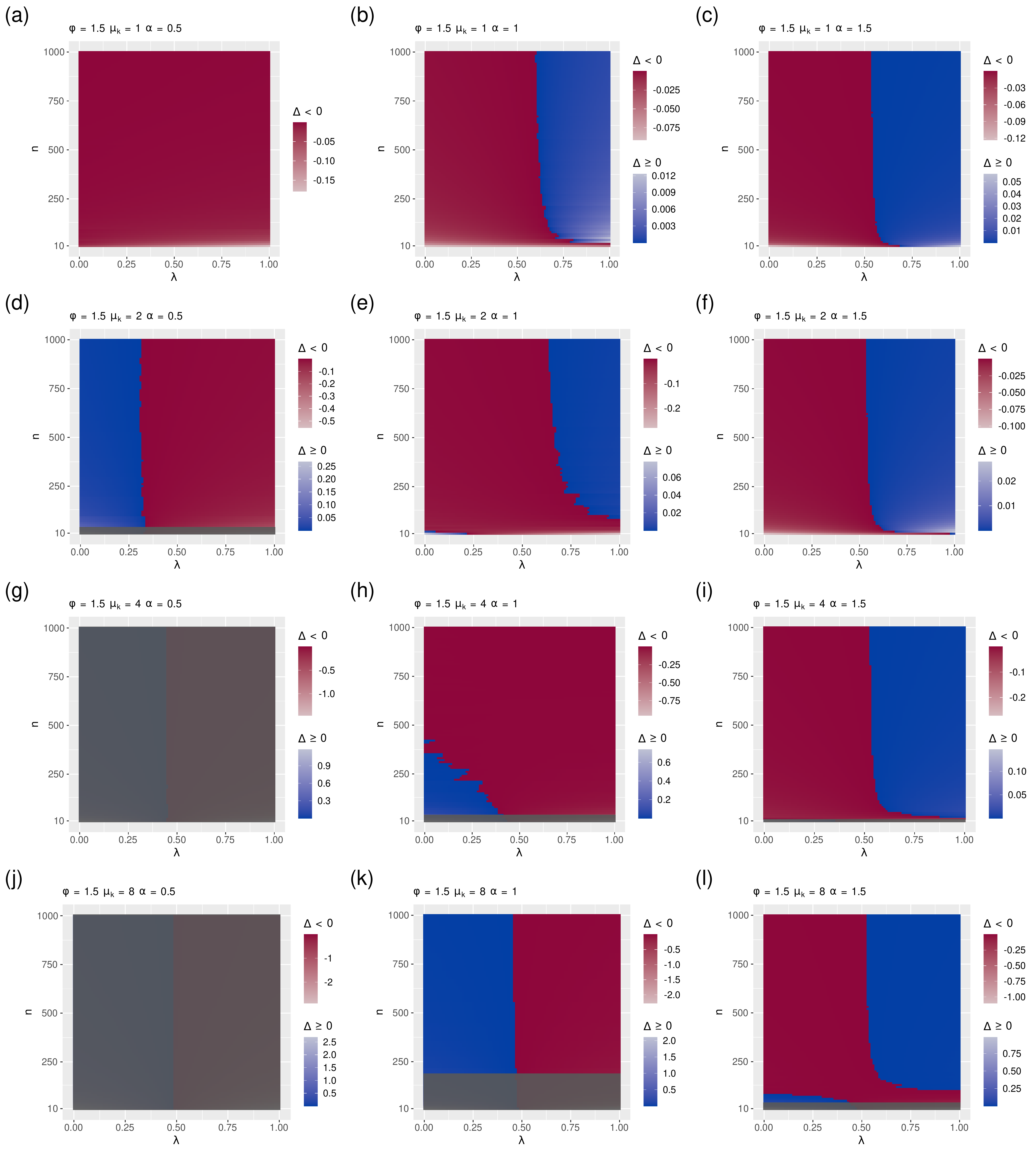}
\caption{\label{heatmap_lambda_vs_n_w_in_0_1_phi_1.5_figure} Figure equivalent to Fig. \ref{heatmap_lambda_vs_n_w_in_0_1_mu_continuous_phi_1.5_figure} after discretization of the $\mu_i'$s.
}
\end{figure}

\begin{figure}
\centering
\includegraphics[width = \textwidth]{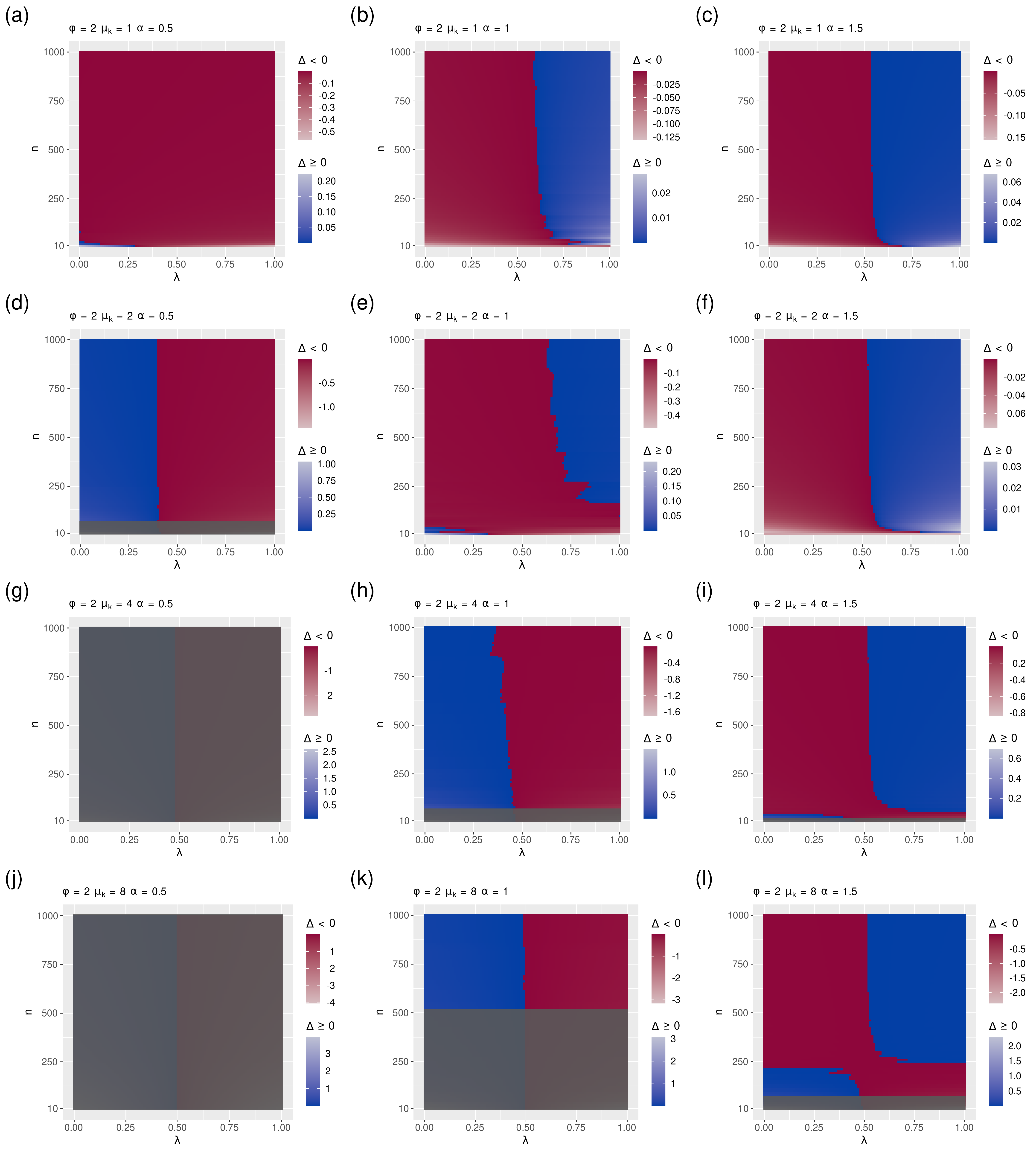}
\caption{\label{heatmap_lambda_vs_n_w_in_0_1_phi_2_figure} Figure equivalent to Fig. \ref{heatmap_lambda_vs_n_w_in_0_1_mu_continuous_phi_2_figure} after discretization of the $\mu_i'$s.
}
\end{figure}

\begin{figure}
\centering
\includegraphics[width = \textwidth]{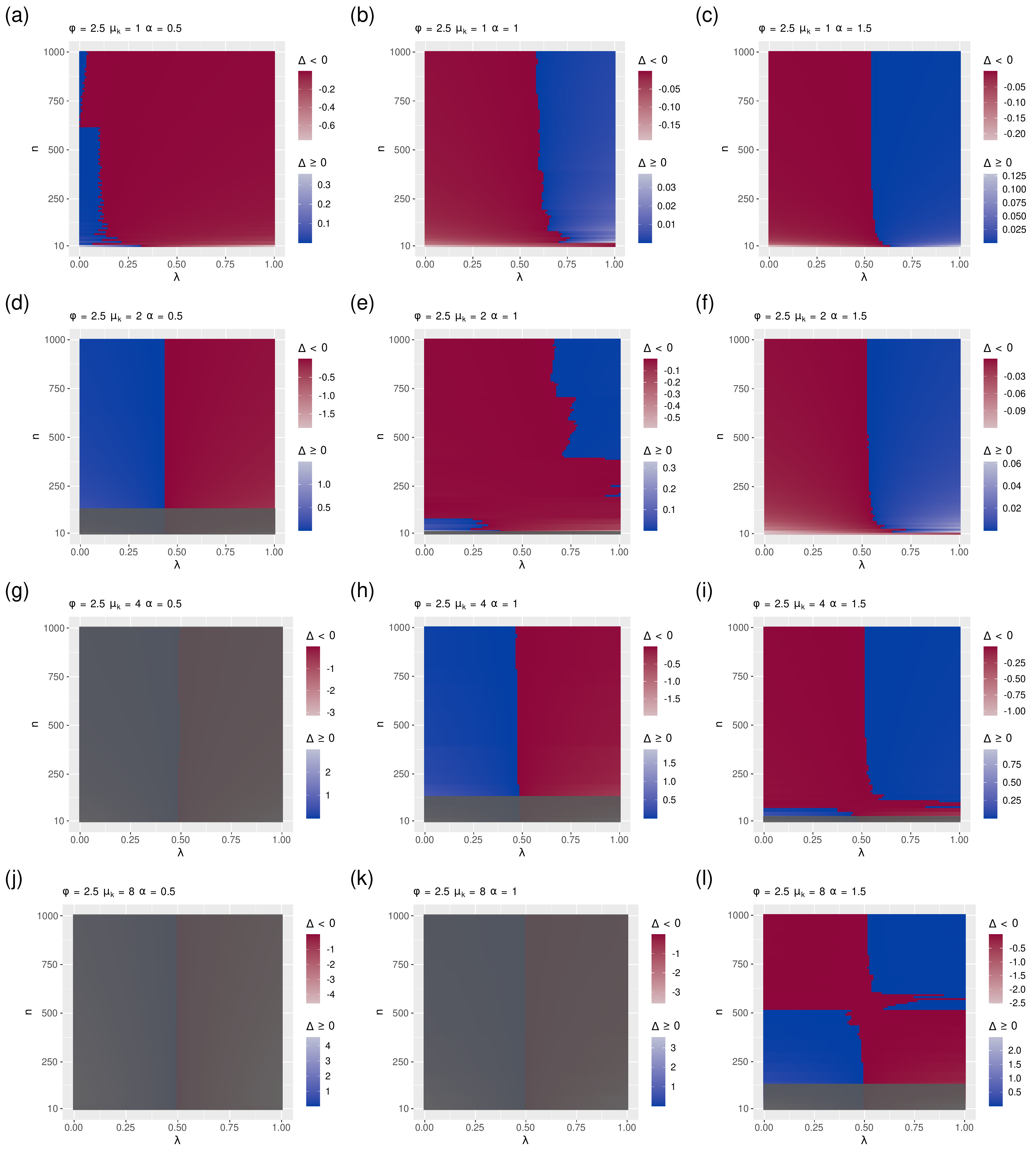}
\caption{\label{heatmap_lambda_vs_n_w_in_0_1_phi_2.5_figure} Figure equivalent to Fig. \ref{heatmap_lambda_vs_n_w_in_0_1_mu_continuous_phi_2.5_figure} after discretization of the $\mu_i'$s.
}
\end{figure}

\begin{figure}
\centering
\includegraphics[width = \textwidth]{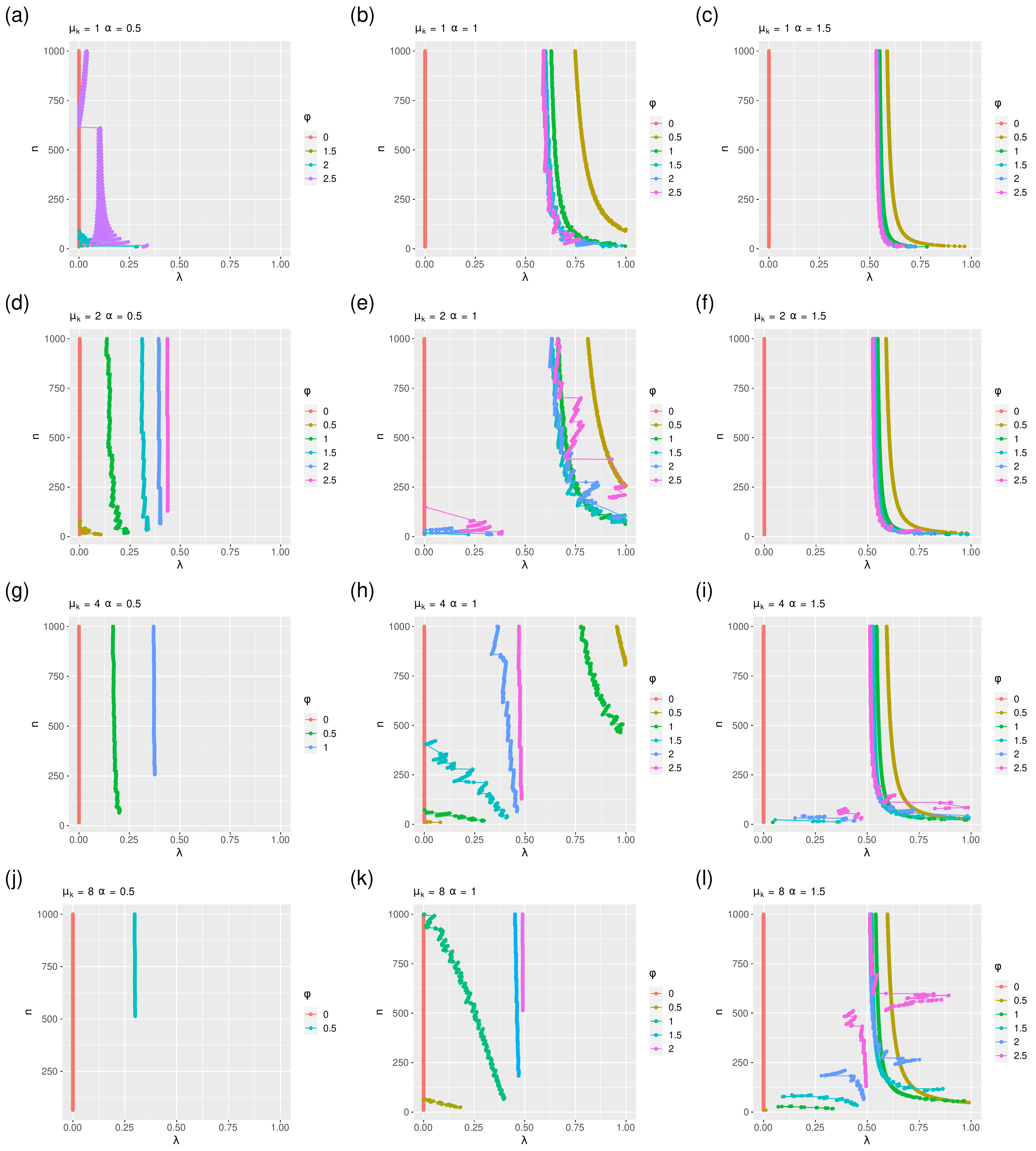}
\caption{\label{curves_delta_0_lambda_vs_n_w_in_0_1_various_phi_figure} Figure equivalent to Fig. \ref{curves_delta_0_lambda_vs_n_w_in_0_1_mu_continuous_various_phi_figure} after discretization of the $\mu_i'$s. It summarizes Figs. \ref{heatmap_lambda_vs_n_w_in_0_1_phi_0.5_figure},  \ref{heatmap_lambda_vs_n_w_in_0_1_phi_1_figure},  \ref{heatmap_lambda_vs_n_w_in_0_1_phi_1.5_figure},  \ref{heatmap_lambda_vs_n_w_in_0_1_phi_2_figure} and \ref{heatmap_lambda_vs_n_w_in_0_1_phi_2.5_figure}.
}
\end{figure}

The correspondence between the figures in this appendix with rounded form degrees and the figures in other sections is as follows. Figs. \ref{heatmap_lambda_vs_mu_k_w_in_0_1_phi_0_figure}, \ref{heatmap_lambda_vs_mu_k_w_in_0_1_phi_0.5_figure}, \ref{heatmap_lambda_vs_mu_k_w_in_0_1_phi_1_figure}, \ref{heatmap_lambda_vs_mu_k_w_in_0_1_phi_1.5_figure}, \ref{heatmap_lambda_vs_mu_k_w_in_0_1_phi_2_figure} and \ref{heatmap_lambda_vs_mu_k_w_in_0_1_phi_2.5_figure} are equivalent to Figs. \ref{heatmap_lambda_vs_mu_k_w_in_0_1_mu_continuous_phi_0_figure}, \ref{heatmap_lambda_vs_mu_k_w_in_0_1_mu_continuous_phi_0.5_figure}, \ref{heatmap_lambda_vs_mu_k_w_in_0_1_mu_continuous_phi_1_figure}, \ref{heatmap_lambda_vs_mu_k_w_in_0_1_mu_continuous_phi_1.5_figure}, \ref{heatmap_lambda_vs_mu_k_w_in_0_1_mu_continuous_phi_2_figure} and \ref{heatmap_lambda_vs_mu_k_w_in_0_1_mu_continuous_phi_2.5_figure}, respectively. These are the figures where $\lambda$ is on the $x$-axis and $\mu_k$ on the $y$-axis of the heatmap. Fig. \ref{curves_delta_0_lambda_vs_mu_k_w_in_0_1_various_phi_figure}, that summarizes the boundaries of the heatmaps, corresponds to Fig. \ref{curves_delta_0_lambda_vs_mu_k_w_in_0_1_mu_continuous_various_phi_figure} after discretization. Figs. \ref{heatmap_lambda_vs_alpha_w_in_0_1_phi_0.5_figure}, \ref{heatmap_lambda_vs_alpha_w_in_0_1_phi_1_figure}, \ref{heatmap_lambda_vs_alpha_w_in_0_1_phi_1.5_figure}, \ref{heatmap_lambda_vs_alpha_w_in_0_1_phi_2_figure} and \ref{heatmap_lambda_vs_alpha_w_in_0_1_phi_2.5_figure} are equivalent to Figs. \ref{heatmap_lambda_vs_alpha_w_in_0_1_mu_continuous_phi_0.5_figure}, \ref{heatmap_lambda_vs_alpha_w_in_0_1_mu_continuous_phi_1_figure}, \ref{heatmap_lambda_vs_alpha_w_in_0_1_mu_continuous_phi_1.5_figure}, \ref{heatmap_lambda_vs_alpha_w_in_0_1_mu_continuous_phi_2_figure} and \ref{heatmap_lambda_vs_alpha_w_in_0_1_mu_continuous_phi_2.5_figure}, respectively. In these figures, $\alpha$ is placed on the $y$-axis instead. Fig. \ref{curves_delta_0_lambda_vs_alpha_w_in_0_1_various_phi_figure} summarizes the boundaries and is the discretized version of Fig. \ref{curves_delta_0_lambda_vs_alpha_w_in_0_1_mu_continuous_various_phi_figure}. Finally, Fig. \ref{heatmap_lambda_vs_n_w_in_0_1_phi_0.5_figure}, \ref{heatmap_lambda_vs_n_w_in_0_1_phi_1_figure}, \ref{heatmap_lambda_vs_n_w_in_0_1_phi_1.5_figure}, \ref{heatmap_lambda_vs_n_w_in_0_1_phi_2_figure} and \ref{heatmap_lambda_vs_n_w_in_0_1_phi_2.5_figure} are equivalent to Figs. \ref{heatmap_lambda_vs_n_w_in_0_1_mu_continuous_phi_0.5_figure}, \ref{heatmap_lambda_vs_n_w_in_0_1_mu_continuous_phi_1_figure}, \ref{heatmap_lambda_vs_n_w_in_0_1_mu_continuous_phi_1.5_figure}, \ref{heatmap_lambda_vs_n_w_in_0_1_mu_continuous_phi_2_figure} and \ref{heatmap_lambda_vs_n_w_in_0_1_mu_continuous_phi_2.5_figure}, respectively. This set places $n$ on the $y$-axis. The boundaries in these last discretized figures are summarized by Fig. \ref{curves_delta_0_lambda_vs_n_w_in_0_1_various_phi_figure}, that corresponds to Figure \ref{curves_delta_0_lambda_vs_n_w_in_0_1_mu_continuous_various_phi_figure}.

We have presented two kinds of figures: heatmaps showing the value of $\Delta$ and figures summarizing the boundaries between regions where $\Delta>0$ and $\Delta<0$.
Interestingly, the discretization does not change the presence of regions where $\Delta <0$ and $\Delta>0$ and in general, it does not change the shape of the regions in a qualitative sense except in some cases where remarkable distortions appear (e.g., Figs. \ref{heatmap_lambda_vs_mu_k_w_in_0_1_phi_2_figure} or \ref{heatmap_lambda_vs_mu_k_w_in_0_1_phi_2.5_figure} have one or very few integer values on the $y$-axis for certain combinations of parameters, forming one dimensional bands that don't change over that axis; see also the distorted shapes in Figs. \ref{heatmap_lambda_vs_alpha_w_in_0_1_phi_2_figure} and specially \ref{heatmap_lambda_vs_n_w_in_0_1_phi_2.5_figure}). In contrast, the discretization has drastic impact on the summary plots of the boundary curves, where the curvy shapes of the continuous case are lost and altered substantially in many cases (Fig. \ref{curves_delta_0_lambda_vs_mu_k_w_in_0_1_various_phi_figure}, where some curves become one or a few points, or Fig. \ref{curves_delta_0_lambda_vs_alpha_w_in_0_1_various_phi_figure}, reflecting the loss of the curvy shapes).

%% file: article.bbl
\begin{thebibliography}{84}
\providecommand{\natexlab}[1]{#1}
\providecommand{\url}[1]{{#1}}
\providecommand{\urlprefix}{URL }
\expandafter\ifx\csname urlstyle\endcsname\relax
  \providecommand{\doi}[1]{DOI~\discretionary{}{}{}#1}\else
  \providecommand{\doi}{DOI~\discretionary{}{}{}\begingroup
  \urlstyle{rm}\Url}\fi
\providecommand{\eprint}[2][]{\url{#2}}

\bibitem[{Altmann(1993)}]{Altmann1993}
Altmann G (1993) Science and linguistics. In: R\"ohler R, Rieger B (eds)
  Contributions to Quantitative Linguistics, Kluwer, Dordrecht, Boston, London,
  pp 3--10

\bibitem[{Baixeries et~al.(2013)Baixeries, Elvev{\aa}g, and
  {Ferrer-i-Cancho}}]{Baixeries2012c}
Baixeries J, Elvev{\aa}g B, {Ferrer-i-Cancho} R (2013) The evolution of the
  exponent of {Zipf}'s law in language ontogeny. PLoS ONE 8(3):e53227

\bibitem[{Baronchelli et~al.(2013)Baronchelli, {Ferrer-i-Cancho},
  Pastor-Satorras, Chatter, and Christiansen}]{Baronchelli2013a}
Baronchelli A, {Ferrer-i-Cancho} R, Pastor-Satorras R, Chatter N, Christiansen
  M (2013) Networks in cognitive science. Trends in Cognitive Sciences
  17:348--360

\bibitem[{Bentz and {Ferrer-i-Cancho}(2016)}]{Bentz2016a}
Bentz C, {Ferrer-i-Cancho} R (2016) {Zipf's law of abbreviation as a language
  universal}. In: Bentz C, J{\"a}ger G, Yanovich I (eds) {Proceedings of the
  Leiden Workshop on Capturing Phylogenetic Algorithms for Linguistics},
  University of T{\"u}bingen

\bibitem[{Bion et~al.(2013)Bion, Borovsky, and Fernald}]{Bion2013a}
Bion RA, Borovsky A, Fernald A (2013) Fast {mapping,} slow learning:
  {Disambiguation} of novel word-object mappings in relation to vocabulary
  learning at {18,} {24,} and 30 months. Cognition 126(1):39--53,
  \doi{10.1016/j.cognition.2012.08.008}

\bibitem[{Brochhagen(2021)}]{Brochhagen2021a}
Brochhagen T (2021) Brief at the risk of being misunderstood: Consolidating
  population-and individual-level tendencies. Computational Brain {\&} Behavior
  \doi{10.1007/s42113-021-00099-x}

\bibitem[{Bunge(2001)}]{Bunge2001a}
Bunge M (2001) La science, sa m\'ethode et sa philosophie. Vigdor

\bibitem[{Byers-Heinlein and Werker(2013)}]{Byers-Heinlein2013a}
Byers-Heinlein K, Werker JF (2013) Lexicon structure and the disambiguation of
  novel words: {Evidence} from bilingual infants. Cognition 128(3):407--416,
  \doi{10.1016/j.cognition.2013.05.010}

\bibitem[{Casas et~al.(2018)Casas, Catal{\`a}, {Ferrer-i-Cancho},
  Hern\'andez-Fern\'andez, and Baixeries}]{Casas2019a}
Casas B, Catal{\`a} N, {Ferrer-i-Cancho} R, Hern\'andez-Fern\'andez A,
  Baixeries J (2018) The polysemy of the words that children learn over time.
  Interaction Studies 19(3):389 -- 426

\bibitem[{Chater and Brown(1999)}]{Chater1999a}
Chater N, Brown GDA (1999) Scale invariance as a unifying psychological
  principle. Cognition 69:1999

\bibitem[{Clark(1987)}]{Clark1987a}
Clark E (1987) The principle of contrast: {A} constraint on language
  acquisition. In: MacWhinney B (ed) Mechanisms of language acquisition,
  Lawrence Erlbaum Associates, Hillsdale, NJ

\bibitem[{Clark(1993)}]{Clark1993}
Clark E (1993) The lexicon in acquisition. Cambridge University Press

\bibitem[{Cormen et~al.(1990)Cormen, Leiserson, and
  Rivest}]{Cormen1990_summations}
Cormen TH, Leiserson CE, Rivest RL (1990) Introduction to algorithms, The MIT
  Press, Cambridge, MA, chap Chapter4. Summations

\bibitem[{Cover and Thomas(2006)}]{Cover2006a}
Cover TM, Thomas JA (2006) Elements of information theory. Wiley, New York, 2nd
  edition

\bibitem[{Dangli and Abazaj(2009)}]{Dangli2009a}
Dangli L, Abazaj G (2009) Absolute versus relative synonymy. Linguistic and
  Communicative Performance Journal 2:64--68

\bibitem[{Deacon(1997)}]{Deacon1997a}
Deacon TW (1997) The Symbolic Species: the Co-evolution of Language and the
  Brain. W. W. Norton \& Company, New York

\bibitem[{Deacon(2015)}]{Deacon2015a}
Deacon TW (2015) Steps to a science of biosemiotics. Green Letters
  19(3):293--311, \doi{10.1080/14688417.2015.1072948}

\bibitem[{Debowski(2020)}]{Debowski2020a}
Debowski L (2020) Information Theory Meets Power Laws: Stochastic Processes and
  Language Models. Wiley, Hoboken, NJ

\bibitem[{Eco(1986)}]{Eco1986a}
Eco U (1986) Semiotics and the philosophy of language. Indiana University
  Press, Bloomington

\bibitem[{Ellis and Hitchcock(1986)}]{Ellis1986}
Ellis SR, Hitchcock RJ (1986) The emergence of {Z}ipf's law: spontaneous
  encoding by users of a command language. IEEE Trans Syst Man Cyber
  16(3):423--427

\bibitem[{{Eun-Nam}(2017)}]{EunNam2017a}
{Eun-Nam} S (2017) Word learning characteristics of 3-to 6-year-olds: Focused
  on the mutual exclusivity assumption. Journal of speech-language {\&} hearing
  disorders 26(4):33--40

\bibitem[{{Ferrer-i-Cancho}(2005{\natexlab{a}})}]{Ferrer2004a}
{Ferrer-i-Cancho} R (2005{\natexlab{a}}) The variation of {Zipf's} law in human
  language. European Physical Journal B 44:249--257

\bibitem[{{Ferrer-i-Cancho}(2005{\natexlab{b}})}]{Ferrer2004e}
{Ferrer-i-Cancho} R (2005{\natexlab{b}}) {Zipf's} law from a communicative
  phase transition. European Physical Journal B 47:449--457,
  \doi{10.1140/epjb/e2005-00340-y}

\bibitem[{{Ferrer-i-Cancho}(2006)}]{Ferrer2005e}
{Ferrer-i-Cancho} R (2006) When language breaks into pieces. {A} conflict
  between communication through isolated signals and language. Biosystems
  84:242--253

\bibitem[{{Ferrer-i-Cancho}(2016{\natexlab{a}})}]{Ferrer2016b}
{Ferrer-i-Cancho} R (2016{\natexlab{a}}) Compression and the origins of
  {Zipf's} law for word frequencies. Complexity 21:409--411

\bibitem[{{Ferrer-i-Cancho}(2016{\natexlab{b}})}]{Ferrer2014d}
{Ferrer-i-Cancho} R (2016{\natexlab{b}}) The meaning-frequency law in {Zipfian}
  optimization models of communication. Glottometrics 35:28--37

\bibitem[{{Ferrer-i-Cancho}(2017{\natexlab{a}})}]{Ferrer2013g}
{Ferrer-i-Cancho} R (2017{\natexlab{a}}) The optimality of attaching unlinked
  labels to unlinked meanings. Glottometrics 36:1--16

\bibitem[{{Ferrer-i-Cancho}(2017{\natexlab{b}})}]{Ferrer2013f}
{Ferrer-i-Cancho} R (2017{\natexlab{b}}) The placement of the head that
  maximizes predictability. {An} information theoretic approach. Glottometrics
  39:38--71

\bibitem[{{Ferrer-i-Cancho}(2018)}]{Ferrer2015b}
{Ferrer-i-Cancho} R (2018) Optimization models of natural communication.
  Journal of Quantitative Linguistics 25(3):207--237

\bibitem[{{Ferrer-i-Cancho} and D{\'{i}}az-Guilera(2007)}]{Ferrer2007a}
{Ferrer-i-Cancho} R, D{\'{i}}az-Guilera A (2007) {The global minima of the
  communicative energy of natural communication systems}. Journal of
  Statistical Mechanics: Theory and Experiment 06009(6),
  \doi{10.1088/1742-5468/2007/06/P06009}

\bibitem[{{Ferrer-i-Cancho} and Sole(2003)}]{Ferrer2002a}
{Ferrer-i-Cancho} R, Sole RV (2003) {Least effort and the origins of scaling in
  human language}. Proceedings of the National Academy of Sciences of the
  United States of America 100(3):788--791, \doi{10.1073/pnas.0335980100}

\bibitem[{{Ferrer-i-Cancho} and Vitevitch(2018)}]{Ferrer2017b}
{Ferrer-i-Cancho} R, Vitevitch M (2018) The origins of {Zipf's}
  meaning-frequency law. Journal of the American Association for Information
  Science and Technology 69(11):1369--1379

\bibitem[{{Ferrer-i-Cancho} et~al.(2005){Ferrer-i-Cancho}, Riordan, and
  Bollob\'as}]{Ferrer2004f}
{Ferrer-i-Cancho} R, Riordan O, Bollob\'as B (2005) The consequences of
  {Zipf's} law for syntax and symbolic reference. Proceedings of the Royal
  Society of London B 272:561--565

\bibitem[{{Ferrer-i-Cancho} et~al.(2019){Ferrer-i-Cancho}, Bentz, and
  Seguin}]{Ferrer2019c}
{Ferrer-i-Cancho} R, Bentz C, Seguin C (2019) Optimal coding and the origins of
  {Zipfian} laws. Journal of Quantitative Linguistics p in press,
  \doi{10.1080/09296174.2020.1778387}

\bibitem[{Frank and Poulin-Dubois(2002)}]{Frank2002a}
Frank I, Poulin-Dubois D (2002) Young monolingual and bilingual children's
  responses to violation of the mutual exclusivity principle. International
  Journal of Bilingualism 6(2):125--146, \doi{10.1177/13670069020060020201}

\bibitem[{Frank et~al.(2009)Frank, Goodman, and Tenenbaum}]{Frank2009a}
Frank MC, Goodman ND, Tenenbaum JB (2009) Using speakers' referential
  intentions to model early cross-situational word learning. Psychological
  Science 20(5):578--585, \doi{10.1111/j.1467-9280.2009.02335.x}

\bibitem[{Fromkin et~al.(2014)Fromkin, Rodman, and Hyams}]{Fromkin2014a}
Fromkin V, Rodman R, Hyams N (2014) An Introduction to Language, 10th edn.
  Wadsworth Publishing, Boston, MA

\bibitem[{Futrell(2020)}]{Futrell2020_twitter}
Futrell R (2020)
  \url{https://twitter.com/rljfutrell/status/1275834876055351297}

\bibitem[{Gandhi and Lake(2020)}]{Gandhi2019a}
Gandhi K, Lake B (2020) Mutual exclusivity as a challenge for deep neural
  networks. In: Advances in Neural Information Processing Systems (NeurIPS), 33

\bibitem[{Genty and Zuberbühler(2014)}]{Genty2014a}
Genty E, Zuberbühler K (2014) Spatial reference in a bonobo gesture. Current
  Biology 24(14):1601--1605, \doi{https://doi.org/10.1016/j.cub.2014.05.065}

\bibitem[{Gibson et~al.(2019)Gibson, Futrell, Piantadosi, Dautriche, Mahowald,
  Bergen, and Levy}]{Gibson2019a}
Gibson E, Futrell R, Piantadosi S, Dautriche I, Mahowald K, Bergen L, Levy R
  (2019) How efficiency shapes human language. Trends in Cognitive Sciences
  23:389--407

\bibitem[{Greene et~al.(2013)Greene, Pe{\~{n}}a, and Bedore}]{Greene2013a}
Greene KJ, Pe{\~{n}}a ED, Bedore LM (2013) Lexical choice and language
  selection in bilingual preschoolers. Child Language Teaching and Therapy
  29(1):27--39, \doi{10.1177/0265659012459743}

\bibitem[{Gulordava et~al.(2020)Gulordava, Brochhagen, and
  Boleda}]{Gulordava2020a}
Gulordava K, Brochhagen T, Boleda G (2020) Deep daxes: {Mutual} exclusivity
  arises through both learning biases and pragmatic strategies in neural
  networks. In: Proceedings of CogSci 2020, pp 2089--2095

\bibitem[{Gustison et~al.(2016)Gustison, Semple, {Ferrer-i-Cancho}, and
  Bergman}]{Gustison2016a}
Gustison ML, Semple S, {Ferrer-i-Cancho} R, Bergman T (2016) Gelada vocal
  sequences follow {Menzerath}'s linguistic law. Proceedings of the National
  Academy of Sciences USA 13(19):E2750--E2758, \doi{10.1073/pnas.1522072113}

\bibitem[{Halberda(2003)}]{Halberda2003a}
Halberda J (2003) The development of a word-learning strategy. Cognition
  87(1):23--34, \doi{10.1016/S0010-0277(02)00186-5}

\bibitem[{Haryu(1991)}]{Haryu1991a}
Haryu E (1991) A developmental study of children's use of mutual exclusivity
  and context to interpret novel words. The Japanese Journal of Educational
  Psychology 39(1):11--20, \doi{10.5926/jjep1953.39.1_11}

\bibitem[{Hendrickson and Perfors(2019)}]{Hendrickson2019a}
Hendrickson AT, Perfors A (2019) Cross-situational learning in a {Zipfian}
  environment. Cognition 189(February):11--22,
  \doi{10.1016/j.cognition.2019.03.005}

\bibitem[{Hobaiter and Byrne(2014)}]{Hobaiter2014a}
Hobaiter C, Byrne RW (2014) The meanings of chimpanzee gestures. Current
  Biology 24:1596--1600

\bibitem[{Houston-Price et~al.(2010)Houston-Price, Caloghiris, and
  Raviglione}]{Houston-Price2010a}
Houston-Price C, Caloghiris Z, Raviglione E (2010) Language experience shapes
  the development of the mutual exclusivity bias. Infancy 15(2):125--150,
  \doi{10.1111/j.1532-7078.2009.00009.x}

\bibitem[{Hung et~al.(2015)Hung, Patrycia, and Yow}]{Hung2015a}
Hung WY, Patrycia F, Yow WQ (2015) Bilingual children weigh speaker's
  referential cues and word-learning heuristics differently in different
  language contexts when interpreting a speaker's intent. Frontiers in
  Psychology 6(JUN):1--9, \doi{10.3389/fpsyg.2015.00796}

\bibitem[{Hurford(1989)}]{Hurford1989a}
Hurford J (1989) Biological evolution of the {Saussurean} sign as a component
  of the language acquisition device. Lingua 77:187--222,
  \doi{doi:10.1016/0024-3481(89)90015-6}

\bibitem[{Kalashnikova et~al.(2015)Kalashnikova, Mattock, and
  Monaghan}]{Kalashnikova2015a}
Kalashnikova M, Mattock K, Monaghan P (2015) The effects of linguistic
  experience on the flexible use of mutual exclusivity in word learning.
  Bilingualism 18(4):626--638, \doi{10.1017/S1366728914000364}

\bibitem[{Kalashnikova et~al.(2016)Kalashnikova, Mattock, and
  Monaghan}]{Kalashnikova2016a}
Kalashnikova M, Mattock K, Monaghan P (2016) Flexible use of mutual exclusivity
  in word learning. Language Learning and Development 12(1):79--91,
  \doi{10.1080/15475441.2015.1023443}

\bibitem[{Kalashnikova et~al.(2019)Kalashnikova, Oliveri, and
  Mattock}]{Kalashnikova2019a}
Kalashnikova M, Oliveri A, Mattock K (2019) Acceptance of lexical overlap by
  monolingual and bilingual toddlers. International Journal of Bilingualism
  23(6):1517--1530, \doi{10.1177/1367006918808041}

\bibitem[{Kaminski et~al.(2004)Kaminski, Call, and Fischer}]{Kaminski2004a}
Kaminski J, Call J, Fischer J (2004) Word learning in a domestic dog:
  {Evidence} for ``fast mapping''. Science 304(5677):1682--1683,
  \doi{10.1126/science.1097859}

\bibitem[{Kanwal et~al.(2017)Kanwal, Smith, Culbertson, and
  Kirby}]{Kanwal2017a}
Kanwal J, Smith K, Culbertson J, Kirby S (2017) Zipf's law of abbreviation and
  the principle of least effort: Language users optimise a miniature lexicon
  for efficient communication. Cognition 165:45--52

\bibitem[{Kello et~al.(2010)Kello, Brown, {Ferrer-i-Cancho}, Holden,
  Linkenkaer-Hansen, Rhodes, and Orden}]{Kello2010a}
Kello CT, Brown GDA, {Ferrer-i-Cancho} R, Holden JG, Linkenkaer-Hansen K,
  Rhodes T, Orden GCV (2010) Scaling laws in cognitive sciences. Trends in
  cognitive sciences 14(5):223--232, \doi{10.1016/j.tics.2010.02.005}

\bibitem[{K\"{o}hler(1987)}]{Koehler1987}
K\"{o}hler R (1987) System theoretical linguistics. Theor Linguist
  14(2-3):241--257

\bibitem[{Kull(1999)}]{Kull1999a}
Kull K (1999) Biosemiotics in the twentieth century: {A} view from biology.
  Semiotica 127(1/4):385–414

\bibitem[{Kull(2018)}]{Kull2018a}
Kull K (2018) Choosing and learning: {Semiosis} means choice. Sign Systems
  Studies 46(4):452–466

\bibitem[{Kull(2020)}]{Kull2020a}
Kull K (2020) Codes: {Necessary,} but not sufficient for meaning-making.
  Constructivist Foundations 15(2):137–139

\bibitem[{Li et~al.(2010)Li, Miramontes, and Cocho}]{Li2010a}
Li W, Miramontes P, Cocho G (2010) Fitting ranked linguistic data with
  two-parameter functions. Entropy 12(7):1743--1764

\bibitem[{Liittschwager and Markman(1994)}]{Liittschwager1994a}
Liittschwager JC, Markman EM (1994) Sixteen-and 24-month-olds' use of mutual
  exclusivity as a default assumption in second-label learning. Developmental
  Psychology 30(6):955--968, \doi{10.1037/0012-1649.30.6.955}

\bibitem[{Lund and Burgess(1996)}]{Lund1996a}
Lund K, Burgess C (1996) Producing high-dimensional semantic spaces from
  lexical co-occurrence. Behavior Research Methods, Instruments, and Computers
  28(2):203--208

\bibitem[{Markman and Wachtel(1988)}]{Markman1988a}
Markman E, Wachtel G (1988) Children's use of mutual exclusivity to constrain
  the meanings of words. Cognitive Psychology 20:121--157

\bibitem[{Merriman and Bowman(1989)}]{Merriman1989a}
Merriman WW, Bowman LL (1989) The mutual exclusivity bias in children's word
  learning. Monographs of the Society for Research in Child Development
  54:1‐129

\bibitem[{Meylan and Griffiths(2021)}]{Meylan2021a}
Meylan S, Griffiths T (2021) The challenges of large-scale, web-based language
  datasets: Word length and predictability revisited. PsyArXiv
  \doi{10.31234/osf.io/6832r}, \urlprefix\url{psyarxiv.com/6832r}

\bibitem[{Moore(2014)}]{Moore2014a}
Moore R (2014) Ape gestures: Interpreting chimpanzee and bonobo minds. Current
  Biology 24(14):R645--R647, \doi{10.1016/j.cub.2014.05.072}

\bibitem[{Nicoladis and Laurent(2020)}]{Nicoladis2020a}
Nicoladis E, Laurent A (2020) When knowing only one word for ``car'' leads to
  weak application of mutual exclusivity. Cognition 196(February 2019):104087,
  \doi{10.1016/j.cognition.2019.104087}

\bibitem[{Nicoladis and Secco(2000)}]{Nicoladis2000a}
Nicoladis E, Secco G (2000) The role of a child's productive vocabulary in the
  language choice of a bilingual family. First Language 20(58):003--28,
  \doi{10.1177/014272370002005801}

\bibitem[{Piantadosi(2014)}]{Piantadosi2014a}
Piantadosi S (2014) Zipf's law in natural language: a critical review and
  future directions. Psych onomic Bulletin and Review 21:1112--1130

\bibitem[{Piantadosi et~al.(2011)Piantadosi, Tily, and
  Gibson}]{Piantadosi2011a}
Piantadosi ST, Tily H, Gibson E (2011) Word lengths are optimized for efficient
  communication. Proceedings of the National Academy of Sciences
  108(9):3526--3529

\bibitem[{Piotrowski and Spivak(2007)}]{Piotrowski2007a}
Piotrowski RG, Spivak DL (2007) Linguistic disorders and pathologies:
  synergetic aspects. In: Grzybek P, K\"ohler R (eds) Exact methods in the
  study of language and text. To honor Gabriel Altmann, Gruyter, Berlin, pp
  545--554

\bibitem[{Pulvermuller(2001)}]{Pulvermuller2001}
Pulvermuller F (2001) Brain reflections of words and their meaning. Trends in
  Cognitive Sciences 5(12):517--524

\bibitem[{Pulvermüller(2013)}]{Pulvermuller2013a}
Pulvermüller F (2013) How neurons make meaning: brain mechanisms for embodied
  and abstract-symbolic semantics. Trends in Cognitive Sciences 17(9):458--470,
  \doi{https://doi.org/10.1016/j.tics.2013.06.004}

\bibitem[{Saxton(2010)}]{Saxton2010a_Chapter6}
Saxton M (2010) Child language. Acquisition and development, {SAGE}, Los
  Angeles, chap 6. The developing lexicon: what's in a name?, pp 133--158

\bibitem[{Steels(1996)}]{Steels1996a}
Steels L (1996) The spontaneous self-organization of an adaptive language.
  Machine Intelligence 15:205--224

\bibitem[{Stewart and Plotkin(2021)}]{Stewart2021a}
Stewart AJ, Plotkin JB (2021) The natural selection of good science. Nature
  Human Behaviour \doi{10.1038/s41562-021-01111-x}

\bibitem[{Yildiz(2020)}]{Yildiz2020a}
Yildiz M (2020) Conflicting nature of social-pragmatic cues with mutual
  exclusivity regarding three-year-olds' label-referent mappings. Psychology of
  Language and Communication 24(1):124--141, \doi{10.2478/plc-2020-0008}

\bibitem[{Yurovsky and Yu(2008)}]{Yurovsky2008a}
Yurovsky D, Yu C (2008) Mutual exclusivity in crosssituational statistical
  learning. Proceedings of the annual meeting of the cognitive science society
  pp 715--720

\bibitem[{Zaslavsky et~al.(2018)Zaslavsky, Kemp, Regier, and
  Tishby}]{Zaslavsky2018a}
Zaslavsky N, Kemp C, Regier T, Tishby N (2018) Efficient compression in color
  naming and its evolution. Proceedings of the National Academy of Sciences
  115(31):7937--7942, \doi{10.1073/pnas.1800521115}

\bibitem[{Zaslavsky et~al.(2021)Zaslavsky, Maldonado, and
  Culbertson}]{Zaslavsky2021a}
Zaslavsky N, Maldonado M, Culbertson J (2021) Let's talk (efficiently) about
  us: {Person} systems achieve near-optimal compression. PsyArXiv
  \doi{10.31234/osf.io/kcu27}, \urlprefix\url{psyarxiv.com/kcu27}

\bibitem[{Zipf(1945)}]{Zipf1945a}
Zipf GK (1945) The meaning-frequency relationship of words. Journal of General
  Psychology 33:251--266

\bibitem[{Zipf(1949)}]{Zipf1949a}
Zipf GK (1949) Human behaviour and the principle of least effort.
  Addison-Wesley, Cambridge (MA), USA

\end{thebibliography}
